\documentclass[journal]{IEEEtran}
\ifCLASSINFOpdf
\else
\fi
\ifCLASSOPTIONcompsoc
\usepackage[caption=false,font=normalsize,labelfont=sf,textfont=sf,farskip=0pt,justification=centering]{subfig}
\else
\usepackage[caption=false,font=footnotesize,farskip=0pt,justification=centering]{subfig}
\fi

\usepackage{hyperref}
\usepackage{algorithm}
\usepackage[noend]{algpseudocode}
\usepackage{pgfplots}
\usepackage{tikz}

\usepackage{amsfonts}
\usepackage{amsthm}
\usepackage{amsmath} 

\usepackage{booktabs}
\usepackage{multirow}

\usepackage{amssymb}

\usepackage{bm}
\usepackage{mathtools}

\usepackage{setspace}

\usepackage[noadjust]{cite}

\usepackage{csquotes}

\usetikzlibrary{automata}

\pgfplotsset{compat=1.17}

\DeclareMathOperator*{\argmin}{arg\,min}

\renewcommand{\vec}[1]{\mathbf{#1}}
\newcommand{\mtx}[1]{\bm{\mathit{#1}}}

\usepackage{tabularx}
\newcolumntype{C}{>{\centering\arraybackslash}X} 
\newcolumntype{F}{>{\hsize=.65\hsize}C}
\newcolumntype{S}{>{\hsize=.6\hsize}C}
\newcolumntype{B}{>{\hsize=1.15\hsize}C}
\newcolumntype{N}{>{\hsize=2.4\hsize}C}
\newcolumntype{R}{>{\hsize=1.3\hsize}C}
\newcolumntype{I}{>{\hsize=1\hsize}C}

\newcommand{\example}[1]{\tilde{#1}}

\makeatletter
\renewcommand{\ALG@name}{Pseudocode}
\makeatother

\algrenewcommand\alglinenumber[1]{\tiny #1:}
\algrenewcommand\algorithmicindent{0.99em}

\begin{document}
	
	%
	\title{Incremental Recursive Ranking Grouping for Large Scale Global Optimization}
	%
	%
	%
	
	\author{Marcin~Michal~Komarnicki,
		Michal~Witold~Przewozniczek,
		Halina~Kwasnicka,
		and~Krzysztof~Walkowiak
		\thanks{This work was supported by the Polish National Science Centre (NCN) under Grant 2020/38/E/ST6/00370.}
		
		\thanks{The authors are with the Wroclaw University of Science and Technology, 50-370 Wroclaw, Poland
			(e-mail: \href{mailto:marcin.komarnicki@pwr.edu.pl}{marcin.komarnicki@pwr.edu.pl}; \href{mailto:michal.przewozniczek@pwr.edu.pl}{michal.przewozniczek@pwr.edu.pl}; \href{mailto:halina.kwasnicka@pwr.edu.pl}{halina.kwasnicka@pwr.edu.pl}; \href{mailto:krzysztof.walkowiak@pwr.edu.pl}{krzysztof.walkowiak@pwr.edu.pl}).}}

	\maketitle
	
	\begin{abstract}
		
		Real-world optimization problems may have a different underlying structure. In black-box optimization, the dependencies between decision variables remain unknown. However, some techniques can discover such interactions accurately. In Large Scale Global Optimization~(LSGO), problems are high-dimensional. It was shown effective to decompose LSGO problems into subproblems and optimize them separately. The effectiveness of such approaches may be highly dependent on the accuracy of problem decomposition. Many state-of-the-art decomposition strategies are derived from Differential Grouping~(DG). However, if a given problem consists of non-additively separable subproblems, DG-based strategies may discover many non-existing interactions. On the other hand, monotonicity checking strategies proposed so far do not report non-existing interactions for any separable subproblems but may miss discovering many of the existing ones. Therefore, we propose Incremental Recursive Ranking Grouping~(IRRG) that suffers from none of these flaws. IRRG consumes more fitness function evaluations than the recent DG-based propositions, e.g., Recursive DG 3 (RDG3). Nevertheless, the effectiveness of the considered Cooperative Co-evolution frameworks after embedding IRRG or RDG3 was similar for problems with additively separable subproblems that are suitable for RDG3. After replacing the additive separability with non-additive, embedding IRRG leads to results of significantly higher quality.
		
	\end{abstract}
	
	

	%
	\IEEEpeerreviewmaketitle

	\section{Introduction}
	\label{sec:introduction}
	%
	%
	%
	%
	
	\IEEEPARstart{I}{N} practice, many continuous Large Scale Global Optimization (LSGO) problems emerge~\cite{lsgoIntro,rw1,rw2}. They were classified as \textit{complex continuous optimization problems}~\cite{lsgoComplex}. Thus, solving them may be considered important but difficult. Real-value encoded instances are classified as large if they have at least $500$ decision variables. However, it is frequent to consider the LSGO instances with at least $1000$ variables~\cite{lsgoIntro,lsgoComplex}. The size of the search space is exponentially proportional to the number of problem dimensions~\cite{lsgoComplex}. Thus, increasing the dimensionality may cause the exponential growth of the number of local optima~\cite{lsgoIntro} or the enormously high cost of running Estimation of Distribution Algorithms~\cite{edaLSGO}. Therefore, tackling LSGO instances may be hard for Evolutionary Algorithms designed to solve less dimensional optimization problems.
	
	In the state-of-the-art LSGO-dedicated optimizers, two main approaches are considered. The first, involves hybrid optimization methods, which do not utilize problem decomposition, e.g., Success-History Based Parameter Adaptation for Differential Evolution with Iterative Local Search~(SHADE-ILS)~\cite{shade-ils}. The other approach is to decompose a problem first and then optimize its subproblems independently, frequently using Cooperative Co-evolution (CC)~\cite{cc,cbcc,ccfr2,cc2018}. Nowadays, the state-of-the-art problem decomposition strategies derive from Differential Grouping (DG)~\cite{dg,xdg,dg2,rdg,rdg2,rdg3}. These strategies assume that problem consists of additively separable subproblems. If this assumption is false, the DG-based strategies may report \textit{false linkage} that takes place when two independent variables are found dependent~\cite{3lo}. False linkage may significantly decrease the quality of the decomposition and decrease the overall method effectiveness~\cite{linkageQuality}. Decomposition strategies that do not assume the existence of only additively separable subproblems are based on monotonicity checking~\cite{lvi,ccvil,svil,fvil}. However, they may miss finding some of the interactions (\textit{missing linkage} \cite{3lo}), and DG-based strategies where shown to be less vulnerable to this inconsistency~\cite{dg,fvil,dg2,rdg}.
	
	In this article, we propose IRRG, a new problem decomposition strategy that does not assume the existence of only additively separable subproblems. IRRG is derived from monotonicity checking, thus, it never reports false linkage. Additionally, it significantly limits the issue of missing linkage. This paper has two main objectives. First, we show that IRRG can decompose continuous LSGO problems accurately regardless of the additive or non-additive separability. Second, we show that although IRRG is more expensive than Recursive DG 3 (RDG3)~\cite{rdg3}, its influence on the overall optimization cost is negligible also when the problems are DG-suitable. To meet these objectives, we embed both, IRRG and RDG3, into two different CC frameworks. We consider a set of $45$ problems including additively and non-additively separable ones. This set is built from the CEC'2013 functions~\cite{cec2013}. Since the standard CEC'2013 set contains only one function with non-additively separable subfunctions, we propose two transformations of the additive separability into non-additive.\par
	
	The rest of this article is organized as follows. The related work is discussed in the next section. In Section~\ref{sec:comp}, we compare DG and monotonicity checking. Section~\ref{sec:irrg} presents the details of IRRG, whereas the fifth section reports the results of the experiments. Finally, in the last section, we conclude this paper and propose the directions of future work.\par

	\section{Related Work}
	
	\subsection{Problem Decomposition}
	\label{sec:related:decomposition}
	The quality of the problem decomposition is related to the separability structure of its objective function~\cite{dg2}. A function $f: \Omega \rightarrow \mathbb R$ is partially separable~\cite{lsgo_bench_design}, i.e., consists of $m$ independent subfunctions, $2 \le m \le n$, if
	\begin{equation}
	\small
	\label{eq:decomposition}
	\argmin_{\vec{x}} f(\vec{x}) = \bigg[\argmin_{\vec{x_1}} f(\vec{x_1}, ...), ..., \argmin_{\vec{x_m}} f(..., \vec{x_m})\bigg]
	\end{equation}
	where $\vec{x} \in \Omega$ is a vector of $n$ decision variables $[x_1, ..., x_n]$, $\forall_{i \in \{1, ..., m\}} \vec{x_i} \in \Omega_i$, and $\Omega = \Omega_1 \times ... \times \Omega_m$. When $m = n$ then each decision variable does not interact with the others and a function is fully separable.
	
	Problem decomposition strategies include uniformed decomposition~\cite{uniformedDecomposition}, random grouping~\cite{randomGrouping}, delta grouping~\cite{deltaGrouping}, meta modelling~\cite{metaModelling}, formula-based grouping~\cite{formulaGrouping}, differential grouping~\cite{dg,xdg,dg2,rdg,rdg2,rdg3}, and monotonicity checking~\cite{lvi, ccvil, svil, fvil}. The last two approaches are highly related to this paper's scope. Thus, we describe them in detail.
	
	\subsubsection{Differential Grouping}
	\label{sec:related:dec:dg}
	
	Many real-world optimization problems are partially separable~\cite{modularNature}, e.g., additively separable~\cite{dg}. The definition of an additively separable function is similar to formula~(\ref{eq:decomposition}): $f(\vec{x}) = \sum_{i = 1}^{m} f_i(\vec{x_i})$. Differential Grouping (DG)~\cite{dg} is a problem decomposition strategy proven to group only dependent variables if a problem is additively separable. DG marks two variables as interacting if $\exists a, b_1 \ne b_2, \delta > 0, \vec{x^*} \in \Omega$ the following condition holds
	\begin{equation}
	\small
	\label{eq:deltaCheck}
	\Delta_{\delta,x_p}[f](\vec{x^*})|_{x_p=a,x_q=b_1} \ne \Delta_{\delta,x_p}[f](\vec{x^*})|_{x_p=a,x_q=b_2}
	\end{equation}
	where $x_p$ and $x_q$ indicate the $p$th and $q$th problem variables, whereas the definition of function $\Delta_{\delta,x_p}[f]$ is as follows
	\begin{equation}
	\small
	\label{eq:deltaCheck:delta}
	\Delta_{\delta,x_p}[f](\vec{x}) = f(..., x_p + \delta, ...) - f(..., x_p, ...)
	\end{equation}
	By $f(\vec{x})|_{x_p=a,x_q=b}$ we mean the value of $f$ after setting the values of $x_p$ and $x_q$ to $a$ and $b$, respectively. Each perturbation of the $p$th variable, i.e., $x_p + \delta$, must produce a feasible solution. Without loss of generality, we may denote the left and right sides of formula~(\ref{eq:deltaCheck}) as $\Delta_1$ and $\Delta_2$, respectively. Due to the inaccuracy of the float numbers representation, a user-defined $\epsilon$ is employed to control the sensitivity of DG: $|\Delta_1 - \Delta_2| > \epsilon$ instead of $\Delta_1 \ne \Delta_2$. Its appropriate value may be different for various optimization problems~\cite{dg2}. DG optimizes the number of pair-wise checks (the single interaction check requires four fitness function evaluations~(FFEs)) by ignoring some pairs of variables. The worst scenario is when problem is fully separable, then, $2n(n-1)$ FFEs are needed. The minimal cost, for a fully non-separable problem, is $4(n-1)$ FFEs. Thus, the DG's time complexity is $\mathcal{O}(n^2)$.\par
	
	DG2 is an improved version of DG~\cite{dg2}. It decreases by half the FFE cost of decomposition for fully separable problems. Additionally, DG2 can detect the overlapping structure of a given optimization problem by checking all possible pairs of variables. In DG2, $\epsilon$ is automatically computed as follows. Each interaction check requires fitness values of four solutions: $\vec{x_1^*}$, $\vec{x_2^*}$, $\vec{x_3^*}$, and $\vec{x_4^*}$. Thus, $|\Delta_1 - \Delta_2| > \epsilon$ can be reformulated as $|[f(\vec{x_1^*}) - f(\vec{x_2^*})] - [f(\vec{x_3^*}) - f(\vec{x_4^*})]| > \epsilon$, where
	\begin{equation}
	\small
	\label{eq:dg2:epsilon}
	\epsilon = f_{\gamma}(\sqrt{n} + 2) \cdot (|f(\vec{x_1^*})| + |f(\vec{x_2^*})| + |f(\vec{x_3^*})| + |f(\vec{x_4^*})|)
	\end{equation}
	\begin{equation}
	\small
	\label{eq:dg2:gamma}
	f_{\gamma}(k) = \frac{k \mu_M}{1 - k \mu_M}
	\end{equation}
	where $\mu_M$ is a machine dependent constant~\cite{floatingPoint}.
	
	\subsubsection{Recursive Differential Grouping}
	
	Recursive Differential Grouping (RDG)~\cite{rdg} reduces the complexity of DG2 from $\mathcal{O}(n^2)$ to $\mathcal{O}(n\log(n))$. DG and DG2 perform the pair-wise interaction check, whereas RDG consider two disjoint groups of variables $X_1$ and $X_2$ that are subsets of $X = \{x_1, ..., x_n\}$. Groups interact if at least one pair of variables $x_p \in X_1$ and $x_q \in X_2$ is non-separable. Then, there exist such values $\delta_1$, $\delta_2 > 0$, unit vectors $\vec{u_1} \in U_{X_1}$, $\vec{u_2} \in U_{X_2}$, and decision vector $\vec{x^*} \in \Omega$ that meet
	\begin{equation}
	\small
	\label{eq:rdg:nonseparability}
	f(\vec{x^*} + \delta_1 \vec{u_1} + \delta_2 \vec{u_2}) - f(\vec{x^*} + \delta_2 \vec{u_2}) \ne  f(\vec{x^*} + \delta_1 \vec{u_1}) - f(\vec{x^*})
	\end{equation}
	where $\vec{u_j} = [u_1, ..., u_n] \in U_{X_j}$ such that $\forall_{i \in \{1, ..., n\}} u_i = 0 \Leftrightarrow x_i \notin X_j$. Thus, RDG is executed recursively, dividing subsequent groups of decision variables by half until a one-element group is reached or the separability is discovered.
	
	RDG3~\cite{rdg3}, introduces two new thresholds, i.e., $\epsilon_s$ and $\epsilon_n$. It tries to join all separable variables into groups of size $\epsilon_s$. The $\epsilon_n$ threshold is useful for overlapping problems. The variables of such problems are divided into groups of approximately $\epsilon_n$ size instead of creating one group. CC frameworks embedding RDG3 were shown more effective than those using its predecessors~\cite{rdg3}.
	
	\subsubsection{Monotonicity Checking}
	
	Monotonicity checking strategies~\cite{lvi, ccvil, svil} do not assume that a problem is additively separable. Two decision variables $x_p$ and $x_q$ are interacting if $\exists a_1 \ne a_2, b_1 \ne b_2, \vec{x^*} \in \Omega$ such that
	\begin{equation}
	\small
	\label{eq:nmd:nonseparability}
	\begin{aligned}
	& f(\vec{x^*})|_{x_p=a_1,x_q=b_1} \le f(\vec{x^*})|_{x_p=a_2,x_q=b_1} \land \\
	& f(\vec{x^*})|_{x_p=a_1,x_q=b_2} > f(\vec{x^*})|_{x_p=a_2,x_q=b_2}
	\end{aligned}
	\end{equation}
	
	Fast variable interdependence learning~(FVIL)~\cite{fvil} replaces this pair-wise interaction check by examining two disjoint groups of decision variables $X_1$ and $X_2$. They are interacting if $\exists \delta_1, \delta_2 > 0$, $\vec{u_1} \in U_{X_1}$, $\vec{u_2} \in U_{X_2}$, $\vec{x^*} \in \Omega$ such that
	\begin{equation}
	\small
	\label{eq:nmd:nonseparability:r}
	f(\vec{x^*}) \le f(\vec{x^*} + \delta_1 \vec{u_1}) \land f(\vec{x^*} + \delta_2 \vec{u_2}) > f(\vec{x^*} + \delta_1 \vec{u_1} + \delta_2 \vec{u_2})
	\end{equation}
	To check if $X_1$ and $X_2$ interact, randomly created $\delta_1$, $\delta_2$, $\vec{u_1}$, $\vec{u_2}$, and $\vec{x^*}$ are used. This procedure is repeated at most $N$ times. As in RDG, subsequent groups of variables are being recursively divided. Thus, the complexity is also $\mathcal{O}(n\log(n))$.
	
	\subsection{Cooperative Co-Evolution and Hybrid Optimization}
	\label{sec:cc}
	
	The idea behind CC is to optimize each component (subproblem) separately. Thus, the decomposition strategies are useful for CC~\cite{cc}. The decomposition quality is not the only factor that influences the effectiveness of CC\textemdash the lower quality decomposition may lead to better results~\cite{rdg3}. In the original CC, the expenses for optimizing each component are similar. Since different subproblems may have a different (high or low) impact on global fitness improvements~\cite{cbcc}, it seems reasonable to concentrate the optimization on those components that will have the highest impact on fitness.
	
	Contribution-based CC~(CBCC)~\cite{cbcc, cc2018} computes an accumulated contribution $\Delta F_i$ of the $i$th component considering all previous fitness improvements. Improvements found earlier have a lower impact on the accumulated contribution. CBCC does not require components of equal sizes. A subpopulation of a different size may optimize each component, but the FFE budget of the single run is the same for each component. In CCFR2 (another extended CC framework)~\cite{ccfr2}, each subpopulation (that may be of a different size as in CBCC) is executed every time for the same number of iterations. Additionally, in CCFR2, the accumulated contribution considers the computational cost of finding the improvement.
	
	In CBCC and CCFR2, we provide them the decomposition of a considered problem and execute separate optimization processes for separate components. In hybrid methods, we search for the most appropriate optimizer for optimizing all variables together or their random subset~\cite{shade-ils,mosINS,mlshade-spa}. SHADE-ILS~\cite{shade-ils} hybridizes SHADE~\cite{shade} and two local search methods. SHADE is used to explore the search space, while local search is used to improve the quality of promising solutions. SHADE-ILS employs MTS-L1~\cite{mts} and L-BFGS-B~\cite{L-BFGS-B} that have complementary pros and conns. MTS-L1 is effective for separable problems, while L-BFGS-B approximates the gradient and is more robust. Additionally, SHADE-ILS restarts itself when the improvement ratio is low.
	
	\section{Differential Grouping and Monotonicity Checking\textemdash Analysis of Main Differences}
	\label{sec:comp}
	
	This section presents a comparison between differential grouping and monotonicity checking. We focus on functions for which at least one of these two strategies reports false or missing linkage. To this end, we adapt the definition of the variable separability from~\cite{3lo} to continuous domains. For the sake of clarity, we consider a two variable function $\example{f}$ and two univariate subfunctions that simplify $\example{f}$ by replacing the second variable by a constant value: $\example{g}(x) = \example{f}(\vec{x})|_{x_1=x,x_2=b_1}$ and $\example{h}(x) = \example{f}(\vec{x})|_{x_1=x,x_2=b_2}$, where $b_1$ and $b_2$ are two different constant values. Conclusions drawn for a two variable function can be easily generalized to more variables.
	
	\subsection{Monotonicity Checking as Empirical Linkage Learning}
	\label{sec:mc:ell}
	
\begin{figure}
		\centering
		\subfloat[Possibly separable]{
			\resizebox{0.47\linewidth}{!}{%
				\begin{tikzpicture}
				\begin{axis}[%
				extra x ticks={0},
				extra x tick labels=\empty,
				extra x tick style={grid=major,major grid style={draw=black, dotted}},
				yticklabels=\empty,
				xticklabels=\empty,
				xmin=-5,
				xmax=5,
				ymin=-1.5,
				ymax=-0.5,
				legend entries={$\boldsymbol{\example{g}(x)}$, $\boldsymbol{\example{h}(x)}$},
				legend columns = -1,
				legend style={font=\fontsize{13}{0}\selectfont},
				legend pos = north east,
				label style={inner sep=0pt}, 
				tick label style={inner sep=0pt}, 
				]
				\addplot[color=black,samples=500]{-1.1*e^(-(x-4.11)^2 / 3) -1.3*e^(-(x+0)^2 / 5) -0.9*e^(-(x+4.11)^2 / 3) + 0.13};
				\addplot[color=black,samples=500,dashed]{-0.95*e^(-(x-4.11)^2 / 3) -1.3*e^(-(x+0)^2 / 5) -1.15*e^(-(x+4.11)^2 / 3) - 0.12};
				\draw [-latex, line width=1.5pt](-4,-0.82) -- (-4,-1.32);
				\draw [-latex, line width=1.5pt](4,-1.02) -- (4,-1.12);
				\node at (-3, -1.4) {\large $\boldsymbol{\bar{3^{rd}}} \rightarrow \boldsymbol{\ddot{2^{nd}}}$};
				\node at (3, -1.2) {\large $\boldsymbol{\bar{2^{nd}}} \rightarrow \boldsymbol{\ddot{3^{rd}}}$};
				\end{axis}
				\end{tikzpicture}
			}
			\label{fig:decompositionSeparability:a}
		}
		\hfil
		\subfloat[Non-separable]{
			\resizebox{0.47\linewidth}{!}{%
				\begin{tikzpicture}
				\begin{axis}[%
				extra x ticks={0,4},
				extra x tick labels=\empty,
				extra x tick style={grid=major,major grid style={draw=black, dotted}},
				yticklabels=\empty,
				xticklabels=\empty,
				xmin=-5,
				xmax=5,
				ymin=-1.5,
				ymax=-0.5,
				legend entries={$\boldsymbol{\example{g}(x)}$, $\boldsymbol{\example{h}(x)}$},
				legend columns = -1,
				legend style={font=\fontsize{13}{0}\selectfont},
				legend pos = north east,
				label style={inner sep=0pt}, 
				tick label style={inner sep=0pt}, 
				]
				\addplot[color=black,samples=500]{-1.1*e^(-(x-4.11)^2 / 3) -1.3*e^(-(x+0)^2 / 5) -0.9*e^(-(x+4.11)^2 / 3) + 0.13};
				\addplot[color=black,samples=500,dashed]{-1.2*e^(-(x-4.11)^2 / 3) -1.11*e^(-(x+0)^2 / 5) -0.9*e^(-(x+4.11)^2 / 3) - 0.19};
				\draw [-latex, line width=1.5pt](4,-1.02) -- (4,-1.43);
				\draw [-latex, line width=1.5pt](0,-1.18) -- (0,-1.31);
				\node at (-2, -1.31) {\large $\boldsymbol{\bar{1^{st}}} \rightarrow \boldsymbol{\ddot{2^{nd}}}$};
				\node at (2, -1.43) {\large $\boldsymbol{\bar{2^{nd}}} \rightarrow \boldsymbol{\ddot{1^{st}}}$};
				\end{axis}
				\end{tikzpicture}
			}
			\label{fig:decompositionSeparability:b}
		}
		\hfil
		\\[4pt]
		\subfloat[Possibly separable]{
			\resizebox{0.47\linewidth}{!}{%
				\begin{tikzpicture}
				\begin{axis}[%
				extra x ticks={0},
				extra x tick labels=\empty,
				extra x tick style={grid=major,major grid style={draw=black, dotted}},
				yticklabels=\empty,
				xticklabels=\empty,
				xmin=-5,
				xmax=5,
				ymin=-1.5,
				ymax=-0.5,
				legend entries={$\boldsymbol{\example{g}(x)}$, $\boldsymbol{\example{h}(x)}$},
				legend columns = -1,
				legend style={font=\fontsize{13}{0}\selectfont},
				legend pos = north east,
				label style={inner sep=0pt}, 
				tick label style={inner sep=0pt}, 
				]
				\addplot[color=black,samples=500]{-1.1*e^(-(x-4.11)^2 / 3) -1.3*e^(-(x+0)^2 / 5) -0.9*e^(-(x+4.11)^2 / 3) + 0.13};
				\addplot[color=black,dashed]{(x/4)^2 - 1.42};
				\end{axis}
				\end{tikzpicture}
			}
			\label{fig:decompositionSeparability:c}
		}
		\hfil
		\subfloat[Non-separable]{
			\resizebox{0.47\linewidth}{!}{%
				\begin{tikzpicture}
				\begin{axis}[%
				extra x ticks={0,4},
				extra x tick labels=\empty,
				extra x tick style={grid=major,major grid style={draw=black, dotted}},
				yticklabels=\empty,
				xticklabels=\empty,
				xmin=-5,
				xmax=5,
				ymin=-1.5,
				ymax=-0.5,
				legend entries={$\boldsymbol{\example{g}(x)}$, $\boldsymbol{\example{h}(x)}$},
				legend columns = -1,
				legend style={font=\fontsize{13}{0}\selectfont},
				legend pos = north east,
				label style={inner sep=0pt}, 
				tick label style={inner sep=0pt}, 
				]
				\addplot[color=black,samples=500]{-1.1*e^(-(x-4.11)^2 / 3) -1.3*e^(-(x+0)^2 / 5) -0.9*e^(-(x+4.11)^2 / 3) + 0.13};
				\addplot[color=black,dashed]{((x-4)/4)^2 - 1.42};
				\end{axis}
				\end{tikzpicture}
			}
			\label{fig:decompositionSeparability:d}
		}
		\caption{Separability according to formula~(\ref{eq:decomposition})}
		\label{fig:decompositionSeparability}
	\end{figure}
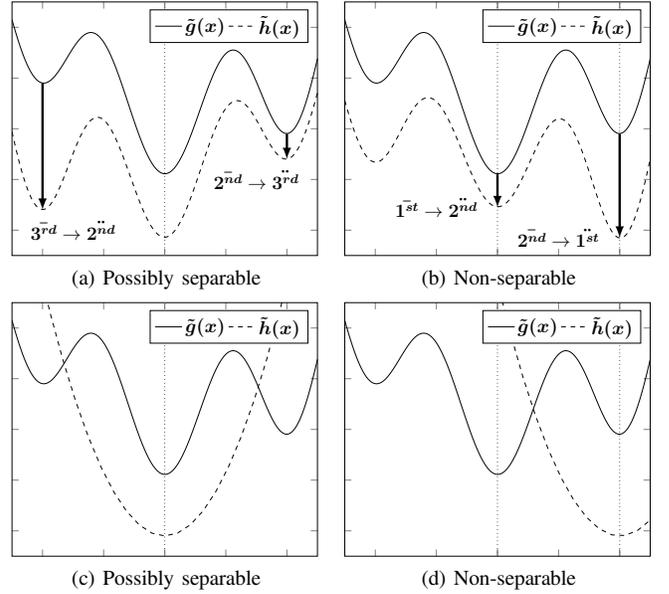

	According to formula~(\ref{eq:decomposition}), and taking subfunctions $\example{g}$ and $\example{h}$ into account, we can state that if $\argmin_x \example{g}(x) \ne \argmin_x \example{h}(x)$, then $x_1$ and $x_2$ are interacting. Otherwise, although the global optimum is not changed, it is not certain that $x_1$ and $x_2$ are separable, because only two different values of $x_2$, i.e., $b_1$ and $b_2$, have been considered. After setting $x_2$ to $b_3$ and $b_4$, such that $b_3 \ne b_1 \land b_4 \ne b_2$, the condition $\argmin_x \example{g}(x) \ne \argmin_x \example{h}(x)$ could be met. This kind of interaction condition can omit fitness landscape characteristics~\cite{flc}, which may significantly affect the optimizer effectiveness~\cite{flcPSO,fla}. Fig.~\ref{fig:decompositionSeparability:a} and~\ref{fig:decompositionSeparability:b} present cases when one of optima becomes better than another. However, only if a global one changes, the variables are found interacting. Similarly, if a subfunction  transforms from multi- into unimodal for different values of $x_2$, as in Fig.~\ref{fig:decompositionSeparability:c} and~\ref{fig:decompositionSeparability:d}, the dependency will be discovered only when a global optimum is affected.
	
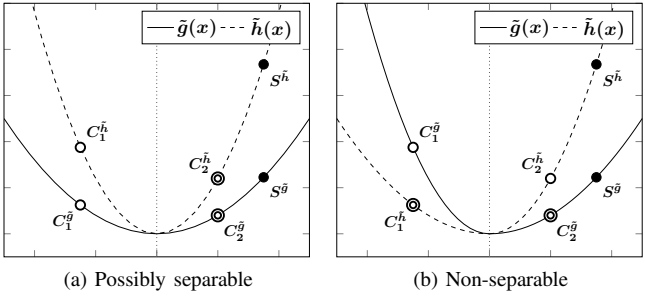
\begin{figure}
		\centering
		\subfloat[Possibly separable]{
			\resizebox{0.47\linewidth}{!}{%
				\begin{tikzpicture}
				\begin{axis}[%
				extra x ticks={0}, 
				extra x tick labels=\empty,
				extra x tick style={grid=major,major grid style={draw=black, dotted}},
				yticklabels=\empty,
				xticklabels=\empty,
				xmin=-5,
				xmax=5,
				ymin=-5,
				ymax=50,
				legend entries={$\boldsymbol{\example{g}(x)}$, $\boldsymbol{\example{h}(x)}$},
				legend columns = -1,
				legend style={font=\fontsize{13}{0}\selectfont},
				legend pos = north east,
				label style={inner sep=0pt}, 
				tick label style={inner sep=0pt}, 
				]
				\addplot[color=black]{x^2)};
				\addplot[color=black, dashed]{3*x^2)};
				\filldraw[black] (3.5, 12.25) circle (3pt) node[anchor=north west, outer sep=0.5pt] {$\boldsymbol{S^{\example{g}}}$};
				\filldraw[color=black, fill=white, very thick, accepting] (2,4) circle (3pt) node[anchor=north west, outer sep=0.5pt] {$\boldsymbol{C^{\example{g}}_2}$};
				\filldraw[color=black, fill=white, very thick] (-2.5,6.25) circle (3pt) node[anchor=north east, outer sep=0.5pt] {$\boldsymbol{C^{\example{g}}_1}$};
				\filldraw[black] (3.5, 36.75) circle (3pt) node[anchor=north west, outer sep=0.5pt] {$\boldsymbol{S^{\example{h}}}$};
				\filldraw[color=black, fill=white, very thick, accepting] (2,12) circle (3pt) node[anchor=south east, outer sep=0.5pt] {$\boldsymbol{C^{\example{h}}_2}$};
				\filldraw[color=black, fill=white, very thick] (-2.5,18.75) circle (3pt) node[anchor=south west, outer sep=0.5pt] {$\boldsymbol{C^{\example{h}}_1}$};
				\end{axis}
				\end{tikzpicture}
			}
			\label{fig:greedyAndSeparability:a}
		}
		\hfil
		\subfloat[Non-separable]{
			\resizebox{0.47\linewidth}{!}{%
				\begin{tikzpicture}
				\begin{axis}[%
				extra x ticks={0}, 
				extra x tick labels=\empty,
				extra x tick style={grid=major,major grid style={draw=black, dotted}},
				yticklabels=\empty,
				xticklabels=\empty,
				xmin=-5,
				xmax=5,
				ymin=-5,
				ymax=50,
				legend entries={$\boldsymbol{\example{g}(x)}$, $\boldsymbol{\example{h}(x)}$},
				legend columns = -1,
				legend style={font=\fontsize{13}{0}\selectfont},
				legend pos = north east,
				label style={inner sep=0pt}, 
				tick label style={inner sep=0pt}, 
				]
				\addplot[color=black, domain=\pgfkeysvalueof{/pgfplots/xmin}:0]{3*x^2)};
				\addplot[color=black, domain=0:\pgfkeysvalueof{/pgfplots/xmax}, forget plot]{x^2)};
				\addplot[color=black, dashed, domain=\pgfkeysvalueof{/pgfplots/xmin}:0]{x^2)};
				\addplot[color=black, dashed, domain=0:\pgfkeysvalueof{/pgfplots/xmax}, forget plot]{3*x^2)};
				\filldraw[black] (3.5, 12.25) circle (3pt) node[anchor=north west, outer sep=0.5pt] {$\boldsymbol{S^{\example{g}}}$};
				\filldraw[color=black, fill=white, very thick, accepting] (2,4) circle (3pt) node[anchor=north west, outer sep=0.5pt] {$\boldsymbol{C^{\example{g}}_2}$};
				\filldraw[color=black, fill=white, very thick, accepting] (-2.5,6.25) circle (3pt) node[anchor=north east, outer sep=0.5pt] {$\boldsymbol{C^{\example{h}}_1}$};
				\filldraw[black] (3.5, 36.75) circle (3pt) node[anchor=north west, outer sep=0.5pt] {$\boldsymbol{S^{\example{h}}}$};
				\filldraw[color=black, fill=white, very thick] (2,12) circle (3pt) node[anchor=south east, outer sep=0.5pt] {$\boldsymbol{C^{\example{h}}_2}$};
				\filldraw[color=black, fill=white, very thick] (-2.5,18.75) circle (3pt) node[anchor=south west, outer sep=0.5pt] {$\boldsymbol{C^{\example{g}}_1}$};
				\end{axis}
				\end{tikzpicture}
			}
			\label{fig:greedyAndSeparability:b}
		}
		\caption{Separability in terms of greedy optimizer behavior}
		\label{fig:greedyAndSeparability}
	\end{figure}

	The provided examples show that even if two variables are possibly separable (according to formula~(\ref{eq:decomposition})), an optimizer may behave differently depending on which subfunctions will be taken into account. Thus, the condition comparing only global optimum changes to distinguish between the separability and the non-separability seems insufficient. In the example presented in Fig.~\ref{fig:greedyAndSeparability}, we consider a greedy optimizer that starts its search in point $S^{\example{g}}$ or $S^{\example{h}}$ for subfunctions $\example{g}$ and $\example{h}$, respectively. Since $\example{g}$ and $\example{h}$ have the same global optimum, then $x_1$ and $x_2$ are possibly separable according to formula (\ref{eq:decomposition}). Let us consider the following candidate solutions, $C_1^{\example{g}}$, $C_2^{\example{g}}$ for subfunction $\example{g}$, and $C_1^{\example{h}}$, $C_2^{\example{h}}$ for subfunction $\example{h}$. Double circles indicate the better candidate solution for $\example{g}$ and $\example{h}$. In Fig.~\ref{fig:greedyAndSeparability:a}, the choice is the same for $\example{g}$ and $\example{h}$. Thus, $x_1$ and $x_2$ may be, indeed, separable. In Fig.~\ref{fig:greedyAndSeparability:b} the choice is different for each subfunction. The optimization process of $\example{g}$ and $\example{h}$ may result in different points in the search space. Therefore, it seems reasonable to state that the variables $x_1$ and $x_2$ are interacting. For more complex subfunctions, an optimizer may be deceived to a local optimum only for one subfunction, although both have the same global optimum. The selection pressure favors better fitted individuals. Thus, a single change in the outcome of the individuals' fitness comparison may influence a direction of the search~\cite{selection}.
	
	The above remarks are consistent with the idea of empirical linkage learning~(ELL)~\cite{3lo}, a new class of linkage learning techniques. In ELL, interactions are discovered by comparing local search results before and after perturbing one of the decision variables. Two variables interact if a local search method converges to a different value after perturbing one of them. Such techniques are proven never to report false linkage. The formal proof assumes the following definition of the variable separability~\cite{3lo}. Two disjoint sets of variables $X_1$ and $X_2$ are independent if for each values $\delta_1$, $\delta_2 > 0$, unit vectors $\vec{u_1} \in U_{X_1}$, $\vec{u_2} \in U_{X_2}$, and decision vector $\vec{x^*} \in \Omega$ the following condition holds: $f(\vec{x^*}) < f(\vec{x^*} + \delta_1 \vec{u_1}) \Leftrightarrow f(\vec{x^*} + \delta_2 \vec{u_2}) < f(\vec{x^*} + \delta_1 \vec{u_1} + \delta_2 \vec{u_2})$ and $f(\vec{x^*}) = f(\vec{x^*} + \delta_1 \vec{u_1}) \Leftrightarrow f(\vec{x^*} + \delta_2 \vec{u_2}) = f(\vec{x^*} + \delta_1 \vec{u_1} + \delta_2 \vec{u_2})$ on the premise that all arguments of $f$ are feasible solutions. In the latter part of this paper, unless stated otherwise, we consider the above (ELL-based) definition of separability. Thus, if exists such $\delta_1$, $\delta_2$, $\vec{u_1}$, $\vec{u_2}$, and $\vec{x^*}$ that $f(\vec{x^*}) \le f(\vec{x^*} + \delta_1 \vec{u_1})$ and $f(\vec{x^*} + \delta_2 \vec{u_2}) > f(\vec{x^*} + \delta_1 \vec{u_1} + \delta_2 \vec{u_2})$ (formula~(\ref{eq:nmd:nonseparability:r})), then $X_1$ and $X_2$ sets are interacting. Thus, monotonicity checking strategies may be classified as ELL techniques. In~\cite{ccvil}, they are also identified as strategies that do not report false linkage.
	
	\subsection{Decomposition Inaccuracies: False and Missing Linkage}
	\label{sec:comp:falseOrMissingLinkage}
	
	DG-based decomposition strategies are proven that they will never report false linkage if a function is additively separable~\cite{dg,rdg}. A function is additively separable if every pair of its subfunctions is also additively separable. Let $f_i$ and $f_j$ be separable subfunctions. They are additively separable if $\forall_{x_p, x_q} \frac{\partial^2 f}{\partial x_q \partial x_p} = 0$, where $x_p$ and $x_q$ are arguments of subfunctions $f_i$ and $f_j$, respectively~\cite{rdg}. Otherwise, $f_i$ and $f_j$ are non-additively separable\footnote{See the supplementary material for examples and additional explanations.}. If a function is non-additively separable, i.e., every pair of its subfunctions is also non-additively separable, DG-based decomposition strategies may report false linkage. Let us consider the minimization of the following function: $\bar{f}_{c,1}: [-5, 5]^2 \rightarrow [0, 100]$ defined as $\bar{f}_{c,1}(\vec{x}) = (|x_1| + |x_2|)^2$. A greedy optimizer that is deterministic (e.g., with a fixed value of the random seed) applied to optimize $x_1$, for any value of $x_2$, will converge to the same value and vice versa. If this greedy optimizer has infinite precision, it will converge to 0 for both variables. Therefore, $\bar{f}_{c,1}$ is fully separable. However, DG-based strategies will find $x_1$ and $x_2$ dependent, because using formula~(\ref{eq:deltaCheck}), we get $\Delta_1 = 5$ and $\Delta_2 = 7$ for $(a, a + \delta, b_1, b_2) = (1, 2, 1, 2)$.
	
	On the other hand, monotonicity checking never reports false linkage (see Section~\ref{sec:mc:ell}). However, in practice, due to the inaccuracy of the representation of float numbers, the appropriate value of $\epsilon$ is necessary to determine if the inequalities are satisfied (see formula~(\ref{eq:nmd:nonseparability})). Then, monotonicity checking strategies will not report false linkage also for problems with non-additively separable subproblems. DG-based strategies do not have this advantage (and also suffer from the floating-point error). However, the results presented in~\cite{dg, dg2, rdg, fvil} show that DG-based strategies are more accurate. The reason may be that monotonicity checking strategies proposed so far are more vulnerable to missing linkage. Let us consider more examples that employ function $\example{f}$, and its subfunctions $\example{g}$ and $\example{h}$, defined at the beginning of this subsection. We simplify formula~(\ref{eq:deltaCheck}) to $\example{g}(a + \delta) - \example{g}(a) \ne \example{h}(a + \delta) - \example{h}(a)$ that is equivalent to $\Delta_1 \ne \Delta_2$. Similarly, we simplify formula~(\ref{eq:deltaCheck:delta}) to $\example{g}(a_1) \le \example{g}(a_2) \land \example{h}(a_1) > \example{h}(a_2)$.
	
	Fig.~\ref{fig:falseLinkage} presents another example in which DG may report false linkage. We multiply a function by a positive number, which does not change relations between its outputs, i.e., the relation between function outputs (smaller, greater, equal) for any two arguments will be the same for the original and multiplied functions. In Fig.~\ref{fig:falseLinkage:a}, $\Delta_1$ and $\Delta_2$ are indicated by vectors for various values of $a$ spread along the horizontal axis and the same value of $\delta$. For at least one pair ($\Delta_1$, $\Delta_2$) the referring vectors differ. Thus, although, the considered problem is separable, DG may detect (false) interaction between $x_1$ and $x_2$. Oppositely, monotonicity checking will consider the monotonicity intervals (Fig.~\ref{fig:falseLinkage:b}), which are the same for $\example{g}$ and $\example{h}$. Note that it is a necessary, but not sufficient, condition. Hence, we can use another condition, this time sufficient. If $\example{h}(x) = \alpha \cdot \example{g}(x)$, where $\alpha > 0$, then $\forall_{a_1 \ne a_1} \example{g}(a_1) > \example{g}(a_2) \lor \example{h}(a_1) \le \example{h}(a_2)$, because $\forall_{\alpha > 0} \example{h}(a_1) \le \example{h}(a_2) \Leftrightarrow \example{g}(a_1) \le \example{g}(a_2)$. Therefore, monotonicity checking will report $x_1$ and $x_2$ as independent.
	
	\begin{figure}
		\centering
		\subfloat[Differential grouping (false linkage)]{
			\resizebox{0.47\linewidth}{!}{%
				\begin{tikzpicture}
				\begin{axis}[%
				extra x ticks={0}, 
				extra x tick labels=\empty,
				extra x tick style={grid=major,major grid style={draw=black, dotted}},
				yticklabels=\empty,
				xticklabels=\empty,
				xmin=-3.9,
				xmax=3.9,
				ymin=-5,
				ymax=50,
				legend entries={$\boldsymbol{\example{g}(x)}$, $\boldsymbol{\example{h}(x)}$},
				legend columns = -1,
				legend style={font=\fontsize{13}{0}\selectfont},
				legend pos = north east,
				label style={inner sep=0pt}, 
				tick label style={inner sep=0pt}, 
				]
				\addplot[color=black]{x^2)};
				\addplot[color=black, dashed]{3*x^2)};
				\draw [-latex, line width=1.5pt](-3,9) -- (-3,4);
				\draw [-latex, line width=1.5pt, dashed](-3,27) -- (-3,12);
				\draw [-latex, line width=1.5pt](2.5,6.25) -- (2.5,12.25);
				\draw [-latex, line width=1.5pt](2,4) -- (2,9);
				\draw [-latex, line width=1.5pt](1.5,2.25) -- (1.5,6.25);
				\draw [-latex, line width=1.5pt, dashed](2.5,18.75) -- (2.5,36.75);
				\draw [-latex, line width=1.5pt, dashed](2,12) -- (2,27);
				\draw [-latex, line width=1.5pt, dashed](1.5,6.75) -- (1.5,18.75);
				\node at (-3.45, 8) {\large $\vec{\Delta_1}$};
				\node at (-3.45, 20) {\large $\vec{\Delta_2}$};
				\draw [latex-latex, line width=1.5pt](2.5,12.25) -- (3.5,12.25);
				\draw [latex-latex, line width=1.5pt, dashed](2.5,36.75) -- (3.5,36.75);
				\node at (3, 14.5) {\large $\boldsymbol{\delta}$};
				\node at (3, 39) {\large $\boldsymbol{\delta}$};
				\end{axis}
				\end{tikzpicture}
			}
			\label{fig:falseLinkage:a}
		}
		\hfil
		\subfloat[Monotonicity checking (no false linkage)]{
			\resizebox{0.47\linewidth}{!}{%
				\begin{tikzpicture}
				\begin{axis}[%
				extra x ticks={0}, 
				extra x tick labels=\empty,
				extra x tick style={grid=major,major grid style={draw=black, dotted}},
				yticklabels=\empty,
				xticklabels=\empty,
				xmin=-3.9,
				xmax=3.9,
				ymin=-5,
				ymax=50,
				legend entries={$\boldsymbol{\example{g}(x)}$, $\boldsymbol{\example{h}(x)}$},
				legend columns = -1,
				legend style={font=\fontsize{13}{0}\selectfont},
				legend pos = north east,
				label style={inner sep=0pt}, 
				tick label style={inner sep=0pt}, 
				]
				\addplot[color=black]{x^2)};
				\addplot[color=black, dashed]{3*x^2)};
				\draw [-latex, line width=1.5pt](-2,35) -- (-1,30);
				\draw [-latex, line width=1.5pt, dashed](-2,27) -- (-1,22);
				\draw [-latex, line width=1.5pt](1,30) -- (2,35);
				\draw [-latex, line width=1.5pt, dashed](1,22) -- (2,27);
				\end{axis}
				\end{tikzpicture}
			}
			\label{fig:falseLinkage:b}
		}
		\caption{Influence of function scaling on false linkage discovery}
		\label{fig:falseLinkage}
	\end{figure}
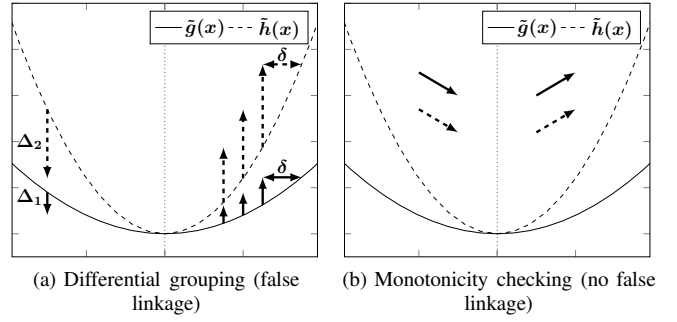

	\begin{figure}
		\centering
		\subfloat[Differential grouping (high sensitivity)]{
			\resizebox{0.47\linewidth}{!}{%
				\begin{tikzpicture}
				\begin{axis}[%
				yticklabels=\empty,
				xticklabels=\empty,
				xmin=-5,
				xmax=5,
				ymin=-5,
				ymax=50,
				legend entries={$\boldsymbol{\example{g}(x)}$, $\boldsymbol{\example{h}(x)}$},
				legend columns = -1,
				legend style={font=\fontsize{13}{0}\selectfont},
				legend pos = north east,
				label style={inner sep=0pt}, 
				tick label style={inner sep=0pt}, 
				]
				\addplot[color=black]{2*(x + 1.5)^2)};
				\addplot[color=black,dashed]{2*(x - 1.5)^2)};
				\draw [-latex, line width=1.5pt](-3,4.5) -- (-3,0);
				\draw [-latex, line width=1.5pt](-3.5,8) -- (-3.5,0.5);
				\draw [-latex, line width=1.5pt](-1.5,0) -- (-1.5,4.5);
				\draw [-latex, line width=1.5pt](1.5,18) -- (1.5,40.5);
				\draw [latex-latex, line width=1.5pt](-1.5,4.5) -- (0,4.5);
				\draw [-latex, line width=1.5pt, dashed](-3,40.5) -- (-3,18);
				\draw [-latex, line width=1.5pt, dashed](-3.5,50) -- (-3.5,24.5);
				\draw [-latex, line width=1.5pt, dashed](-1.5,18) -- (-1.5,4.5);
				\draw [-latex, line width=1.5pt, dashed](1.5,0) -- (1.5,4.5);
				\draw [latex-latex, line width=1.5pt, dashed](1.5,4.5) -- (3,4.5);
				\node at (0.8, 29) {\large $\vec{\Delta_1}$};
				\node at (-4.2, 37) {\large $\vec{\Delta_2}$};
				\node at (2.25, 7) {\large $\boldsymbol{\delta}$};
				\node at (-0.75, 7) {\large $\boldsymbol{\delta}$};
				\end{axis}
				\end{tikzpicture}
			}
			\label{fig:sensitivity:a}
		}
		\hfil
		\subfloat[Monotonicity checking (low sensitivity)]{
			\resizebox{0.47\linewidth}{!}{%
				\begin{tikzpicture}
				\begin{axis}[%
				extra x ticks={-1.5, 1.5}, 
				extra x tick labels=\empty,
				extra x tick style={grid=major,major grid style={draw=black, dotted}},
				yticklabels=\empty,
				xticklabels=\empty,
				xmin=-5,
				xmax=5,
				ymin=-5,
				ymax=50,
				legend entries={$\boldsymbol{\example{g}(x)}$, $\boldsymbol{\example{h}(x)}$},
				legend columns = -1,
				legend style={font=\fontsize{13}{0}\selectfont},
				legend pos = north east,
				label style={inner sep=0pt}, 
				tick label style={inner sep=0pt}, 
				]
				\addplot[color=black]{2*(x + 1.5)^2)};
				\addplot[color=black,dashed]{2*(x - 1.5)^2)};
				\draw [-latex, line width=1.5pt](-4.0,35) -- (-3.0,30);
				\draw [-latex, line width=1.5pt, dashed](-4.0,27) -- (-3.0,22);
				\draw [-latex, line width=1.5pt](-0.5,30) -- (0.5,35);
				\draw [-latex, line width=1.5pt, dashed](-0.5,27) -- (0.5,22);
				\draw [-latex, line width=1.5pt](3.0,30) -- (4.0,35);
				\draw [-latex, line width=1.5pt, dashed](3.0,22) -- (4.0,27);
				\end{axis}
				\end{tikzpicture}
			}
			\label{fig:sensitivity:b}
		}
		\caption{Sensitivity to interaction discovery}
		\label{fig:sensitivity}
	\end{figure}
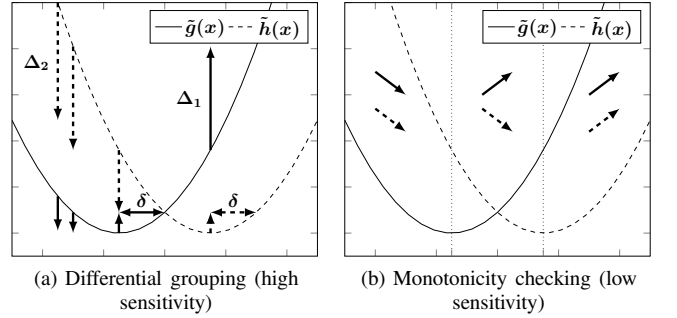
	
	One of the main advantages of DG is its high resistance to missing linkage. In Fig.~\ref{fig:sensitivity:a}, we show that the probability of selecting $a$ that leads to the dependency discovery is high. Monotonicity checking will always report $x_1$ and $x_2$ as dependent only if $a_1$ and $a_2$ come from the middle interval (Fig.~\ref{fig:sensitivity:b}). Otherwise, the interaction may not be discovered.
	
	Feasible values of decision variables may also influence monotonicity checking. Let us consider the decomposition of $\bar{f}_{c,2}: [-8, 8] \times [-2, 2] \rightarrow [0, 100]$ defined as $\bar{f}_{c,2}(\vec{x})=(x_1+x_2)^2$. For $(a_1, a_2, b_1, b_2) = (-2, -1, 2, 1)$, monotonicity checking will discover the dependency between $x_1$ and $x_2$.However, if $a_1 = -6$ and $a_2 = -5$, then $\forall_{b_1, b_2 \in [-2,2]} \bar{f}_{c,2}(a_1,b_1) >  \bar{f}_{c,2}(a_2,b_1) \land \bar{f}_{c,2}(a_1,b_2) >  \bar{f}_{c,2}(a_2,b_2)$. Thus, monotonicity checking will not find the dependency for any feasible values of $b_1$ and $b_2$. Note that $\bar{f}_{c,2}$ is fully non-separable, because the value of $\bar{f}_{c,2}$ is minimal when $x_1 = -x_2$. Finally, DG-based strategies should discover the interaction between $x_1$ and $x_2$ easily.
	
	\section{Incremental Recursive Ranking Grouping}
	\label{sec:irrg}
	
	Decomposition strategies based on DG and monotonicity checking may report false or missing linkage, respectively. Therefore, we propose Incremental Recursive Ranking Grouping~(IRRG) employing the idea of monotonicity checking. IRRG never reports false linkage and significantly mitigates the issue of missing linkage. First, we describe Recursive Ranking Grouping~(RRG) that is a key-part part of our proposition, then we present the general view of IRRG. Finally, we compare IRRG with other monotonicity checking strategies and discuss the time complexity of IRRG.
	
	\subsection{Basic Concept}
	\label{sec:irrg:basic}
	
	The simplest way to decrease the level of missing linkage in monotonicity checking is to create uniformly and randomly a huge number of samples. A single sample is defined by $a_1$, $a_2$, $b_1$, $b_2$, $\vec{x^*}$ and by $\delta_1$, $\delta_2$, $\vec{u_1}$, $\vec{u_2}$, $\vec{x^*}$ for formulas~(\ref{eq:nmd:nonseparability}) and~(\ref{eq:nmd:nonseparability:r}), respectively. Such an approach increases the probability that at least one sample will result in dependency discovery if two variables or two sets of variables indeed interact. However, its cost may not be reasonable for the optimization process.
	
	To increase the sensitivity of monotonicity checking, IRRG creates two rankings of samples. Each ranking consists of $n_s$ (a user-defined parameter) samples. To check if variables $x_p$ and $x_q$ interact, we need $b_1$, $b_2$ ($b_1 \neq b_2$), $\vec{x^*}$, and $n_s$ values that are evenly taken from the feasible set of $x_p$. Let us denote these $n_s$ values as $\example{a}_1$, ..., $\example{a}_{n_s}$. Then, we create two rankings $\vec{r_1} = [r_{1,1}, ..., r_{1,n_s}]$ and $\vec{r_2} = [r_{2,1}, ..., r_{2,n_s}]$, where $r_{j,i}$ is computed for $f(\vec{x^*})|_{x_p=\example{a}_i,x_q=b_j}$. If $\exists_{i \in \{1, ..., n_s\}} r_{1,i} \ne r_{2,i}$ then the variables $x_p$ and $x_q$ interact. For instance, for $\bar{f}_{c,3}: [-3, 3]^4 \rightarrow [-111, 111]$ defined as $\bar{f}_{c,3}(\vec{x})=(x_1+x_2)^2 \cdot x_3+x_4$, $p=1$, $q=2$, $b_1=1$, $b_2=2$, $\vec{x^*} = [1,0,3,2]$, and $n_s=3$, we get $\example{a}_1=-3$, $\example{a}_2=0$, and $\example{a}_3=3$. Thus, $\bar{f}_{c,3}(\vec{x^*})|_{x_1=-3,x_2=1}=14$, $\bar{f}_{c,3}(\vec{x^*})|_{x_1=0,x_2=1}=5$, $\bar{f}_{c,3}(\vec{x^*})|_{x_1=3,x_2=1}=50$, and $\bar{f}_{c,3}(\vec{x^*})|_{x_1=-3,x_2=2}=5$, $\bar{f}_{c,3}(\vec{x^*})|_{x_1=0,x_2=2}=14$, $\bar{f}_{c,3}(\vec{x^*})|_{x_1=3,x_2=2}=77$. Therefore, $\vec{r_1} = [2, 1, 3]$ and $\vec{r_2} = [1, 2, 3]$. Since $r_{1,1}\ne r_{2,1}$, $x_1$ and $x_2$ are dependent. Comparing rankings may be found equivalent to searching for $a_1$, $a_2$, $b_1$, $b_2$, and $\vec{x^*}$ that satisfy formula~(\ref{eq:nmd:nonseparability}).
	
	Using formula~(\ref{eq:nmd:nonseparability:r}), we define the recursive ranking check that can detect interaction between two disjoint sets of variables $X_1$ and $X_2$. First, $n_s$ values for each $x_j \in X_1$, namely $\example{a}_{1,j}$, ..., $\example{a}_{n_s,j}$, are evenly generated from the $x_j$ feasible set. Then, we define $n_s$ decision vectors $\example{\vec{x}}_\vec{i} = [\example{x}_{i,1}, ..., \example{x}_{i,n}]$ based on a given vector $\vec{x^*} = [x_1^*, ..., x_n^*]$, where
	\begin{equation}
	\small
	\example{x}_{i,j} = 
	\begin{cases}
	\example{a}_{\pi_j(i),j} &, x_j \in X_1 \\
	x_j^* &, x_j \notin X_1
	\end{cases}
	\end{equation}
	where $\pi_j$ denotes a random permutation of $\{1, ..., n_s\}$ for the $j$th variable. These vectors are then used to generate the rankings. Using all $n_s^{|X_1|}$ available vectors is impossible due to practical reasons. The random permutation is used to overcome the possible bias. The $i$th value of the first ranking $r_{1,i}$ is based on $f(\example{\vec{x}}_\vec{i})$, whereas $r_{2,i}$ is computed for $f(\example{\vec{x}}_\vec{i} + \delta_2 \vec{u_2})$. Analogously to the pair-wise interaction check, If $\exists_{i^* \in \{1, ..., n_s\}} r_{1,i^*} \ne r_{2,i^*}$ then the sets $X_1$ and $X_2$ interact, because for at least one $i^* \in \{1, ..., n_s\}$, condition $f\big(\example{\vec{x}}_\vec{r_{1,i^*}}\big) \le f\big(\example{\vec{x}}_\vec{r_{2,i^*}}\big) \land  f\big(\example{\vec{x}}_\vec{r_{1,i^*}} + \delta_2 \vec{u_2}\big) > f\big(\example{\vec{x}}_\vec{r_{2,i^*}} + \delta_2 \vec{u_2}\big)$ holds. Note that this condition specifies formula~(\ref{eq:nmd:nonseparability:r}).
	
	\subsection{Recursive Ranking Grouping}
	\label{sec:irrg:rrg}
	
	RRG is the main part of IRRG. It discovers dependencies between disjoint groups of variables. First, we define a set of functions that will be used in the description of RRG. A set of groups of variables that were already found interacting will be denoted as $\mtx{G}$, while $V$ is a set of variables for which no interactions were discovered yet. In matrix $\bar{\mtx{X}}_\mathbf{1}$ of size $n_s \times n$, each $j$th column consists of $n_s$ randomly ordered values that are evenly generated from the feasible set of the $j$th variable. \par
	
	Function \textproc{ConsiderVariables}~(Pseudocode~\ref{alg:considerVariables}) decides if the variables in $V$ should be considered in the interaction check. It will happen if one of the following conditions holds. (1) $\mtx{G}$ is empty, i.e., no dependencies were discovered yet~(line~\ref{line:cv:g0v1}). (2) $V$ contains only one variable~(line~\ref{line:cv:g0v1}). (3) $V$ is randomly divided into two disjoint subsets $V_1$ and $V_2$. The sizes of both are equal or differ by one. If we detect interaction between $V_1$ and $V_2$, then the variables in $V$ will be considered in the interaction search~(lines~\ref{line:cv:v:start}--\ref{line:cv:v:end}). (4) $V$ interacts with at least one group from $\mtx{G}$ (lines~\ref{line:cv:g:start}--\ref{line:cv:g:end}). \textproc{ConsiderVariables} uses \textproc{CreateFirstRanking}~(Pseudocode~\ref{alg:createRankingR1}) that creates a ranking using the concept presented in Section~\ref{sec:irrg:basic}. Except for ranking $\vec{r_1}$, this function returns a vector of values $\bar{\vec{y}}_\vec{1}$ for which the ranking was computed. $\bar{\vec{y}}_\vec{1}$ is used by other functions defined in this section.

	\begin{algorithm}
		\caption{\textproc{ConsiderVariables}; Checking if it is worth considering separable so far variables in the interaction search}
		\scriptsize
		\begin{algorithmic}[1]
			\setstretch{1.1}
			\Statex \textbf{input:} $V$: variables to check, $\mtx{G}$: groups of non-separable variables, $\vec{x_{hq}}$: a high-quality decision vector, $\bar{\mtx{X}}_\mathbf{1}$: a $n_s \times n$ matrix of samples' values, $\bar{\vec{x}}_\vec{2}$: a decision vector different than $\vec{x_{hq}}$, $f$: an optimization problem, $n_s$: the number of samples
			\Statex \textbf{output:} a decision if at least one variable from $V$ can be non-separable
			\Function{ConsiderVariables}{$V, \mtx{G}, \vec{x_{hq}}, \bar{\mtx{X}}_\mathbf{1}, \bar{\vec{x}}_\vec{2}, f, n_s$}
			\If {$|\mtx{G}| = 0$ \textbf{or} $|V| = 1$} \label{line:cv:g0v1}
			\State \Return $true$     
			\EndIf
			\If {$|V| = 0$}
			\State \Return $false$     
			\EndIf
			\State $V \gets$ shuffle $V$ \label{line:cv:v:start}
			\State $V_1 \gets$ the first half of $V$ 
			\State $V_2 \gets$ the second half of $V$
			\State $(\bar{\vec{y}}_\vec{1}, \vec{r_1}) \gets$ \Call{CreateFirstRanking}{$V_1, \vec{x_{hq}}, \bar{\mtx{X}}_\mathbf{1}, f, n_s$}
			\If {\Call{IsInteraction}{$V_1, V_2, \vec{x_{hq}}, \bar{\mtx{X}}_\mathbf{1}, \bar{\vec{x}}_\vec{2}, \bar{\vec{y}}_\vec{1}, \vec{r_1}, f, n_s$}} \label{line:cv:v:interact1}
			\State \Return $true$ 
			\EndIf
			\State $(\bar{\vec{y}}_\vec{1}, \vec{r_1}) \gets$ \Call{CreateFirstRanking}{$V_2, \vec{x_{hq}}, \bar{\mtx{X}}_\mathbf{1}, f, n_s$}
			\If {\Call{IsInteraction}{$V_2, V_1, \vec{x_{hq}}, \bar{\mtx{X}}_\mathbf{1}, \bar{\vec{x}}_\vec{2}, \bar{\vec{y}}_\vec{1}, \vec{r_1}, f, n_s$}} \label{line:cv:v:interact2}
			\State \Return $true$ 
			\EndIf \label{line:cv:v:end}
			\For{$i \gets 1 \textbf{ to } |\mtx{G}|$} \label{line:cv:g:start}
			\State $(\bar{\vec{y}}_\vec{1}, \vec{r_1}) \gets$ \Call{CreateFirstRanking}{$V, \vec{x_{hq}}, \bar{\mtx{X}}_\mathbf{1}, f, n_s$}
			\If {\Call{IsInteraction}{$V, \mtx{G}[i], \vec{x_{hq}}, \bar{\mtx{X}}_\mathbf{1}, \bar{\vec{x}}_\vec{2}, \bar{\vec{y}}_\vec{1}, \vec{r_1}, f, n_s$}} \label{line:cv:g:interact1}
			\State \Return $true$ 
			\EndIf
			\State $(\bar{\vec{y}}_\vec{1}, \vec{r_1}) \gets$ \Call{CreateFirstRanking}{$\mtx{G}[i], \vec{x_{hq}}, \bar{\mtx{X}}_\mathbf{1}, f, n_s$}
			\If {\Call{IsInteraction}{$\mtx{G}[i], V, \vec{x_{hq}}, \bar{\mtx{X}}_\mathbf{1}, \bar{\vec{x}}_\vec{2}, \bar{\vec{y}}_\vec{1}, \vec{r_1}, f, n_s$}} \label{line:cv:g:interact2}
			\State \Return $true$ 
			\EndIf
			\EndFor \label{line:cv:g:end}
			\State \Return $false$
			\EndFunction
		\end{algorithmic}
		\label{alg:considerVariables}
	\end{algorithm}
	
	\begin{algorithm}
		\caption{\textproc{CreateFirstRanking}}
		\scriptsize
		\begin{algorithmic}[1]
			\setstretch{1.1}
			\Statex \textbf{input:} $X_1$: set of variables, $\vec{x}$: a decision vector, $\bar{\mtx{X}}_\mathbf{1}$: a $n_s \times n$ matrix of samples' values, $f$: an optimization problem, $n_s$: the number of samples
			\Statex \textbf{output:} $\bar{\vec{y}}_\vec{1}$: values of $f$ calculated based on $\vec{x}$ and $\bar{\mtx{X}}_\mathbf{1}$, $\vec{r_1}$: the ranking of $\bar{\vec{y}}_\vec{1}$
			\Function{CreateFirstRanking}{$X_1, \vec{x}, \bar{\mtx{X}}_\mathbf{1}, f, n_s$}
			\State $\bar{\vec{x}} \gets \vec{x}$
			\For{$i \gets 1 \textbf{ to } n_s$} \label{line:ing:samples:start}
			\State $\bar{\vec{x}}[X_1] \gets \bar{\mtx{X}}_\mathbf{1}[i][X_1]$
			\State $\bar{\vec{y}}_\vec{1}[i] \gets f(\bar{\vec{x}})$ \Comment{values of $f$ calculated based on $\vec{x}$ and $\bar{\mtx{X}}_\mathbf{1}$}
			\EndFor \label{line:ing:samples:end}
			\State $\vec{r_1} \gets$ indices $i$ from $1$ to $n_s$ ordered by $\bar{\vec{y}}_\vec{1}[i]$ \label{line:ing:ranking}
			\State \Return $(\bar{\vec{y}}_\vec{1},\vec{r_1})$
			\EndFunction
		\end{algorithmic}
		\label{alg:createRankingR1}
	\end{algorithm}
	
	Function \textproc{IsInteraction}~(Pseudocode~\ref{alg:interacting}) is used to check the dependency between two groups of variables denoted as $X_1$ and $X_2$. To this end, we create ranking $\vec{r_2}$ and check if it differs from $\vec{r_1}$. To save computational effort, the order of ranking $\vec{r_2}$ is checked during its creation.
	
	\begin{algorithm}
		\caption{\textproc{IsInteraction}; Checking if two sets of variables are interacting}
		\scriptsize
		\begin{algorithmic}[1]
			\setstretch{1.1}
			\Statex \textbf{input:} $X_1$, $X_2$: disjoint sets of variables, $\vec{x_{hq}}$: a high-quality decision vector, $\bar{\mtx{X}}_\mathbf{1}$: a $n_s \times n$ matrix of samples' values, $\bar{\vec{x}}_\vec{2}$: a decision vector different than $\vec{x_{hq}}$, $\bar{\vec{y}}_\vec{1}$: values of $f$ calculated based on $\vec{x_{hq}}$ and $\bar{\mtx{X}}_\mathbf{1}$, $\vec{r_1}$: the ranking of $\bar{\vec{y}}_\vec{1}$, $f$: an optimization problem, $n_s$: the number of samples
			\Statex \textbf{output:} a decision if $X_1$ and $X_2$ are interacting
			\Function{IsInteraction}{$X_1, X_2, \vec{x_{hq}}, \bar{\mtx{X}}_\mathbf{1}, \bar{\vec{x}}_\vec{2}, \bar{\vec{y}}_\vec{1}, \vec{r_1}, f, n_s$}
			\State $\bar{\vec{x}} \gets \vec{x_{hq}}$
			\State $\bar{\vec{x}}[X_2] \gets \bar{\vec{x}}_\vec{2}[X_2]$ \label{line:ing:x2:start}
			\State $\bar{\vec{x}}[X_1] \gets \bar{\mtx{X}}_\mathbf{1}[\vec{r_1}[1]][X_1]$
			\State $\bar{\vec{y}}_\vec{2}[1] \gets f(\bar{\vec{x}})$ \Comment{values of $f$ calculated based on $\vec{x_{hq}}$, $\bar{\vec{x}}_\vec{2}$, and $\bar{\mtx{X}}_\mathbf{1}$}
			\For{$i \gets 2 \textbf{ to } n_s$} \label{line:ing:it}
			\State $\epsilon_1 \gets f_{\gamma}(\sqrt{n} + 1) \cdot (|\bar{\vec{y}}_\vec{1}[\vec{r_1}[i]]| + |\bar{\vec{y}}_\vec{1}[\vec{r_1}[i-1]]|)$ \Comment{formula~(\ref{eq:dg2:gamma})} \label{line:ing:e1}
			\If {$\text{sgn}(\bar{\vec{y}}_\vec{1}[\vec{r_1}[i]] - \bar{\vec{y}}_\vec{1}[\vec{r_1}[i-1]], \epsilon_1) = 0$} \Comment{formula~(\ref{eq:irrg:sgn})} \label{line:ing:equal}
			\State \textbf{continue}
			\EndIf
			\State $\bar{\vec{x}}[X_1] \gets \bar{\mtx{X}}_\mathbf{1}[\vec{r_1}[i]][X_1]$
			\State $\bar{\vec{y}}_\vec{2}[i] \gets f(\bar{\vec{x}})$
			\State $\epsilon_2 \gets f_{\gamma}(\sqrt{n} + 1) \cdot (|\bar{\vec{y}}_\vec{2}[i]| + |\bar{\vec{y}}_\vec{2}[i-1]|)$ \Comment{formula~(\ref{eq:dg2:gamma})} \label{line:ing:e2}
			\If {$\text{sgn}(\bar{\vec{y}}_\vec{2}[i] - \bar{\vec{y}}_\vec{2}[i-1], \epsilon_2) < 0$} \Comment{formula~(\ref{eq:irrg:sgn})} \label{line:ing:check}
			\State \Return $true$  
			\EndIf
			\EndFor \label{line:ing:x2:end}
			\State \Return $false$
			\EndFunction
		\end{algorithmic}
		\label{alg:interacting}
	\end{algorithm}
	
	Similarly to DG-based strategies~\cite{dg,xdg,dg2,rdg,rdg2,rdg3}, the result of this comparison of rankings may be affected by the inaccuracy of the representation of float numbers. Instead of using a user-defined $\epsilon$, we employ the automatic estimation of its value derived from~\cite{rdg2}~(lines~\ref{line:ing:e1} and~\ref{line:ing:e2}). If we denote two vectors as $\vec{x_1^*}$ and $\vec{x_2^*}$, then, analogously to formula~(\ref{eq:dg2:epsilon})~\cite{rdg2}, the estimation of $\epsilon$ is computed as $\epsilon = f_{\gamma}(\sqrt{n} + 1) \cdot (|f(\vec{x_1^*})| + |f(\vec{x_2^*})|)$ to decide which solution is better. Hence, the standard signum function may be reformulated to a version that takes $\epsilon$ into account (lines~\ref{line:ing:equal} and~\ref{line:ing:check}):
	\begin{equation}
	\small
	\label{eq:irrg:sgn}
	\text{sgn}(x, \epsilon) = \left \{
	\begin{array}{@{}rl@{}}
	-1, & \text{if}\ x < -\epsilon \\
	0, & \text{if}\ |x| \le \epsilon \\
	1, & \text{if}\ x > \epsilon
	\end{array}\right.
	\end{equation} 
	
	High-quality solutions occupy local optima or are close to them. Therefore, if the interaction exists, then it should reveal mainly while observing high-quality solutions~\cite{3lo}. To achieve this goal we start the analysis of rankings $\vec{r_1}$ and $\vec{r_2}$ from the samples of the highest quality in terms of $\vec{r_1}$. We believe that the best samples of different rankings may correspond to different optima if an interaction exists. Additionally, to decrease the probability that too small $\epsilon$ value causes false linkage, we discover an interaction only if both results of signum function (formula~(\ref{eq:irrg:sgn})) are decisive~(lines~\ref{line:ing:equal} and~\ref{line:ing:check}).
	
	To discover interactions, we use $n_s$ values for $X_1$ and only two values for $X_2$. Thus, \textproc{IsInteraction} is not a symmetric function. We call it twice in \textproc{ConsiderVariables} for each pair of sets of variables we examine~(lines~\ref{line:cv:v:interact1}, \ref{line:cv:v:interact2}, \ref{line:cv:g:interact1}, and~\ref{line:cv:g:interact2}).\par
	
	In the last paragraph of Section~\ref{sec:comp:falseOrMissingLinkage}, we show that constraints, which are a part of constrained optimization problems, e.g., bounding-box constraints, may prevent monotonicity checking from discovering interactions. To this end, we prefer to use a high-quality solution $\vec{x_{hq}}$ that was optimized before RRG execution. The motivation behind this step is as follows. If $\vec{x_{hq}}$ is close to a local optimum and $X_1$, $X_2$ interact, then the following situations are possible. (1) $\vec{x_{hq}} + \delta_2\vec{u_2}$ is not close to the local optimum, meaning that monotonicity intervals around $\vec{x_{hq}}$ changed and that RRG will likely detect them. (2) The monotonicity intervals around $\vec{x_{hq}}$ did not change, but the relations between function values around $\vec{x_{hq}}$ changed (see Fig.~\ref{fig:greedyAndSeparability:b}). In such a situation, RRG is likely to discover interaction. (3) The monotonicity intervals and the relations of function values around $\vec{x_{hq}}$ remain the same. RRG will not discover interaction in this case, but such a situation is not likely if $X_1$ and $X_2$ interact. Additionally, the number of local optima also influences the RRG sensitivity. The higher number of local optima, the higher number of points from different basins of attraction that have a similar interaction discovery potential as $\vec{x_{hq}}$. Thus, RRG will have a higher chance of interaction discovery.
	
	Function \textproc{Interact}~(Pseudocode~\ref{alg:interact}) extends function \textproc{IsInteraction}. Its objective is to search for groups of variables in $\mtx{G}_\mathbf{2}$ that interact with at least one group from $\mtx{G}_\mathbf{1}$. The output of \textproc{Interact} is a set of groups $\mtx{G}_\mathbf{1}^\mathbf{*}$ such that $\mtx{G}_\mathbf{1} \subseteq \mtx{G}_\mathbf{1}^\mathbf{*}$. Initially, $\mtx{G}_\mathbf{1}^\mathbf{*}$ is a copy of $\mtx{G}_\mathbf{1}$~(line~\ref{line:interact:copy}). To check if $\mtx{G}_\mathbf{1}$ and $\mtx{G}_\mathbf{2}$ interact, they are flatten (let $\mtx{S} = \{S_1, ..., S_k\}$ be a set of $k$ sets, then flattening $\mtx{S}$ results in a set $\bigcup \mtx{S} = S_1 \cup ... \cup S_k$) to $X_1 = \bigcup \mtx{G}_\mathbf{1}$ and $X_2 = \bigcup \mtx{G}_\mathbf{2}$~(lines~\ref{line:interact:flatten1} and~\ref{line:interact:flatten2}), and \textproc{IsInteraction} is called~(line~\ref{line:interacting:exec}). If an interaction is discovered, $\mtx{G}_\mathbf{2}$ is divided into two subsets (preferably of equal size) $\mtx{G}_\mathbf{2}^\mathbf{1}$ and $\mtx{G}_\mathbf{2}^\mathbf{2}$ (lines~\ref{line:interacting:g21} and~\ref{line:interacting:g22}). \textproc{Interact} is then called recursively for $\mtx{G}_\mathbf{2}^\mathbf{1}$ and $\mtx{G}_\mathbf{2}^\mathbf{2}$ (lines~\ref{line:interacting:callG21} and~\ref{line:interacting:callG22}). Once an interaction is found for $\mtx{G}_\mathbf{2}$ containing only one element (group), then this single group is added to $\mtx{G}_\mathbf{1}^\mathbf{*}$ (line~\ref{line:interact:oneElement}).
	
	\begin{algorithm}
		\caption{\textproc{Interact}; Searching for groups of variables that are interacting, but were not discovered yet}
		\scriptsize
		\begin{algorithmic}[1]
			\setstretch{1.1}
			\Statex \textbf{input:} $\mtx{G}_\mathbf{1}$, $\mtx{G}_\mathbf{2}$: disjoint sets of groups of non-separable variables, $\vec{x_{hq}}$: a high-quality decision vector, $\bar{\mtx{X}}_\mathbf{1}$: a $n_s \times n$ matrix of samples' values, $\bar{\vec{x}}_\vec{2}$: a decision vector different than $\vec{x_{hq}}$, $\bar{\vec{y}}_\vec{1}$: values of $f$ calculated based on $\vec{x_{hq}}$ and $\bar{\mtx{X}}_\mathbf{1}$, $\vec{r_1}$: the ranking of $\bar{\vec{y}}_\vec{1}$, $f$: an optimization problem, $n_s$: the number of samples
			\Statex \textbf{output:} $\mtx{G}_\mathbf{1}^\mathbf{*}$: the union of $\mtx{G}_\mathbf{1}$ and some groups from $\mtx{G}_\mathbf{2}$ that interact with at least one group from $\mtx{G}_\mathbf{1}$
			\Function{Interact}{$\mtx{G}_\mathbf{1}, \mtx{G}_\mathbf{2}, \vec{x_{hq}}, \bar{\mtx{X}}_\mathbf{1}, \bar{\vec{x}}_\vec{2}, \bar{\vec{y}}_\vec{1}, \vec{r_1}, f, n_s$}
			\State $\mtx{G}_\mathbf{1}^\mathbf{*} \gets \mtx{G}_\mathbf{1}$ \label{line:interact:copy}
			\State $X_1 \gets $ flatten $\mtx{G}_\mathbf{1}$ \label{line:interact:flatten1}
			\State $X_2 \gets $ flatten $\mtx{G}_\mathbf{2}$ \label{line:interact:flatten2}
			\If {\Call{IsInteraction}{$X_1, X_2, \vec{x_{hq}}, \bar{\mtx{X}}_\mathbf{1}, \bar{\vec{x}}_\vec{2}, \bar{\vec{y}}_\vec{1}, \vec{r_1}, f, n_s$}} \label{line:interacting:exec}
			\If {$|\mtx{G}_\mathbf{2}| = 1$} \label{line:interact:detected:start}
			\State Add the group from $\mtx{G}_\mathbf{2}$ to $\mtx{G}_\mathbf{1}^\mathbf{*}$ \label{line:interact:oneElement}
			\Else
			\State $\mtx{G}_\mathbf{2}^\mathbf{1} \gets$ the first half of $\mtx{G}_\mathbf{2}$ \label{line:interacting:g21}
			\State $\mtx{G}_\mathbf{2}^\mathbf{2} \gets$ the second half of $\mtx{G}_\mathbf{2}$ \label{line:interacting:g22}
			\State $\mtx{G}_\mathbf{1}^\mathbf{1*} \gets$ \Call{Interact}{$\mtx{G}_\mathbf{1}, \mtx{G}_\mathbf{1}^\mathbf{1}, \vec{x_{hq}}, \bar{\mtx{X}}_\mathbf{1}, \bar{\vec{x}}_\vec{2}, \bar{\vec{y}}_\vec{1}, \vec{r_1}, f, n_s$} \label{line:interacting:callG21}
			\State $\mtx{G}_\mathbf{1}^\mathbf{2*} \gets$ \Call{Interact}{$\mtx{G}_\mathbf{1}, \mtx{G}_\mathbf{2}^\mathbf{2}, \vec{x_{hq}}, \bar{\mtx{X}}_\mathbf{1}, \bar{\vec{x}}_\vec{2}, \bar{\vec{y}}_\vec{1}, \vec{r_1}, f, n_s$} \label{line:interacting:callG22}
			\State Add all new groups from $\mtx{G}_\mathbf{1}^\mathbf{1*}$ to $\mtx{G}_\mathbf{1}^\mathbf{*}$ \label{line:interact:addingG11}
			\State Add all new groups from $\mtx{G}_\mathbf{1}^\mathbf{2*}$ to $\mtx{G}_\mathbf{1}^\mathbf{*}$ \label{line:interact:addingG21}
			\EndIf \label{line:interact:detected:end}
			\EndIf
			\State \Return $\mtx{G}_\mathbf{1}^\mathbf{*}$ \label{line:interact:independent}
			\EndFunction
		\end{algorithmic}
		\label{alg:interact}
	\end{algorithm}
	
	The general procedure of RRG is presented in Pseudocode~\ref{alg:rrg}. RRG requires providing the interaction matrix $\Theta$, which single element $\theta_{p,q} \in \{0, 1\}$ informs if two variables, $x_p$ and $x_q$, are dependent. $\Theta$ may be represented by a graph. Finding the connected components of this graph leads to identifying groups of interacting variables~\cite{dg2}. Considering remarks to \textproc{IsInteraction} presented above, we assume that at least one provided decision vector is of high quality. First, using $\Theta$, we create groups of interacting variables. The order of variables within groups is random to remove possible bias (lines~\ref{line:rrg:createGroups:start}--\ref{line:rrg:createGroups:end}). Then, according to the description presented in Section~\ref{sec:irrg:basic}, we create $\bar{\mtx{X}}_\mathbf{1}$ (lines~\ref{line:rrg:mtxX1:start}--\ref{line:rrg:mtxX1:end}). Using \textproc{ConsiderVariables} we decide if variables that were not found interacting yet, shall take part in the interaction discovery (line~\ref{line:rrg:addingV}). We shuffle $\mtx{G}$ to remove possible bias and create $\mtx{G}_\mathbf{1}$ and $\mtx{G}_\mathbf{2}$ (lines~\ref{line:rrg:shuffle}--\ref{line:rrg:g2}). \par
	
	\begin{algorithm}
		\caption{\textproc{RRG}; Recursive Ranking Grouping}
		\scriptsize
		\begin{algorithmic}[1]
			\setstretch{1.1}
			\Statex \textbf{input:} $\vec{x_{hq}}$: a high-quality decision vector, $\bar{\vec{x}}_\vec{2}$: a decision vector different than $\vec{x_{hq}}$, $\Theta$: the interaction matrix, $f$: an optimization problem, $n$: the problem size, $\vec{lb}$: the lower bounds of $f$, $\vec{ub}$: the upper bounds of $f$, $n_s$: the number of samples
			\Statex \textbf{output:} $\mtx{NonSeps}$: groups of non-separable variables
			\Function{RRG}{$\vec{x_{hq}}, \bar{\vec{x}}_\vec{2}, \Theta, f, n, \vec{lb}, \vec{ub}, n_s$}
			\State $\mtx{NonSeps} \gets \varnothing$
			\State $\mtx{G} \gets \varnothing$ \Comment{groups of non-separable variables to detect new linkage} \label{line:rrg:formingGroups:start}
			\State $V \gets$ variables from $x_1$ to $x_n$
			\For{$v \in V$} \label{line:rrg:createGroups:start}
			\State $Inters \gets$ extract variables interacting with $v$ from $\Theta$
			\If {$|Inters| > 1$}
			\State $V \gets V - Inters$ \Comment{the difference of sets}
			\State $Inters \gets$ shuffle $Inters$ \label{line:rrg:shuffleG}
			\State Add $Inters$ to $\mtx{G}$ as a group
			\EndIf  \label{line:rrg:createGroups:end}
			\EndFor \label{line:rrg:formingGroups:end}
			\For{$i \gets 1 \textbf{ to } n$} \Comment{building a $n \times n_s$ matrix of samples' values} \label{line:rrg:mtxX1:start}
			\State $\bar{\mtx{X}}_\mathbf{1}^\intercal[i] \gets n_s$ values evenly taken from $[\vec{lb}[i], \vec{ub}[i]]$ 
			\State $\bar{\mtx{X}}_\mathbf{1}^\intercal[i] \gets$ shuffle $\bar{\mtx{X}}_\mathbf{1}^\intercal[i]$
			\EndFor 
			\State $\bar{\mtx{X}}_\mathbf{1} \gets$ transpose $\bar{\mtx{X}}_\mathbf{1}^\intercal$ \Comment{a $n_s \times n$ matrix of samples' values} \label{line:rrg:mtxX1:end}
			\If {\Call{ConsiderVariables}{$V, \mtx{G}, \vec{x_{hq}}, \bar{\mtx{X}}_\mathbf{1}, \bar{\vec{x}}_\vec{2}, f, n_s$}}
			\State Add all variables from $V$ to $\mtx{G}$ as single groups \label{line:rrg:addingV}
			\EndIf
			\State $\mtx{G} \gets$ shuffle the order of groups in $\mtx{G}$ \label{line:rrg:shuffle}
			\State $\mtx{G}_\mathbf{1} \gets$ take the first group from $\mtx{G}$ \label{line:rrg:g1}
			\State $\mtx{G}_\mathbf{2} \gets$ the other groups from $\mtx{G}$ \label{line:rrg:g2}
			\While{$|\mtx{G}_\mathbf{2}| > 0$} \label{line:rrg:stopCondition}
			\State $X_1 \gets$ flatten $\mtx{G}_\mathbf{1}$ \label{line:rrg:flattenG1}
			\State $(\bar{\vec{y}}_\vec{1}, \vec{r_1}) \gets$ \Call{CreateFirstRanking}{$X_1, \vec{x_{hq}}, \bar{\mtx{X}}_\mathbf{1}, f, n_s$}
			\State $\mtx{G}_\mathbf{1}^\mathbf{*} \gets$ \Call{Interact}{$\mtx{G}_\mathbf{1}, \mtx{G}_\mathbf{2}, \vec{x_{hq}}, \bar{\mtx{X}}_\mathbf{1}, \bar{\vec{x}}_\vec{2}, \bar{\vec{y}}_\vec{1}, \vec{r_1}, f, n_s$} \label{line:rrg:interact}
			\If {$|\mtx{G}_\mathbf{1}| = |\mtx{G}_\mathbf{1}^\mathbf{*}|$} \label{line:rrg:g1g11Size}
			\If {$|\mtx{G}_\mathbf{1}| = 1$}
			\State $minSize = \min_{i} |\mtx{G}_\mathbf{2}[i]|$ \label{line:rrg:g1g11EqualNotFound:start}
			\If {$|\mtx{G}_\mathbf{1}[1]| \ge \max \{minSize, 2\}$} \label{line:rrg:minSize:end}
			\State Remove half variables from $\mtx{G}_\mathbf{1}[1]$ \label{line:rrg:half}
			\Else
			\State $\mtx{G}_\mathbf{1} \gets$ take the first group from $\mtx{G}_\mathbf{2}$ \label{line:rrg:g1g11EqualNotFoundNotFound:start}
			\State $\mtx{G}_\mathbf{2} \gets \mtx{G}_\mathbf{2}$ excluding the first group \label{line:rrg:g1g11EqualNotFoundNotFound:end}
			\EndIf \label{line:rrg:g1g11EqualNotFound:end}
			\Else
			\State Add flattened $\mtx{G}_\mathbf{1}$ to $\mtx{NonSeps}$ \label{line:rrg:g1g11EqualFound:start}
			\State $\mtx{G}_\mathbf{1} \gets$ take the first group from $\mtx{G}_\mathbf{2}$
			\State $\mtx{G}_\mathbf{2} \gets \mtx{G}_\mathbf{2}$ excluding the first group \label{line:rrg:g1g11EqualFound:end}
			\EndIf
			\Else 
			\State $\mtx{G}_\mathbf{1} \gets \mtx{G}_\mathbf{1}^\mathbf{*}$ \label{line:rrg:g1g11NotEqual:start}
			\State $\mtx{G}_\mathbf{2} \gets \mtx{G}_\mathbf{2}$ excluding groups from $\mtx{G}_\mathbf{1}$ \label{line:rrg:g1g11NotEqual:end}
			\EndIf
			\EndWhile
			\If {$|\mtx{G}_\mathbf{1}| > 1$} \label{line:rrg:addLastGroup:start}
			\State Add flattened $\mtx{G}_\mathbf{1}$ to $\mtx{NonSeps}$ \label{line:rrg:addLastGroup:end}
			\EndIf
			\State \Return $\mtx{NonSeps}$
			\EndFunction
		\end{algorithmic}
		\label{alg:rrg}
	\end{algorithm}
	
	The latter part of RRG focuses on interaction discovery (lines~\ref{line:rrg:stopCondition}--\ref{line:rrg:addLastGroup:end}). We create ranking $\vec{r_1}$ and call \textproc{Interact} to find new interactions between groups of variables from $\mtx{G}_\mathbf{1}$ and $\mtx{G}_\mathbf{2}$. If interactions were found, we add dependent groups to $\mtx{G}_\mathbf{1}$, remove them from $\mtx{G}_\mathbf{2}$ (lines~\ref{line:rrg:g1g11NotEqual:start} and~\ref{line:rrg:g1g11NotEqual:end}) and call \textproc{Interact} again for the extended $\mtx{G}_\mathbf{1}$. If no new interactions were found, but $\mtx{G}_\mathbf{1}$ was extended and contains more than one group, then $\mtx{G}_\mathbf{1}$ is flattened and added as a single group to the output set of groups of interacting variables, $\mtx{G}_\mathbf{1}$ is replaced with a single group that is excluded from $\mtx{G}_\mathbf{2}$ (it is equivalent to the random choice, because $\mtx{G}_\mathbf{2}$ was shuffled) (lines~\ref{line:rrg:g1g11EqualFound:start}--\ref{line:rrg:g1g11EqualFound:end}). If no extensions to the group initially inserted to $\mtx{G}_\mathbf{1}$ were discovered (i.e. $\mtx{G}_\mathbf{1}$ contains only one group after calling \textproc{Interact}), then we randomly remove half of the variables from the single group in $\mtx{G}_\mathbf{1}$ (lines \ref{line:rrg:g1g11EqualNotFound:start}--\ref{line:rrg:half}). The intuition behind this step is as follows. If $|\bigcup\mtx{G}_\mathbf{1}|$ is large, then some subgroup of variables in $\bigcup\mtx{G}_\mathbf{1}$ may influence rankings $\vec{r_1}$ and $\vec{r_2}$ so significantly that the influence of the variables in $\bigcup\mtx{G}_\mathbf{2}$ will become negligible. Thus, if we remove some of the variables from a single group in $\mtx{G}_\mathbf{1}$ randomly, then some of the overwhelmed variables in $\bigcup\mtx{G}_\mathbf{2}$ may gain their influence on $\vec{r_1}$ and $\vec{r_2}$ rankings, resulting in the interaction discovery and mitigating missing linkage. Usually, the situation described above occurs when most interactions are discovered. Thus, it seems reasonable to decrease the size of a single group in $\mtx{G}_\mathbf{1}$ only if its size is not less than the size of the shortest group in $\mtx{G}_\mathbf{2}$. The above procedure mitigates missing linkage and does not seem computationally expensive.

	\subsection{Incremental Grouping}
	\label{sec:irrg:ig}
	
	IRRG builds the interaction matrix $\Theta$ incrementally. The general IRRG procedure is presented in Pseudocode~\ref{alg:irrg}. First, a user-defined optimizer is executed to find $\vec{x_{hq}}$ (a high-quality solution) (line~\ref{line:irrg:optimize}), and the 
	interaction matrix $\Theta$ is initialized as reporting no interactions (line~\ref{line:irrg:theta}). Then, RRG is called to get $\mtx{NonSepsTmp}$\textemdash the set of groups of interacting variables (line~\ref{line:irrg:rrg}). This set is used to update $\Theta$ by inserting the newly discovered interactions (line~\ref{line:irrg:utheta}). The $\Theta$ updates ensure transitivity. These operations are repeated until RRG will not find any new interaction in $\epsilon_{sti}$ (a user-defined parameter) iterations in a row or if RRG did not discover any new interactions during the first iteration (line \ref{line:irrg:terminate}). The latter part of this condition was introduced to decrease the computational cost of IRRG. The motivation is as follows. The lower is the number of interactions in $\Theta$, the easier is to supplement it. Thus, if RRG can not discover any new interactions in the first iteration, then repeating it does not seem reasonable.
	
	\begin{algorithm}
		\caption{Incremental Recursive Ranking Grouping}
		\scriptsize
		\begin{algorithmic}[1]
			\setstretch{1.1}
			\Statex \textbf{input:} $f$: an optimization problem, $n$: the problem size, $\vec{lb}$: the lower bounds of $f$, $\vec{ub}$: the upper bounds of $f$, $optimizer$: using during the initial optimization, $n_s$: the number of samples, $\epsilon_s$: the preferred size of groups of separable variables, $\epsilon_{sti}$: the stale iterations threshold
			\Statex \textbf{output:} $\mtx{Seps}$: groups of separable variables, $\mtx{NonSeps}$: groups of non-separable variables 
			\State $\mtx{Seps}, \mtx{NonSeps} \gets \varnothing$
			\State $\vec{x_{hq}} \gets$ optimize $f$ using $optimizer$ \label{line:irrg:optimize}
			\State $\Theta \gets n \times n$ identity matrix \Comment{the interaction matrix} \label{line:irrg:theta}
			\State $firstIter \gets true$ \Comment{marks the first iteration of IRRG}
			\State $cnt_{sti} \gets 0$ \Comment{the counter of the consecutive stale iterations}
			\State $terminate \gets false$
			\While {\textbf{not} $terminate$}
			\State $\vec{x_{lq}} \gets$ create a random solution \label{line:irrg:random}
			\State $\mtx{NonSepsTmp} \gets$ \Call{RRG}{$\vec{x_{hq}}, \vec{x_{lq}}, \Theta, f, n, \vec{lb}, \vec{ub}, n_s$} \label{line:irrg:rrg}
			\State Update $\Theta$ taking groups from $\mtx{NonSepsTmp}$ into account \label{line:irrg:utheta}
			\If {any new interactions were not discovered} \Comment{see line~\ref{line:irrg:utheta}}
			\State $cnt_{sti} \gets cnt_{sti} + 1$ \label{line:irrg:ipp}
			\State $terminate \gets firstIter$ \textbf{or} $cnt_{sti} = \epsilon_{sti}$ \label{line:irrg:terminate}
			\Else
			\State $cnt_{sti} \gets 0$
			\EndIf
			\State $firstIter \gets false$
			\EndWhile
			\State $V \gets$ variables from $x_1$ to $x_n$
			\State $S \gets \varnothing$ \Comment{separable variables before their grouping into $\mtx{Seps}$}
			\For{$v \in V$} \label{lines:irrg:sepsNonSeps:start}
			\State $Inters \gets$ extract variables interacting with $v$ from $\Theta$
			\If {$|Inters| = 1$}
			\State Add $v$ to $S$
			\Else
			\State Add $Inters$ to $\mtx{NonSeps}$ as a group
			\EndIf
			\State $V \gets V - Inters$ \Comment{the difference of sets}
			\EndFor \label{lines:irrg:sepsNonSeps:end}
			\While {$|S| > 0$} \label{line:irrg:eps_s:start}
			\If {$|S| < \epsilon_s$}
			\State Add $S$ to $\mtx{Seps}$ as a group
			\State $S \gets \varnothing$
			\Else
			\State Add $\epsilon_s$ variables from $S$ to $\mtx{Seps}$ as a group
			\State Delete $\epsilon_s$ variables from $S$
			\EndIf
			\EndWhile \label{line:irrg:eps_s:end}
			\State \Return $(\mtx{Seps}, \mtx{NonSeps})$
		\end{algorithmic}
		\label{alg:irrg}
	\end{algorithm}
	
	IRRG returns two sets of variable groups. $\mtx{NonSeps}$ contains groups of interacting variables, whereas in $\mtx{Seps}$, separable variables are joined into groups of a user-defined size $\epsilon_s$ (lines~\ref{line:irrg:eps_s:start}--\ref{line:irrg:eps_s:end}). This step is adopted from RDG3~\cite{rdg3} to improve separable variables processing by CC frameworks.
	
	IRRG can be tuned by adjusting values of four parameters: an initial optimizer, the number of samples $n_s$, the stale iteration threshold $\epsilon_{sti}$, and the preferred size of groups of separable variables $\epsilon_s$. The optimizer is executed to find a high-quality solution $\vec{x_{hq}}$. In case of LSGO problems, especially partially separable ones, we recommend exploring the search space first and then exploiting the found promising region. Otherwise, high-quality solutions for some subproblems may not be found. However, for some problems, the initial optimization will not be helpful and will not lead to obtaining better decomposition. The cost of this initial optimization should be reasonable, e.g., at most $2 \cdot n_s \cdot n$ FFEs, which corresponds to the lowest time complexity of IRRG. The higher the values of $n_s$ and $\epsilon_{sti}$, the higher the probability of obtaining better decomposition. However, the decomposition cost increases at the same time. The discussion about $\epsilon_s$ can be found in~\cite{rdg3}.
	
	\subsection{Comparison with Other Monotonicity Checking Strategies}
	
	IRRG and FVIL are recursive monotonicity checking strategies. Both use formula~(\ref{eq:nmd:nonseparability}) to decide if two groups of variables ($X_1$ and $X_2$) are interacting. If the interaction is discovered, $X_2$ is divided into two (nearly) equally sized groups, $X_2^1$ and $X_2^2$. Then, $X_1$ is checked if it interacts with $X_2^1$ or $X_2^2$. FVIL creates a new sample ($\delta_1$, $\delta_2$, $\vec{u_1}$, $\vec{u_2}$, $\vec{x^*}$) for each iteration check randomly. Consequently, if $X_1$ interacts with $X_2$ because variables $x_p \in X_1$ and $x_q \in X_2$ are interacting and $x_q$ is a member of $X_2^1$ afterwards, then the interaction between $X_1$ and $X_2^1$ could not be discovered because of different samples. Therefore, in IRRG, values assigned to variables in $X_1$, $X_2^1$, and $X_2^2$ are the same as in the previous step. Due to repeating value assignments, some FFEs may be cached and reused in the next recursive steps.
	
	Another difference that causes FVIL to be more vulnerable to missing linkage is the size of $X_1$. In FVIL, $X_1$ is always a single-element group. In IRRG, variables found dependent are added to $X_1$ in subsequent interaction checks. Subsequent iterations of IRRG also try to extend the already found groups of interacting variables. Such interaction checks using larger $X_1$ decrease the probability of missing linkage. Finally, FVIL does not apply the initial optimization. As explained in Section~\ref{sec:irrg:rrg}, using $\vec{x_{hq}}$ may significantly mitigate missing linkage for some optimization problems.
	
	Some monotonicity checking strategies employ an optimizer during interaction search~\cite{lvi, ccvil}, e.g., Cooperative Coevolution with Variable Interaction Learning~(CCVIL)~\cite{ccvil}. CCVIL executes an optimizer for each variable in each interaction discovery phase. However, the optimizer is executed only for one iteration, and its population is limited to only three individuals. Thus, we may expect that $\vec{x_{hq}}$ employed by IRRG is of significantly higher quality.
	
	\subsection{Time Complexity}
	\label{sec:irrg:complexity}
	
	IRRG is a recursive decomposition strategy, similarly to RDG-based strategies~\cite{rdg,rdg2,rdg3}. The typical time complexity (considering FFEs) for this class is $\mathcal{O}(n\log(n))$. In this paper, we consider LSGO, thus, we may expect that for large $n$, the cost of initial optimization (held to obtain $\vec{x_{hq}}$) is negligible. Thus, we ignore this part of IRRG in the analysis presented below. We refer to the analogous analysis of RDG~\cite{rdg}. The creation of $\vec{r_1}$ ranking costs $n_s$ FFEs in each IRRG iteration. The cost of a single \textproc{Interact} execution is not higher than $n_s$ FFEs (the analogous operation in RDG requires three FFEs~\cite{rdg}). Thus, the cost of decomposing $n$-dimensional problem in the first iteration of IRRG is as follows.
	
	\begin{enumerate}
		\item Fully separable problems\textemdash about $n_s \cdot n + n_s \cdot n$ FFEs.
		\item Fully non-separable problems\textemdash at most $2 \cdot n_s \cdot n + n_s$ FFEs.
		\item Partially separable problems with $m$ subproblems of size $l$\textemdash at most $2 \cdot n_s \cdot n \cdot \log_2(n) + n_s \cdot m$ FFEs.
		\item  Partially separable problems with one non-separable subproblem of size $l$\textemdash at most $n_s \cdot (n - l) + 2 \cdot n_s \cdot l \cdot \log_2(n) + n_s \cdot (n - l + 1)$ FFEs.
		\item Rosenbrock function~\cite{cec2013} that is an overlapping problem\textemdash about $2 \cdot n_s \cdot n \cdot \log_2(n) + n_s \cdot n/2$ FFEs.
	\end{enumerate}
	
	For consecutive IRRG iterations, we propose the similar analysis. Let us denote the number of variables that were not found interacting yet as $n_V$ and the number of already discovered groups of interacting variables as $n_{\mtx{G}}$. Thus, $n_V + n_{\mtx{G}} < n$. If all possible interactions were already discovered, then similar to fully separable problems, approximately $2 \cdot n_s \cdot (n_V + n_{\mtx{G}})$ additional FFEs are needed. Otherwise, since $n_V + n_{\mtx{G}} < n$, each consecutive IRRG iteration will still use less FFEs than the first iteration.
	
	\textproc{RRG} includes the operation of removing half of variables from the single group belonging to $\mtx{G}_\mathbf{1}$ (Pseudocode~\ref{alg:rrg}, line~\ref{line:rrg:half}). The number of variable removals will be the highest when $|\bigcup \mtx{G}_\mathbf{1}|=n-1 \land |\bigcup \mtx{G}_\mathbf{2}|=1$ and will not exceed $\log_2(n-1)$. The cost of each cut is not higher than $2 \cdot n_s$ FFEs, $n_s$ FFEs for the creation of ranking $\vec{r_1}$ and another $n_s$ FFEs for \textproc{Interact}. Thus, the total cost of each cut is below $2 \cdot n_s \cdot \log_2(n-1)$ FFEs. In the worst scenario, the cost of \textproc{ConsiderVariables} resulting from calling \textproc{IsInteraction} and creating ranking $\vec{r_1}$ twice for single variables and twice for each group of interacting variables is approximately $4 \cdot n_s + 4 \cdot n_s \cdot n_{\mtx{G}}$ FFEs. Therefore, if $n_s \ll n$ and the number of all IRRG iterations is much smaller than $n$, the time complexity of IRRG is $\mathcal{O}(n\log(n))$.\par
	
	One of the cases with a high number of IRRG iterations is as follows. The considered problem consists of $n / 2$ subproblems of size $2$, one interaction is found during the first iteration, and at every $\epsilon_{sti}$th iteration another single interation is found. Then, the number of IRRG iterations is $1 + \epsilon_{sti} \cdot n / 2$. Since the time complexity of finding a new single interaction is $\mathcal{O}(n)$, the worst time complexity of IRRG is $\mathcal{O}(n^2)$.
	
	\section{Experiments and Results Analysis}
	\label{sec:results}
	
	The objective of the experiments was twofold. First\textemdash to check if the proposed problem decomposition strategy is more robust than the DG-based strategies. The second objective was to verify if IRRG can be successfully embedded into CC frameworks. To this end, we integrate IRRG with CCBC~\cite{cbcc, cc2018} and CCFR2~\cite{ccfr2}, which were shown effective in solving LSGO problems. We compare these methods denoted as CBCC-IRRG and CCFR2-IRRG with their versions employing RDG3 denoted as CBCC-RDG3~\cite{rdg3,cbcc,cc2018} and CCFR2-RDG3~\cite{rdg3,ccfr2}, respectively. The competing methods were supplemented by SHADE-ILS~\cite{shade-ils}. Note that CBCC-RDG3 and SHADE-ILS are the state-of-the-art methods when LSGO is considered. To the best of our knowledge, CCFR2-RDG3 was not proposed yet. Nevertheless, considering it among the competing methods seems justified and desirable. IRRG- and RDG3-embedded CCs significantly outperformed CBCC-FVIL~\cite{cbcc,fvil} and CCFR2-FVIL~\cite{ccfr2,fvil}. Therefore, FVIL-based results are reported in the supplementary material.
	
	This section is divided into four subsections. First, we present the setup of all considered optimization methods. Then, we describe the chosen test problems. In Section~\ref{sec:results:decomAccuracy}, we discuss the decomposition accuracy of IRRG and RDG3 as well as their decomposition costs. Finally, the optimization results are reported in the last subsection.
	
	\subsection{Optimization Methods' Setup}
	
	CC frameworks require component optimizers. For this purpose, we employ CMA-ES~\cite{cmaes}, which was shown to be the best option for both considered CC frameworks~\cite{rdg,ccfr2}. We use CMA-ES implementation\footnote{\url{https://github.com/CMA-ES/pycma}}, which was shown the most efficient in~\cite{cmaes-python}. The original implementation of CBCC-RDG3 was supported in Matlab\footnote{\url{https://bitbucket.org/yuans/rdg3}}. It was rewritten to Python to integrate it with the source code of CMA-ES. The implementation of CCFR2 was made based on its pseudocodes presented in~\cite{ccfr2}. For RDG3, we use its original Matlab source code. The decomposition that resulted from using this source code was used as an input to the Python implementations of CBCC and CCFR2. Similarly, IRRG was written in C++, but its output was passed to Python. The pack with the source code, the detailed results of all runs, and the results of all performed statistical tests can be downloaded from the public repository\footnote{\url{https://github.com/kommar/IRRG}}. For SHADE-ILS, we took the source code provided by its Authors\footnote{\url{https://github.com/dmolina/shadeils}}.
	
	The parameters of the considered decomposition strategies, CC frameworks, and SHADE-ILS are adopted from papers in which they were proposed or recently applied. For RDG3, we use $\epsilon_s = 100$ and $\epsilon_n = 50$~\cite{rdg3}. CBCC is executing CMA-ES for $1000$ FFEs and the smoothing factor $w$ is $0.5$~\cite{rdg3,cc2018}. For CCFR2, we use $w=0.1$ and CMA-ES (executed for $100$ iterations) as a component optimizer~\cite{ccfr2}. SHADE-ILS uses the population size of $100$ for its underlying differential evolution, MTS-LS1~\cite{mts} with the initial step set to $20$, the minimum improvement ratio is $5\%$ and the allowed number of iterations without the improvement is $3$~\cite{shade-ils}.
	
	The configuration of the proposed IRRG is as follows. To obtain $\vec{x_{hq}}$, we use two optimizers. First, we use SHADE with the recommended settings~\cite{shade} (population size: $100$; arc ratio: $2$; $p$best ratio: $0.1$; memory size: $10^3$). Then, the solution proposed by SHADE is optimized by MTS-LS1~\cite{mts} (search range: $20\%$ of the difference between the upper bound and the lower bound). The motivation behind using two optimizers is to consider SHADE as a global search optimizer and MTS-LS1 as a local search optimizer. They are executed for $5 \cdot 10^3$ and $1.5 \cdot 10^4$ FFEs, respectively. The other IRRG parameters are: $\epsilon_{sti} = 15$, $n_s = 10$, and $\epsilon_s = 100$ (the value of $\epsilon_s$ is adopted from~\cite{rdg3}). The FFE-based stop conditions and parameters $\epsilon_{sti}$ and $n_s$ were tuned to find their minimal values that still lead to the perfect decomposition of considered non-overlapping benchmarks.
	
	\subsection{Test Problems}
	\label{sec:res:problems}
	
	As benchmarks we use functions from IEEE CEC’2013 special session on LSGO~\cite{cec2013}. CEC'2013 was also considered by other up-to-date papers that considered LSGO~\cite{dg2,rdg,rdg2,rdg3,ccfr2,shade-ils}. However, in CEC'2013 almost all separable subfunctions are additively separable, which favors DG-based strategies (see Section~\ref{sec:comp:falseOrMissingLinkage}). To this end, we propose two new sets based on CEC'2013, which eliminate this drawback.
	
	Let us consider a function $f$ and its range $Y$. Let $g$ be a function that is strictly increasing over $Y$. Then, we can state that $\forall_{\vec{x_1^*}, \vec{x_2^*} \in \Omega} f(\vec{x_1^*}) < f(\vec{x_2^*}) \Rightarrow h(\vec{x_1^*}) < h(\vec{x_2^*})$, where $h = g \circ f$. For each such function composition, any result of a comparison between solutions does not change. Therefore, $f$ and $h$ have the same intervals of monotonicity, optima, and variable interactions (see Section~\ref{sec:mc:ell}). However, even if $f$ is additively separable, $h$ may be non-additively separable, e.g., when $g$ is a non-linear function. Nevertheless, their ideal interaction matrices are the same.
	
	The minimal value of all fifteen CEC'2013 functions is not lower than~$0$. Thus, as $g$ we need functions that are strictly increasing over non-negative arguments. In this paper, we use $g_1(y)=y^2$ and $g_2(y)=\sqrt{y}$. They were chosen, because they increase in different ways. The test cases created using $g_1$ and $g_2$ are denoted as $(f_i)^2$ and $\sqrt{f_i}$, respectively. For the sake of clarity, we define a set of the $i$th functions $F_i$ as $\{f_i, (f_i)^2, \sqrt{f_i}\}$ for $i \in \{1, ..., 15\}$. To summarize, we consider the following five categories of functions~\cite{ccfr2,cec2013}.
	
	\begin{enumerate}
		\item Fully separable functions ($F_1$--$F_3$).
		\item Partially separable functions with separable variables ($F_4$--$F_7$).
		\item Partially separable functions with no separable variables ($F_8$--$F_{11}$).
		\item Overlapping functions ($F_{12}$--$F_{14}$).
		\item Fully non-separable functions ($F_{15}$).
	\end{enumerate}
	Considering the type of interaction between separable subfunctions, we split the first four categories into:
	\begin{enumerate}
		\item Functions with additively separable subfunctions ($f_1$, $f_2$, and $f_4$--$f_{14}$).
		\item Functions with non-additively separable subfunctions ($f_3$, $(f_{1})^2$--$(f_{14})^2$, and $\sqrt{f_{1}}$--$\sqrt{f_{14}}$).
	\end{enumerate}
	The computation budget was set to $3 \cdot 10^6$ FFEs and each experiment was repeated $25$ times~\cite{cec2013,dg2,rdg,rdg2,rdg3,ccfr2,shade-ils}.
	
	As a real-world optimization problem with non-additively separable subproblems, we analyze a multi-path routing problem in computer and communication networks. In contrast to the traditional routing approach that transmits all traffic of a given demand along a single path, multi-path routing splits the traffic among several paths. Multi-path routing can improve network efficiency in terms of various measures of Quality of Service, including survivability, congestion and throughput~\cite{Pioro_Book_2004, Cideon_1999, Banner_2007}. More details, including the results, can be found in the supplementary material.
	
	\subsection{Accuracy and Costs of Decomposition}
	\label{sec:results:decomAccuracy}
	
	IRRG is a decomposition strategy that never reports false linkage (in contrast to DG-based strategies) and should not suffer from missing linkage, which is a typical drawback of monotonicity checking. We employ three measures proposed in~\cite{measures} to verify the accuracy of the decomposition. The first measure ($\rho_1$) returns the accuracy of finding interaction, and its value is maximum when there is no missing linkage. The second measure ($\rho_2$) refers to false linkage. The higher its value is, the lower is the number of non-existing interactions detected. The value of the third measure ($\rho_3$) is maximal when there is neither false nor missing linkage.
	\begin{equation}
	\small
	\rho_1 = \frac{\sum_{i=1}^n \sum_{j = 1, j \ne i}^n (\Theta \circ \Theta^*)_{i,j}}{\sum_{i=1}^n \sum_{j = 1, j \ne i}^n (\Theta^*)_{i,j}} \cdot 100\%
	\end{equation}
	\begin{equation}
	\small
	\rho_2 = \frac{\sum_{i=1}^n \sum_{j = 1, j \ne i}^n ((\vec{1} - \Theta) \circ (\vec{1} - \Theta^*))_{i,j}}{\sum_{i=1}^n \sum_{j = 1, j \ne i}^n (\vec{1} - \Theta^*)_{i,j}} \cdot 100\%
	\end{equation}
	\begin{equation}
	\small
	\rho_3 = \frac{\sum_{i=1}^n \sum_{j = 1, j \ne i}^n (\vec{1} - |\Theta - \Theta^*|)_{i,j}}{n \cdot (n - 1) / 2} \cdot 100\%
	\end{equation}
	where $n$ is the problem size, $\Theta^*$ is the ideal interaction matrix, $\vec{1}$ denotes the all-ones matrix, $\circ$ indicates the Hadamard product of two matrices, and $|\Theta - \Theta^*|$ is a matrix which elements are defined as $|(\Theta)_{i,j} - (\Theta^*)_{i,j}|$.
	
	\begin{table*}
		\caption{Decomposition accuracy}
		\label{tab:accuracy}
		\scriptsize
		\begin{tabularx}{\linewidth}{CCCCCCCCCCCCC}
			\toprule
			\multirow{4}{*}{Func} & \multicolumn{9}{>{\hsize=\dimexpr9\hsize+9\tabcolsep+\arrayrulewidth\relax}C}{IRRG} & \multicolumn{3}{>{\hsize=\dimexpr3\hsize+3\tabcolsep+\arrayrulewidth\relax}C}{RDG3} \\
			\cmidrule{2-13}
			& \multicolumn{3}{>{\hsize=\dimexpr3\hsize+3\tabcolsep+\arrayrulewidth\relax}C}{$\rho_1$} & \multicolumn{3}{>{\hsize=\dimexpr3\hsize+3\tabcolsep+\arrayrulewidth\relax}C}{$\rho_2$} & \multicolumn{3}{>{\hsize=\dimexpr3\hsize+3\tabcolsep+\arrayrulewidth\relax}C}{$\rho_3$} & \multirow{2}{*}{$\rho_1$} & \multirow{2}{*}{$\rho_2$} & \multirow{2}{*}{$\rho_3$} \\
			\cmidrule{2-10}
			& Med & Avg & Std & Med & Avg & Std & Med & Avg & Std & & & \\
			\midrule
			
			$f_{1}$ & N/A & N/A & N/A & \textbf{100\%} & \textbf{100\%} & \textbf{0\%} & \textbf{100\%} & \textbf{100\%} & \textbf{0\%} & N/A & \textbf{100\%} & \textbf{100\%} \\
			$(f_{1})^2$ & N/A & N/A & N/A & \textbf{100\%} & \textbf{100\%} & \textbf{0\%} & \textbf{100\%} & \textbf{100\%} & \textbf{0\%} & N/A & 39.97\% & 39.97\% \\
			$\sqrt{f_{1}}$ & N/A & N/A & N/A & \textbf{100\%} & \textbf{100\%} & \textbf{0\%} & \textbf{100\%} & \textbf{100\%} & \textbf{0\%} & N/A & 50.40\% & 50.40\% \\
			
			\addlinespace[0.5em]
			
			$f_{2}$ & N/A & N/A & N/A & \textbf{100\%} & \textbf{100\%} & \textbf{0\%} & \textbf{100\%} & \textbf{100\%} & \textbf{0\%} & N/A & \textbf{100\%} & \textbf{100\%} \\
			$(f_{2})^2$ & N/A & N/A & N/A & \textbf{100\%} & \textbf{100\%} & \textbf{0\%} & \textbf{100\%} & \textbf{100\%} & \textbf{0\%} & N/A & 0\% & 0\% \\
			$\sqrt{f_{2}}$ & N/A & N/A & N/A & \textbf{100\%} & \textbf{100\%} & \textbf{0\%} & \textbf{100\%} & \textbf{100\%} & \textbf{0\%} & N/A & 0\% & 0\% \\
			
			\addlinespace[0.5em]
			
			$f_{3}$ & N/A & N/A & N/A & \textbf{100\%} & \textbf{100\%} & \textbf{0\%} & \textbf{100\%} & \textbf{100\%} & \textbf{0\%} & N/A & 0\% & 0\% \\
			$(f_{3})^2$ & N/A & N/A & N/A & \textbf{100\%} & \textbf{100\%} & \textbf{0\%} & \textbf{100\%} & \textbf{100\%} & \textbf{0\%} & N/A & 0\% & 0\% \\
			$\sqrt{f_{3}}$ & N/A & N/A & N/A & \textbf{100\%} & \textbf{100\%} & \textbf{0\%} & \textbf{100\%} & \textbf{100\%} & \textbf{0\%} & N/A & 0\% & 0\% \\
			
			\addlinespace[0.5em]
			
			$f_{4}$ & \textbf{100\%} & \textbf{100\%} & \textbf{0\%} & \textbf{100\%} & \textbf{100\%} & \textbf{0\%} & \textbf{100\%} & \textbf{100\%} & \textbf{0\%} & \textbf{100\%} & \textbf{100\%} & \textbf{100\%} \\
			$(f_{4})^2$ & \textbf{100\%} & \textbf{100\%} & \textbf{0\%} & \textbf{100\%} & \textbf{100\%} & \textbf{0\%} & \textbf{100\%} & \textbf{100\%} & \textbf{0\%} & 89.26\% & 56.57\% & 57.13\% \\
			$\sqrt{f_{4}}$ & \textbf{100\%} & \textbf{100\%} & \textbf{0\%} & \textbf{100\%} & \textbf{100\%} & \textbf{0\%} & \textbf{100\%} & \textbf{100\%} & \textbf{0\%} & 70.70\% & 73.09\% & 73.05\% \\
			
			\addlinespace[0.5em]
			
			$f_{5}$ & \textbf{100\%} & \textbf{100\%} & \textbf{0\%} & \textbf{100\%} & \textbf{100\%} & \textbf{0\%} & \textbf{100\%} & \textbf{100\%} & \textbf{0\%} & 98.85\% & \textbf{100\%} & 99.98\% \\
			$(f_{5})^2$ & \textbf{100\%} & \textbf{100\%} & \textbf{0\%} & \textbf{100\%} & \textbf{100\%} & \textbf{0\%} & \textbf{100\%} & \textbf{100\%} & \textbf{0\%} & 89.22\% & 51.78\% & 52.42\% \\
			$\sqrt{f_{5}}$ & \textbf{100\%} & \textbf{100\%} & \textbf{0\%} & \textbf{100\%} & \textbf{100\%} & \textbf{0\%} & \textbf{100\%} & \textbf{100\%} & \textbf{0\%} & 89.83\% & 83.17\% & 83.29\% \\
			
			\addlinespace[0.5em]
			
			$f_{6}$ & \textbf{100\%} & \textbf{100\%} & \textbf{0\%} & \textbf{100\%} & \textbf{100\%} & \textbf{0\%} & \textbf{100\%} & \textbf{100\%} & \textbf{0\%} & \textbf{100\%} & 71.93\% & 72.42\% \\
			$(f_{6})^2$ & \textbf{100\%} & \textbf{100\%} & \textbf{0\%} & \textbf{100\%} & \textbf{100\%} & \textbf{0\%} & \textbf{100\%} & \textbf{100\%} & \textbf{0\%} & 70.13\% & 67.36\% & 67.41\% \\
			$\sqrt{f_{6}}$ & \textbf{100\%} & \textbf{100\%} & \textbf{0\%} & \textbf{100\%} & \textbf{100\%} & \textbf{0\%} & \textbf{100\%} & \textbf{100\%} & \textbf{0\%} & 84.86\% & 58.02\% & 58.48\% \\
			
			\addlinespace[0.5em]
			
			$f_{7}$ & \textbf{100\%} & \textbf{100\%} & \textbf{0\%} & \textbf{100\%} & \textbf{100\%} & \textbf{0\%} & \textbf{100\%} & \textbf{100\%} & \textbf{0\%} & \textbf{100\%} & \textbf{100\%} & \textbf{100\%} \\
			$(f_{7})^2$ & \textbf{100\%} & \textbf{100\%} & \textbf{0\%} & \textbf{100\%} & \textbf{100\%} & \textbf{0\%} & \textbf{100\%} & \textbf{100\%} & \textbf{0\%} & \textbf{100\%} & 92.62\% & 92.74\% \\
			$\sqrt{f_{7}}$ & \textbf{100\%} & \textbf{100\%} & \textbf{0\%} & \textbf{100\%} & \textbf{100\%} & \textbf{0\%} & \textbf{100\%} & \textbf{100\%} & \textbf{0\%} & \textbf{100\%} & 92.62\% & 92.74\% \\
			
			\addlinespace[0.5em]
			
			$f_{8}$ & \textbf{100\%} & \textbf{100\%} & \textbf{0\%} & \textbf{100\%} & \textbf{100\%} & \textbf{0\%} & \textbf{100\%} & \textbf{100\%} & \textbf{0\%} & 70.20\% & \textbf{100\%} & 97.98\% \\
			$(f_{8})^2$ & \textbf{100\%} & \textbf{100\%} & \textbf{0\%} & \textbf{100\%} & \textbf{100\%} & \textbf{0\%} & \textbf{100\%} & \textbf{100\%} & \textbf{0\%} & 63.88\% & 99.72\% & 97.28\% \\
			$f_{8}$ & \textbf{100\%} & \textbf{100\%} & \textbf{0\%} & \textbf{100\%} & \textbf{100\%} & \textbf{0\%} & \textbf{100\%} & \textbf{100\%} & \textbf{0\%} & 69.57\% & 38.80\% & 40.89\% \\
			
			\addlinespace[0.5em]
			
			$f_{9}$ & \textbf{100\%} & \textbf{100\%} & \textbf{0\%} & \textbf{100\%} & \textbf{100\%} & \textbf{0\%} & \textbf{100\%} & \textbf{100\%} & \textbf{0\%} & \textbf{100\%} & \textbf{100\%} & \textbf{100\%} \\
			$(f_{9})^2$ & \textbf{100\%} & \textbf{100\%} & \textbf{0\%} & \textbf{100\%} & \textbf{100\%} & \textbf{0\%} & \textbf{100\%} & \textbf{100\%} & \textbf{0\%} & 84.28\% & 44.36\% & 47.07\% \\
			$\sqrt{f_{9}}$ & \textbf{100\%} & \textbf{100\%} & \textbf{0\%} & \textbf{100\%} & \textbf{100\%} & \textbf{0\%} & \textbf{100\%} & \textbf{100\%} & \textbf{0\%} & 92.47\% & 63.37\% & 65.35\% \\
			
			\addlinespace[0.5em]
			
			$f_{10}$ & \textbf{100\%} & \textbf{100\%} & \textbf{0\%} & \textbf{100\%} & \textbf{100\%} & \textbf{0\%} & \textbf{100\%} & \textbf{100\%} & \textbf{0\%} & 93.28\% & \textbf{100\%} & 99.54\% \\
			$(f_{10})^2$ & \textbf{100\%} & \textbf{100\%} & \textbf{0\%} & \textbf{100\%} & \textbf{100\%} & \textbf{0\%} & \textbf{100\%} & \textbf{100\%} & \textbf{0\%} & 93.21\% & 98.82\% & 98.44\% \\
			$\sqrt{f_{10}}$ & \textbf{100\%} & \textbf{100\%} & \textbf{0\%} & \textbf{100\%} & \textbf{100\%} & \textbf{0\%} & \textbf{100\%} & \textbf{100\%} & \textbf{0\%} & 89.85\% & 98.84\% & 98.23\% \\
			
			\addlinespace[0.5em]
			
			$f_{11}$ & \textbf{100\%} & \textbf{100\%} & \textbf{0\%} & \textbf{100\%} & \textbf{100\%} & \textbf{0\%} & \textbf{100\%} & \textbf{100\%} & \textbf{0\%} & \textbf{100\%} & \textbf{100\%} & \textbf{100\%} \\
			$(f_{11})^2$ & \textbf{100\%} & \textbf{100\%} & \textbf{0\%} & \textbf{100\%} & \textbf{100\%} & \textbf{0\%} & \textbf{100\%} & \textbf{100\%} & \textbf{0\%} & 98.60\% & 43.58\% & 47.32\% \\
			$\sqrt{f_{11}}$ & \textbf{100\%} & \textbf{100\%} & \textbf{0\%} & \textbf{100\%} & \textbf{100\%} & \textbf{0\%} & \textbf{100\%} & \textbf{100\%} & \textbf{0\%} & 95.09\% & 49.35\% & 52.45\% \\
			
			\addlinespace[0.5em]
			
			$f_{12}$ & \textbf{99.20\%} & \textbf{99.07\%} & \textbf{0.32\%} & 84.86\% & 83.94\% & 4.80\% & 84.89\% & 83.97\% & 4.79\% & 98.10\% & \textbf{95.28\%} & \textbf{95.29\%} \\
			$(f_{12})^2$ & 99.20\% & 99.14\% & 0.29\% & \textbf{84.56\%} & \textbf{81.80\%} & \textbf{9.14\%} & \textbf{84.59\%} & \textbf{81.83\%} & \textbf{9.12\%} & \textbf{100\%} & 0\% & 0.20\% \\
			$\sqrt{f_{12}}$ & 99.20\% & 99.14\% & 0.36\% & \textbf{83.54\%} & \textbf{81.77\%} & \textbf{7.51\%} & \textbf{83.57\%} & \textbf{81.80\%} & \textbf{7.49\%} & \textbf{100\%} & 0\% & 0.20\% \\
			
			\addlinespace[0.5em]
			
			$f_{13}$ & \textbf{99.93\%} & \textbf{99.84\%} & \textbf{0.22\%} & 0.23\% & 18.04\% & 23.88\% & 8.44\% & 24.78\% & 21.90\% & 92.65\% & \textbf{98.61\%} & \textbf{98.12\%} \\
			$(f_{13})^2$ & 99.93\% & 99.83\% & 0.18\% & \textbf{0.23\%} & \textbf{21.33\%} & \textbf{24.36\%} & \textbf{8.44\%} & \textbf{27.79\%} & \textbf{22.34\%} & \textbf{100\%} & 0\% & 8.23\% \\
			$\sqrt{f_{13}}$ & 99.85\% & 99.81\% & 0.20\% & \textbf{0.23\%} & \textbf{21.77\%} & \textbf{26.30\%} & \textbf{8.44\%} & \textbf{28.20\%} & \textbf{24.12\%} & \textbf{100\%} & 0\% & 8.23\% \\
			
			\addlinespace[0.5em]
			
			$f_{14}$ & \textbf{99.70\%} & \textbf{99.70\%} & \textbf{0.24\%} & 37.54\% & 33.07\% & 28.28\% & 42.66\% & 38.55\% & 25.93\% & 94.55\% & \textbf{96.88\%} & \textbf{96.68\%} \\
			$(f_{14})^2$ & \textbf{99.78\%} & \textbf{99.78\%} & \textbf{0.22\%} & 13.35\% & 22.66\% & 25.41\% & 20.47\% & 29.01\% & 23.30\% & 87.83\% & \textbf{53.26\%} & \textbf{56.11\%} \\
			$\sqrt{f_{14}}$ & \textbf{99.89\%} & \textbf{99.78\%} & \textbf{0.24\%} & 0.35\% & 24.52\% & 27.28\% & 8.55\% & 30.72\% & 25.01\% & 73.85\% & \textbf{56.69\%} & \textbf{58.11\%} \\
			
			\addlinespace[0.5em]
			
			$f_{15}$ & \textbf{100\%} & \textbf{100\%} & \textbf{0\%} & N/A & N/A & N/A & \textbf{100\%} & \textbf{100\%} & \textbf{0\%} & \textbf{100\%} & N/A & \textbf{100\%} \\
			$(f_{15})^2$ & \textbf{100\%} & \textbf{100\%} & \textbf{0\%} & N/A & N/A & N/A & \textbf{100\%} & \textbf{100\%} & \textbf{0\%} & \textbf{100\%} & N/A & \textbf{100\%} \\
			$\sqrt{f_{15}}$ & \textbf{100\%} & \textbf{100\%} & \textbf{0\%} & N/A & N/A & N/A & \textbf{100\%} & \textbf{100\%} & \textbf{0\%} & \textbf{100\%} & N/A & \textbf{100\%} \\
			
			\bottomrule
		\end{tabularx}
	\end{table*}
	
	In Table~\ref{tab:accuracy}, we present the decomposition accuracy of IRRG and RDG3. IRRG is non-deterministic, thus, we report the summarized results of 50 runs. We summarize the runs of CBCC-IRRG and CCFR2-IRRG to show the quality of decomposition that was later on used during the optimization. For all non-overlapping functions, IRRG is highly repeatable\textemdash the standard deviation is equal to $0\%$ for these test cases. For functions with separable variables, $\rho_2$ and $\rho_3$ were computed just before joining them into groups of $\epsilon_s$ size (otherwise, IRRG and RDG3 would report false linkage). For the overlapping functions the $\rho_2$ values are below $100\%$ for IRRG, although it never reports false linkage. This is caused by the fact that $\rho_1$, $\rho_2$, and $\rho_3$ allow only for direct interactions, while IRRG and RDG3 consider also the indirect ones~\cite{xdg,dled}. \par

	For the separable and partially separable functions (additive or non-additive), the values of all measures are optimal for all IRRG runs, i.e., IRRG found all interactions and never reported false linkage regardless of the type of interaction between subfunctions (Table~\ref{tab:accuracy}, $F_1$--$F_{11}$). For the standard CEC'2013 functions, except for $f_3$ and $f_6$, RDG3 did not report false linkage. Such results were expected for $f_3$, because it is the only function from the standard CEC'2013 set that is non-additively separable. Thus, $\rho_2 = \rho_3 = 0\%$ for RDG3 reported for $f_3$ mean that all separable variables were grouped together. On the other hand, $f_6$ is additively separable, which makes $\rho_2 = 71.93\%$ an unexpected result. The possible reason behind this result seems to be the inaccuracy of automatic $\epsilon$ estimation procedure, i.e., in some cases the value of $\epsilon$ might be too low. Additionally, for $f_5$, $f_8$, and $f_{10}$ the value of $\rho_1$ is below $100\%$, which means that RDG3 did not discover all interactions. The most convincing explanation seems to be too high value of $\epsilon$. To omit the issue of missing linkage, IRRG executes RRG several times, which increases the risk of reporting false linkage (in the case when $\epsilon$ is too low). Nevertheless, Table~\ref{tab:accuracy} shows that considering only those samples for which the outcome of $f$ differs, in terms of the $\epsilon$ value, is sufficient to report false linkage never or very rarely.\par
	
	All modified $f_1$--$f_3$ functions are non-additively separable. Thus, we expect that for these functions $\rho_2$ is $0\%$ for RDG3. The results for $(f_1)^2$ and $\sqrt{f_1}$ have significantly higher values (although still much worse than the optimal $100\%$). These results are caused by cases when too high value of $\epsilon$ allows RDG3 to omit discovering some of the false dependencies. Moreover, RDG3 tries to limit the size of a single group to $\epsilon_n$, which may mitigate false linkage phenomenon. Note that the inappropriate value of $\epsilon$ influences RDG3 differently in $(f_1)^2$ and $\sqrt{f_1}$ cases. For the modified versions of partially separable functions, we also expect grouping all variables together by RDG3, i.e., $\rho_1=100\%$ and $\rho_2=0\%$. However, similarly to $(f_1)^2$--$(f_3)^2$ and $\sqrt{f_1}$--$\sqrt{f_3}$, too high value of $\epsilon$ may lead to mitigating the false linkage phenomenon. \par
	
	A perfect, unique decomposition does not exist for overlapping problems ($F_{12}$--$F_{14}$), and it may be beneficial to use various decompositions during the optimization process~\cite{3lo}. In all overlapping test cases, every variable is dependent (directly or indirectly) on all the others. For these problems, the value of $\rho_1$ is maximum when only one group is created, while $\rho_2$ is maximum if all created groups consist of only directly dependent variables. Finally, $\rho_3$ is a trade-off between $\rho_1$ and $\rho_2$. Thus, its value cannot reach $100\%$. Therefore, each IRRG execution may lead to various groupings and the standard deviation values for $F_{12}$--$F_{14}$ are higher than $0\%$.\par
	
	IRRG can detect both types of separability (additive and non-additive). Thus, we expect it to return similar results for each overlapping CEC'2013 function and its modifications. The results reported in Table~\ref{tab:accuracy} confirm this expectation\textemdash for most of the functions the values of all three measures are similar for the standard, squared, and square rooted function versions. This observation was verified by the unpaired Wilcoxon test (significance level of $5\%$) with the Holm-Bonferroni correction for measures $\rho_1$--$\rho_3$ and pairs $\big(f_i, (f_i)^2\big)$, $\big(f_i, \sqrt{f_i}\big)$, and $\big((f_i)^2, \sqrt{f_i}\big)$, where $i \in \{12, 13, 14\}$. Such results of statistical tests may be surprising for $f_{14}$ for which the differences between the median values of $\rho_1$ seem high. Nevertheless, it is justified by the high value of the standard deviation.\par
	
	For RDG3, the observations for $F_{12}$--$F_{14}$ are opposite to the observations for IRRG. For $f_{12}$--$f_{14}$, RDG3 is able to find a balance between missing and false linkage, i.e. the values of all measures are above $90\%$. However, for the modified versions of $f_{12}$ and $f_{13}$, the values of $\rho_2$ are equal to $0\%$. For $(f_{14})^2$ and $\sqrt{f_{14}}$, the results differ most probably due to the influence of the inappropriate value of $\epsilon$ for some interaction checks and the impact of $\epsilon_n$.\par
	
	Finally, both decomposition strategies handle the fully non-separable function $f_{15}$. In this case, the proposed modifications does not affect the quality of the decomposition.
	
	The results presented above show that IRRG outperforms RDG3 in decomposing the fully and partially separable functions. For fully non-separable functions, both strategies perform equally well, whereas for the overlapping functions the situation is indecisive. Hence, we can state that, in general, IRRG proposes the higher quality decomposition than RDG3. However, this advantage has its costs\textemdash as shown in Table~\ref{tab:ffes} for the standard CEC'2013 functions, IRRG consumes from $3.6$ up to even $14.6$ times more FFEs on average. For the modified functions the overhead is similar except the overlapping functions, which are up to $124.2$ times more expensive than for RDG3. The results of the Student's t-test with the Holm-Bonferroni correction confirm that the differences between the RDG3 and IRRG costs are statistically significant.\par
	
	For $F_8$ and $F_{10}$ functions, the cost of IRRG is significantly different for their standard and modified versions. Since IRRG is insensitive to the modifications, it seems these differences were caused by the $\epsilon$ value estimation procedure.
	
	IRRG consumed the most FFEs for $F_{12}$, i.e., 7.4E+05 on average. Functions in $F_{12}$ are overlapping, and their subfunction sizes are equal to $2$. Thus, if two subfunctions overlap, there is one shared variable. IRRG tends to join all overlapping subproblems into one group. Therefore, to join two overlapping subfunctions, there is only one possible interaction between two variables that must be discovered. Since functions in $F_{12}$ consist of many overlapping subfunctions and discovering dependency between two variables is hard for IRRG in this case, many iterations of IRRG are necessary.
	
	The maximum number of IRRG iterations was reached for $\sqrt{f_{12}}$ and was equal to $169$ whereas only one iteration was needed for $F_1$-$F_3$. On average IRRG executed RRG $22.74 \pm 29.47$ times. After excluding the fully separable functions and $F_{12}$, which are special cases for IRRG, the average number of iterations is $19.22 \pm 4.44$. These average values are much smaller than considered problem sizes. Thus, the typical time complexity of IRRG may be found as $\mathcal{O}(n\log(n))$.
	
	\begin{table}
		\caption{FFEs consumed by decomposition strategies}
		\label{tab:ffes}
		\scriptsize
		\setlength{\tabcolsep}{0.3em}
		\begin{tabularx}{\columnwidth}{NRRIIIII}
			\toprule
			\multirow{2}{*}{Func} & \multicolumn{2}{>{\hsize=\dimexpr2\hsize+2\tabcolsep+\arrayrulewidth\relax}R}{RDG3} & \multicolumn{5}{>{\hsize=\dimexpr5\hsize+5\tabcolsep+\arrayrulewidth\relax}I}{IRRG (Avg) / RDG3 (Avg)} \\
			\cmidrule{2-8}
			& Avg & Std & Min & Max & Med & Avg & Std \\
			\midrule
			
			$f_1$--$f_3$ & 4.0E+03 & 1.7E+03 & 6.7 & 13.3 & 13.3 & 11.1 & 3.8 \\
			$f_4$--$f_7$ & 1.0E+04 & 1.1E+03 & 4.8 & 8.8 & 7.8 & 7.3 & 1.8 \\
			$f_8$--$f_{11}$ & 1.9E+04 & 2.5E+02 & \textbf{3.6} & 5.4 & 4.2 & 4.3 & 0.8 \\
			$f_{12}$--$f_{14}$ & 2.7E+04 & 1.9E+04 & 4.2 & \textbf{14.6} & 4.4 & 7.7 & 5.9 \\
			$f_{15}$ & 6.0E+03 & N/A & 4.7 & 4.7 & 4.7 & 4.7 & N/A \\
			
			\addlinespace[0.5em]
			
			$(f_1)^2$--$(f_3)^2$ & 6.3E+03 & 5.3E+02 & 5.8 & 6.7 & 6.7 & 6.4 & 0.5 \\
			$(f_4)^2$--$(f_7)^2$ & 8.1E+03 & 1.7E+03 & 5.6 & 12.7 & 10.0 & 9.6 & 3.1 \\
			$(f_8)^2$--$(f_{11})^2$ & 1.4E+04 & 5.4E+03 & \textbf{3.8} & 9.3 & 6.8 & 6.7 & 2.6 \\
			$(f_{12})^2$--$(f_{14})^2$ & 6.5E+03 & 1.5E+03 & 8.3 & \textbf{124.2} & 12.7 & 48.4 & 65.7 \\
			$(f_{15})^2$ & 6.0E+03 & N/A & 4.8 & 4.8 & 4.8 & 4.8 & N/A \\
			
			\addlinespace[0.5em]
			
			$\sqrt{f_1}$--$\sqrt{f_3}$ & 6.4E+03 & 7.9E+02 & 5.4 & 6.7 & 6.7 & 6.3 & 0.7 \\
			$\sqrt{f_4}$--$\sqrt{f_7}$ & 1.0E+04 & 3.1E+03 & 5.4 & 12.8 & 7.2 & 8.1 & 3.5 \\
			$\sqrt{f_8}$--$\sqrt{f_{11}}$ & 1.2E+04 & 5.0E+03 & \textbf{3.8} & 13.5 & 8.2 & 8.4 & 4.1 \\
			$\sqrt{f_{12}}$--$\sqrt{f_{14}}$ & 6.3E+03 & 1.1E+03 & 8.9 & \textbf{123.0} & 12.8 & 48.2 & 64.8 \\
			$\sqrt{f_{15}}$ & 6.0E+03 & N/A & 4.9 & 4.9 & 4.9 & 4.9 & N/A \\
			
			\bottomrule
		\end{tabularx}
	\end{table}
	
	\subsection{Optimization Results}
	
	Decomposition of a problem is only a part of the whole optimization process. Additionally, even the high-quality decomposition does not guarantee the successful optimization of a given problem. Optimization methods that can use such decomposition wisely are also needed~\cite{3lo}. For instance, CBCC and CCFR2 offer a speed-up of a CC framework by greedily choosing the most promising subproblem to optimize~\cite{cbcc, ccfr2, cc2018}. Another example is joining separable variables into groups and optimizing them by CMA-ES~\cite{rdg3}, which can effectively solve fully separable problems up to $100$ variables~\cite{sep-cmaes}. Thus, in this section, we compare the effectiveness of CBCC and CCFR2 after embedding IRRG and RDG3. Additionally, we also consider SHADE-ILS, which is dedicated to solving LSGO problems, but does not use problem decomposition. The observed differences in the results quality of the considered optimization methods are confirmed by the unpaired Wilcoxon test with the Holm-Bonferroni correction method at the significance level of $5\%$.
	
	\begin{table}
		\setlength{\tabcolsep}{0.5em}
		\caption{IRRG's number of wins, ties, and losses against RDG3 on the basis of CBCC and CCFR2}
		\label{tab:resIRRGvsRDG3}
		\scriptsize
		\begin{tabularx}{\columnwidth}{CCCCCCC}
			\toprule
			
			\multirow{3}{*}{$i$} & \multicolumn{3}{>{\hsize=\dimexpr3\hsize+3\tabcolsep+\arrayrulewidth\relax}C}{CBCC-IRRG vs. CBCC-RDG3}  & \multicolumn{3}{>{\hsize=\dimexpr3\hsize+3\tabcolsep+\arrayrulewidth\relax}C}{CCFR2-IRRG vs. CCFR2-RDG3} \\
			\cmidrule{2-7}
			& $f_i$ & $(f_i)^2$ & $\sqrt{f_i}$    & $f_i$ & $(f_i)^2$ & $\sqrt{f_i}$     \\
			& w/t/l & w/t/l & w/t/l & w/t/l & w/t/l & w/t/l \\
			\midrule
			
			1--3  & 1/\textbf{2}/0    & \textbf{3}/0/0   & \textbf{3}/0/0   & 1/\textbf{2}/0    & \textbf{3}/0/0   & \textbf{3}/0/0  \\
			4--7  & 0/\textbf{4}/0    & \textbf{3}/0/1   & \textbf{3}/1/0  & 0/\textbf{4}/0    & \textbf{3}/0/1   & \textbf{3}/1/0  \\
			8--11  & 0/\textbf{2}/\textbf{2}   & \textbf{4}/0/0   & \textbf{4}/0/0   & 1/\textbf{3}/0    & \textbf{3}/1/0   & \textbf{3}/1/0  \\
			12--14 & 1/0/\textbf{2}    & \textbf{2}/1/0   & 1/\textbf{2}/0   & 1/0/\textbf{2}    & \textbf{2}/1/0   & \textbf{3}/0/0  \\
			15   & 0/\textbf{1}/0    & 0/\textbf{1}/0   & 0/\textbf{1}/0   & 0/\textbf{1}/0    & 0/\textbf{1}/0   & 0/\textbf{1}/0  \\
			\midrule
			Total  & 2/\textbf{9}/4   & \textbf{12}/2/1 & \textbf{11}/4/0 & 3/\textbf{10}/2  & \textbf{11}/3/1 & \textbf{12}/3/0 \\
			\bottomrule
		\end{tabularx}
	\end{table}
	
	Table~\ref{tab:resIRRGvsRDG3} presents the comparison of the effectiveness of CBCC and CCFR2 after embedding IRRG and RDG3. For the standard CEC'2013 functions, the effectiveness of both considered frameworks is similar for both embedded decomposition strategies. The situation changes significantly when the modified CEC'2013 functions are considered. For both sets of modified CEC'2013 functions, CBCC-IRRG and CCFR2-IRRG outperform the RDG3-embedded CC frameworks significantly. Thus, we may state that embedding IRRG into the state-of-the-art CC frameworks is a more robust proposition than RDG3. Although IRRG consumes more FFEs than RDG3, its cost is still only a small part of the overall computational budget. Therefore, the higher cost of IRRG does not influence the quality of results for $f_1$--$f_{15}$.

	\begin{table}
		\setlength{\tabcolsep}{0.5em}
		\caption{CCFR2-IRRG's number of wins, ties, and losses against CBCC-IRRG and SHADE-ILS}
		\label{tab:resIRRGvsOther}
		\scriptsize
		\begin{tabularx}{\columnwidth}{CCCCCCC}
			\toprule
			
			\multirow{3}{*}{$i$} & \multicolumn{3}{>{\hsize=\dimexpr3\hsize+3\tabcolsep+\arrayrulewidth\relax}C}{CCFR2-IRRG vs. CBCC-IRRG}  & \multicolumn{3}{>{\hsize=\dimexpr3\hsize+3\tabcolsep+\arrayrulewidth\relax}C}{CCFR2-IRRG vs. SHADE-ILS} \\
			\cmidrule{2-7}
			& $f_i$ & $(f_i)^2$ & $\sqrt{f_i}$    & $f_i$ & $(f_i)^2$ & $\sqrt{f_i}$     \\
			& w/t/l & w/t/l & w/t/l & w/t/l & w/t/l & w/t/l \\
			\midrule
			1--3   & 0/\textbf{3}/$0$           & 0/\textbf{3}/0        & 0/\textbf{3}/0        & 0/0/\textbf{3}          & 0/0/\textbf{3}         & 0/1/\textbf{2}       \\
			4--7   & 0/\textbf{4}/0           & 0/\textbf{4}/0        & 0/\textbf{4}/0        & \textbf{3}/0/1          & \textbf{3}/0/1         & \textbf{3}/0/1        \\
			8--11  & \textbf{2}/\textbf{2}/0           & 1/\textbf{3}/0        & \textbf{2}/\textbf{2}/0        & \textbf{3}/1/0          & \textbf{3}/1/0         & \textbf{2}/\textbf{2}/0        \\
			12--14 & 0/\textbf{3}/0           & 0/\textbf{3}/0        & 0/\textbf{3}/0        & 0/1/\textbf{2}         & 0/0/\textbf{3}         & 0/1/\textbf{2}       \\
			15    & 0/\textbf{1}/0          & 0/\textbf{1}/0       & 0/\textbf{1}/0       & 0/0/\textbf{1}          & 0/0/\textbf{1}         & 0/0/\textbf{1}        \\
			\midrule
			Total   & 2/\textbf{13}/0          & 1/\textbf{14}/0       & 2/\textbf{13}/0       & 6/2/\textbf{7}         & 6/1/\textbf{8}       & 5/4/\textbf{6}  \\
			\bottomrule
		\end{tabularx}
	\end{table}
	
	In Table~\ref{tab:resIRRGvsOther}, we compare the effectiveness of CCFR2-IRRG to CBCC-IRRG and SHADE-ILS. The results show that the effectiveness of CCFR2-IRRG and CBCC-IRRG is highly similar. CCFR2-IRRG was chosen for the comparison to SHADE-ILS, because it performs slightly better. SHADE-ILS is significantly different from CC-based optimizers, and it does not rely on problem decomposition. As expected, for partially separable problems from the standard and modified CEC'2013 sets, CCFR2-IRRG outperforms SHADE-ILS significantly. However, the situation is the opposite for fully separable, overlapping, and non-separable functions. Such results are a consequence of the construction of CBCC and CCFR2. Additionally, SHADE-ILS employs MTS-L1, a local search optimizer dedicated to fully separable problems~\cite{mts,shade-ils}. MTS-L1 along with the restart mechanism seems to be the reason why SHADE-ILS has outperformed all CC-based optimizers for $F_1$--$F_3$. Note that CMA-ES (employed by all of the considered CC-based optimizers) is dedicated to optimizing problems with at most $100$ variables~\cite{sep-cmaes}, while all of the considered fully non-separable functions are high-dimensional. Finally, all functions in the CEC'2013 set are homogeneous, i.e., all their subfunctions derive from the same base function (e.g., Rastrigin's function)~\cite{cec2013}. Such a feature may favor SHADE-ILS that adjusts the parameter values of its underlying SHADE and the choice of a local search method (MTS-LS1 or L-BFGS-B) to fitness landscape characteristics. If considered problems are not homogeneous, then such problems should be more suitable for CC-based optimizers because they may adjust their behaviour to the subset of interacting variables not to the whole variable set.\par 
	
	Usually, when IRRG-embedded CC frameworks outperform SHADE-ILS, the order of magnitude of optimization result differences is higher when compared to the opposite situations. In supplementary material, we support an extended discussion on this issue and Fig. S-1 to picture it.
	
	\begin{table}
		\setlength{\tabcolsep}{0.5em}
		\caption{Results for the standard CEC'2013 set}
		\label{tab:resultsCec}
		\scriptsize
		\begin{tabularx}{\columnwidth}{FSBBBBB}
			\toprule
			Func & Stats & CBCC-IRRG & CCFR2-IRRG & CBCC-RDG3 & CCFR2-RDG3 & SHADE-ILS \\
			
			\midrule
			
			\multirow{3}{*}{$f_1$} & Med & 7.87E-19 & 6.78E-19 & 8.14E-19 & 7.64E-19 & \textbf{0.00E+00} \\
			& Avg & 9.16E-19 & 7.41E-19 & 8.57E-19 & 8.89E-19 & \textbf{2.40E-28} \\
			& Std & 3.35E-19 & 1.89E-19 & 2.77E-19 & 3.07E-19 & \textbf{5.10E-28} \\
			
			\addlinespace[0.5em]
			
			\multirow{3}{*}{$f_2$} & Med & 2.35E+03 & 2.35E+03 & 2.36E+03 & 2.33E+03 & \textbf{1.03E+03} \\
			& Avg & 2.34E+03 & 2.36E+03 & 2.34E+03 & 2.33E+03 & \textbf{1.06E+03} \\
			& Std & 9.64E+01 & 1.24E+02 & 9.56E+01 & 9.57E+01 & \textbf{1.37E+02} \\
			
			\addlinespace[0.5em]
			
			\multirow{3}{*}{$f_3$} & Med & 2.02E+01 & 2.02E+01 & 2.04E+01 & 2.04E+01 & \textbf{2.01E+01} \\
			& Avg & 2.02E+01 & 2.02E+01 & 2.04E+01 & 2.04E+01 & \textbf{2.01E+01} \\
			& Std & 1.07E-01 & 1.07E-01 & 7.48E-02 & 6.89E-02 & \textbf{1.13E-02} \\
			
			\addlinespace[0.5em]
			
			\multirow{3}{*}{$f_4$} & Med & \textbf{1.58E+04} & \textbf{2.08E+03} & \textbf{5.68E+03} & \textbf{8.74E+02} & 2.52E+08 \\
			& Avg & \textbf{1.97E+04} & \textbf{1.62E+04} & \textbf{1.45E+04} & \textbf{1.77E+04} & 2.54E+08 \\
			& Std & \textbf{2.00E+04} & \textbf{3.03E+04} & \textbf{2.54E+04} & \textbf{3.46E+04} & 9.98E+07 \\
			
			\addlinespace[0.5em]
			
			\multirow{3}{*}{$f_5$} & Med & 2.42E+06 & 2.22E+06 & 2.23E+06 & 2.09E+06 & \textbf{1.31E+06} \\
			& Avg & 2.32E+06 & 2.17E+06 & 2.31E+06 & 2.06E+06 & \textbf{1.29E+06} \\
			& Std & 4.97E+05 & 3.23E+05 & 3.64E+05 & 3.28E+05 & \textbf{2.04E+05} \\
			
			\addlinespace[0.5em]
			
			\multirow{3}{*}{$f_6$} & Med & \textbf{9.96E+05} & 9.96E+05 & \textbf{9.96E+05} & \textbf{9.96E+05} & 1.04E+06 \\
			& Avg & \textbf{1.00E+06} & 1.01E+06 & \textbf{9.99E+05} & \textbf{1.00E+06} & 1.04E+06 \\
			& Std & \textbf{2.27E+04} & 2.27E+04 & \textbf{1.34E+04} & \textbf{2.25E+04} & 5.64E+03 \\
			
			\addlinespace[0.5em]
			
			\multirow{3}{*}{$f_7$} & Med & \textbf{9.43E-22} & \textbf{9.25E-22} & \textbf{9.18E-22} & \textbf{9.19E-22} & 1.07E+02 \\
			& Avg & \textbf{9.33E-22} & \textbf{9.27E-22} & \textbf{9.38E-22} & \textbf{9.25E-22} & 1.18E+03 \\
			& Std & \textbf{9.62E-23} & \textbf{7.24E-23} & \textbf{8.06E-23} & \textbf{6.11E-23} & 4.94E+03 \\
			
			\addlinespace[0.5em]
			
			\multirow{3}{*}{$f_8$} & Med & 9.13E+03 & \textbf{3.50E-05} & 5.91E+03 & 3.48E+03 & 8.65E+11 \\
			& Avg & 9.12E+03 & \textbf{4.47E-05} & 7.58E+03 & 3.41E+03 & 9.32E+11 \\
			& Std & 1.37E+03 & \textbf{2.99E-05} & 5.23E+03 & 1.40E+03 & 5.80E+11 \\
			
			\addlinespace[0.5em]
			
			\multirow{3}{*}{$f_9$} & Med & \textbf{1.59E+08} & \textbf{1.50E+08} & \textbf{1.51E+08} & \textbf{1.65E+08} & \textbf{1.68E+08} \\
			& Avg & \textbf{1.64E+08} & \textbf{1.49E+08} & \textbf{1.56E+08} & \textbf{1.67E+08} & \textbf{1.64E+08} \\
			& Std & \textbf{2.76E+07} & \textbf{3.19E+07} & \textbf{3.05E+07} & \textbf{3.23E+07} & \textbf{2.31E+07} \\
			
			\addlinespace[0.5em]
			
			\multirow{3}{*}{$f_{10}$} & Med & \textbf{9.05E+07} & \textbf{9.05E+07} & \textbf{9.05E+07} & \textbf{9.05E+07} & 9.28E+07 \\
			& Avg & \textbf{9.10E+07} & \textbf{9.17E+07} & \textbf{9.12E+07} & \textbf{9.23E+07} & 9.27E+07 \\
			& Std & \textbf{1.27E+06} & \textbf{1.88E+06} & \textbf{1.54E+06} & \textbf{2.08E+06} & 4.27E+05 \\
			
			\addlinespace[0.5em]
			
			\multirow{3}{*}{$f_{11}$} & Med & 5.49E-12 & \textbf{6.70E-18} & 9.64E-15 & \textbf{7.11E-18} & 4.99E+05 \\
			& Avg & 1.01E-10 & \textbf{1.73E-17} & 2.47E-12 & \textbf{7.13E-18 }& 5.11E+05 \\
			& Std & 1.56E-10 & \textbf{3.36E-17} & 9.11E-12 & \textbf{3.47E-18} & 1.37E+05 \\
			
			\addlinespace[0.5em]
			
			\multirow{3}{*}{$f_{12}$} & Med & 8.12E+02 & 9.04E+02 & 7.24E+02 & 7.88E+02 & \textbf{6.46E-01} \\
			& Avg & 2.32E+03 & 9.06E+02 & 7.10E+02 & 7.98E+02 & \textbf{6.60E+01} \\
			& Std & 7.17E+03 & 7.18E+01 & 9.01E+01 & 7.76E+01 & \textbf{2.38E+02} \\
			
			\addlinespace[0.5em]
			
			\multirow{3}{*}{$f_{13}$} & Med & 1.34E+06 & 1.50E+06 & 4.65E+04 & \textbf{1.75E+04} & 1.04E+06 \\
			& Avg & 1.14E+06 & 1.30E+06 & 5.02E+04 & \textbf{2.43E+04} & 1.13E+06 \\
			& Std & 5.61E+05 & 5.80E+05 & 2.40E+04 & \textbf{1.66E+04} & 1.00E+06 \\
			
			\addlinespace[0.5em]
			
			\multirow{3}{*}{$f_{14}$} & Med & 1.60E+07 & 1.53E+07 & 1.47E+09 & 2.15E+09 & \textbf{7.55E+06} \\
			& Avg & 1.89E+07 & 2.01E+07 & 1.82E+09 & 2.38E+09 & \textbf{7.69E+06} \\
			& Std & 8.75E+06 & 1.14E+07 & 1.52E+09 & 1.70E+09 & \textbf{1.13E+06} \\
			
			\addlinespace[0.5em]
			
			\multirow{3}{*}{$f_{15}$} & Med & 2.22E+06 & 2.27E+06 & 2.20E+06 & 2.18E+06 & \textbf{6.10E+05} \\
			& Avg & 2.21E+06 & 2.29E+06 & 2.19E+06 & 2.27E+06 & \textbf{7.88E+05} \\
			& Std & 2.35E+05 & 1.91E+05 & 1.72E+05 & 2.52E+05 & \textbf{7.62E+05} \\
			\bottomrule
		\end{tabularx}
	\end{table}
	
	In Table~\ref{tab:resultsCec}, we present the results for the standard CEC'2013 functions. These detailed results reveal some phenomena discussed below. According to Table \ref{tab:accuracy}, only IRRG decomposes $f_8$ precisely. Therefore, CCFR2-IRRG produced about $10^8$ and $10^{16}$ times better results than RDG3-embedded CCs and SHADE-ILS, respectively. However, CBCC-IRRG proposed results of quality that was similar to RDG3-embedded CCs. The reason seems to be as follows. CBCC and CCFR2 focus on optimizing those components that have the highest impact on the global fitness. However, in CCFR2, the component's initial relatively poor contribution becomes insignificant faster. Thus, CCFR2 may spend more time improving those initially less significant components. This example shows that even having the perfect decomposition of a partially separable problem may not be enough to reach the high-quality results. The differences between CBCC and CCFR2 are also visible for $f_{11}$, for which both CCFR2 versions outperform the CBCC versions.
	
	For overlapping problems, the perfect decomposition may not exist~\cite{dg2,3lo}. Therefore, the structure of the proposed groups influenced the quality of the final results the most significantly. $f_{13}$ and $f_{14}$ are defined as overlapping problems with conforming and conflicting subproblems, respectively~\cite{lsgoOverlapping}. When subproblems are conforming, shared variables (variables that belong to more than one subproblem) may be optimized for each subproblem separately, i.e., the optimal value of a shared variable is the same for all subproblems the shared variable belongs to. If subproblems are conflicting, optimizing one subproblem collides with the optimization of other subproblems that share the same variables. Note that the considered CC-based optimizers process disjoint components and each shared variable must belong to only one component. Therefore, if subproblems are conforming, it could be better to use smaller components like optimizing each subproblem separately. If subproblems contradict, then it may be better to use larger components to optimize many conflicting subproblems at the same time. In general, RDG3 tends to create more groups containing fewer variables than IRRG due to the $\epsilon_n$ threshold. Therefore, RDG3 seems suitable for $f_{13}$ (RGD3-embedded CCs report approximately $100$ times better results than IRRG-embedded ones). The situation is opposite for $f_{14}$. Finally, for $f_{12}$, which contains neither conforming nor contradicting subproblems, both variable grouping ways seem equally useful. Thus, the differences in the optimization results for $f_{13}$ and $f_{14}$ arise from the differences of the underlying problem structure and the employed grouping manner.
	
	In Tables S-I and S-II (in the supplementary material, Section S-II), we report the results for the modified CEC'2013 sets. As expected, for these functions, RDG3-embedded CCs are significantly less effective when compared to IRRG-embedded ones. The results remain similar for $(f_{15})^2$ and $\sqrt{f_{15}}$, which is expected because these functions are fully non-separable. RDG3-embedded CCs are no longer dominating for $(f_{13})^2$ and $\sqrt{f_{13}}$, because RDG3 detects more (false) dependencies. Therefore, it proposes larger groups, which may not be advantageous for overlapping problems with conforming subproblems. However, despite increasing the size of the groups, RDG3-embedded CCs do not improve their effectiveness for $(f_{14})^2$ and $\sqrt{f_{14}}$ containing conflicting subfunctions. Indeed, RDG3 proposes larger groups for these functions, but since they contain false linkage, these groups mix variables from different subfunctions. Such low-quality groups do not improve the quality of the search.
	
	Functions from $F_5$, i.e., $f_5$, $(f_5)^2$, and $\sqrt{f_5}$, contain separable variables. However, differently to other functions of this type, the quality of the results reported for $F_5$ is similar for all IRRG- and RDG3-embedded CCs. Based on the reported results, it seems that for $F_5$, the influence on the fitness of separable variables is significantly higher than for the other partially separable functions with separable subproblems ($F_4$, $F_6$, $F_7$). Therefore, the quality of the decomposition does not play an important role in the optimization of $f_5$, $(f_5)^2$, and $\sqrt{f_5}$. Note that from all the considered optimization methods, SHADE-ILS is the most effective in optimizing fully separable problems. Hence, SHADE-ILS outperformed the others for $F_5$.
	
	
	%

	

	
	\section{Conclusion}
	
	In this article, we propose a new decomposition strategy, namely IRRG, which derives from monotonicity checking, never reports false linkage, and significantly mitigates the risk of missing linkage. Although IRRG is more expensive than RDG3, its typical time complexity is also $\mathcal{O}(n\log(n))$. Additionally, the FFE cost of IRRG is low when compared to the overall optimization cost. IRRG- and RDG3-embedded CCs report similar quality results for the standard CEC'2013 functions, which are largely suitable for RDG3. However, for both proposed modified CEC'2013 sets with functions with non-additively separable subfunctions, the effectiveness of IRRG-embedded CCs remains the same, while for RDG3-embedded CCs deteriorates significantly. Moreover, IRRG-embedded CCs outperform RDG3-embedded CCs for the considered real-world optimization problem with non-additively separable subproblems\footnote{See the supplementary material.}.
	
	The comparison of IRRG-embedded CCs and SHADE-ILS shows that CCs may choose different component optimizers depending on the component size and the existence of separable variables. Another promising future work direction is adjusting the $\epsilon_{sti}$ value during the IRRG run and further improvements in the procedure of automatic $\epsilon$ estimation. Finally, proposing new benchmark sets with non-homogeneous functions may be helpful for future research.
	
	\section*{Acknowledgment}
	
	The authors wish to express their gratitude to Roza Goscien for valuable discussions on real-world optimization problems.
	
	\ifCLASSOPTIONcaptionsoff
	\newpage
	\fi

	
	
	%
	
	
	
	\bibliographystyle{IEEEtran}
	\bibliography{bibl}
	
	%
	
	
	
	
	
	
	
	
	
\end{document}


%
\title{Supplementary Document of 'Incremental Recursive Ranking Grouping for Large Scale Global Optimization'}
%
%
%

\author{Marcin~Michal~Komarnicki, Michal~Witold~Przewozniczek, Halina~Kwasnicka,~and~Krzysztof~Walkowiak}
\maketitle



%
\IEEEpeerreviewmaketitle

\section{Non-Additive Separability}

Let $f: \Omega \rightarrow \mathbb R$ be a function that consists of $m$ separable subfunctions such that $f_i: \Omega_i \rightarrow \mathbb R$ for $i \in \{1, ..., m\}$. Subfunctions $f_i$ and $f_j$ are additively separable if for any $x_p$ and $x_q$
\begin{equation}
    \frac{\partial^2 f}{\partial x_q \partial x_p } = 0
\end{equation}
where $x_p$ and $x_q$ are arguments of subfunctions $f_i$ and $f_j$, respectively~\cite{rdg}. Otherwise, $f_i$ and $f_j$ are non-additively separable.

Let us consider a function $\bar{f}_{c,4}: [-5, 5]^4 \rightarrow [0, 10\sqrt{2}]$ defined as $\bar{f}_{c,4}(\vec{x}) = \sqrt{x_1^2 + x_2^2} + \sqrt{x_3^2 + x_4^2}$. This function is fully separable (according to both separability definitions considered in this paper\textemdash the ELL-based and the one defined in formula (1) in the main paper) and the following subfunctions can be optimized separately to obtain the optimal solution, i.e., $[0, 0, 0, 0]$:
\begin{enumerate}
    \item $\bar{f}_{c,4,1}(x_1) = x_1^2$,
    \item $\bar{f}_{c,4,2}(x_2) = x_2^2$,
    \item $\bar{f}_{c,4,3}(x_3) = x_3^2$,
    \item $\bar{f}_{c,4,4}(x_4) = x_4^2$.
\end{enumerate}
Subfunctions $\bar{f}_{c,4,1}$ and $\bar{f}_{c,4,3}$ are additively separable, because
\begin{equation}
    \frac{\partial^2 \bar{f}_{c,4}}{\partial x_3 \partial x_1} = \frac{\partial}{\partial x_3} \bigg(\frac{x_1}{\sqrt{x_1^2 + x_2^2}}\bigg) = 0
\end{equation}
On the other hand, subfunctions $\bar{f}_{c,4,1}$ and $\bar{f}_{c,4,2}$ are non-additively separable, because
\begin{equation}
    \frac{\partial^2 \bar{f}_{c,4}}{\partial x_2 \partial x_1} = \frac{\partial}{\partial x_2} \bigg(\frac{x_1}{\sqrt{x_1^2 + x_2^2}}\bigg) = -\frac{x_1 x_2}{(x_1^2 + x_2^2)^{\frac{3}{2}}} \ne 0
\end{equation}

An example of a non-additively separable function, which is commonly used as a benchmark function, is the Ackley's function. The $n$-dimensional Ackley's function is defined as follows
\begin{equation}
    \begin{aligned}
        f_{ackley} &= -20 \exp{\Bigg(-0.2 \sqrt{\frac{1}{n} \sum_{i=1}^n x_i^2}\Bigg)} \\
                   &- \exp{\bigg(\frac{1}{n} \sum_{i=1}^n \cos(2 \pi x_i)\bigg)} \\
                   &+ 20 + e
    \end{aligned}
\end{equation}
Since the Ackley's function is fully separable~\cite{cec2013}, it can be decomposed into $n$ separable subfunctions $f_{ackley, 1}$, ..., $f_{ackley, n}$. Any pair of these separable subfunctions, $f_{ackley, i}$ and $f_{ackley, j}$, is non-additively separable, because
\begin{equation}
    \small
    \begin{aligned}
        \frac{\partial^2 f_{ackley}}{\partial x_q \partial x_p} &= \\
        & \frac{x_p x_q \exp{\big(-0.2 \sqrt{\frac{1}{n} \sum_{k=1}^n x_k^2} \big)} (-\frac{0.8}{n} \sqrt{\sum_{k=1}^n x_k^2} - \frac{4}{\sqrt{n}}) }{(\sum_{k=1}^n x_k^2)^{\frac{3}{2}}} \\
        &- \\
        & \Big(\frac{2 \pi}{n}\Big)^2 \sin(2 \pi x_p) \sin(2 \pi x_q) \exp{\bigg(\frac{1}{n} \sum_{k=1}^n \cos(2 \pi x_k) \bigg)} \\
        &\ne 0
    \end{aligned}
\end{equation}
where $x_p$ and $x_q$ are single arguments of subfunctions $f_{ackley, i}$ and $f_{ackley, j}$, respectively.

\section{Additional Results}

Tables~\ref{tab:resultsSquareCEC} and~\ref{tab:resultsSqrtCEC} present the detailed optimization results for the modified CEC'2013 sets.

\begin{table}
\setlength{\tabcolsep}{0.5em}
 \caption{Results for the squared CEC'2013 set}
\label{tab:resultsSquareCEC}
\scriptsize
\begin{tabularx}{\columnwidth}{FSBBBBB}
\toprule
 Func & Stats & CBCC-IRRG & CCFR2-IRRG & CBCC-RDG3 & CCFR2-RDG3 & SHADE-ILS \\
\midrule

\multirow{3}{*}{$(f_{1})^2$} & Med & 5.75E-37 & 4.90E-37 & 1.11E+06 & 1.41E+06 & \textbf{0.00E+00}\\
& Avg & 9.18E-37 & 8.10E-37 & 7.76E+06 & 7.69E+06 & \textbf{1.46E-53}\\
& Std & 6.67E-37 & 6.99E-37 & 1.15E+07 & 1.38E+07 & \textbf{6.13E-53}\\

\addlinespace[0.5em]

\multirow{3}{*}{$(f_{2})^2$} & Med & 5.25E+06 & 5.53E+06 & 2.23E+07 & 2.29E+07 & \textbf{1.12E+06}\\
& Avg & 5.34E+06 & 5.53E+06 & 2.22E+07 & 2.28E+07 & \textbf{1.19E+06}\\
& Std & 5.25E+05 & 5.09E+05 & 1.81E+06 & 2.12E+06 & \textbf{2.57E+05}\\

\addlinespace[0.5em]

\multirow{3}{*}{$(f_{3})^2$} & Med & \textbf{4.07E+02} & 4.07E+02 & 4.15E+02 & 4.16E+02 & \textbf{4.02E+02}\\
& Avg & \textbf{4.06E+02} & 4.08E+02 & 4.15E+02 & 4.16E+02 & \textbf{4.02E+02}\\
& Std & \textbf{5.12E+00} & 3.74E+00 & 3.07E+00 & 2.13E+00 & \textbf{5.73E-01}\\

\addlinespace[0.5em]

\multirow{3}{*}{$(f_{4})^2$} & Med & \textbf{1.53E+07} & \textbf{3.72E+07} & 2.99E+17 & 3.46E+17 & 6.55E+16\\
& Avg & \textbf{9.45E+08} & \textbf{3.88E+08} & 2.98E+17 & 3.55E+17 & 9.06E+16\\
& Std & \textbf{3.52E+09} & \textbf{5.12E+08} & 5.99E+16 & 7.44E+16 & 7.37E+16\\

\addlinespace[0.5em]

\multirow{3}{*}{$(f_{5})^2$} & Med & 5.18E+12 & 4.24E+12 & 3.07E+12 & 2.90E+12 & \textbf{1.66E+12}\\
& Avg & 5.44E+12 & 4.46E+12 & 3.23E+12 & 2.99E+12 & \textbf{1.88E+12}\\
& Std & 1.84E+12 & 1.46E+12 & 1.05E+12 & 8.07E+11 & \textbf{5.84E+11}\\

\addlinespace[0.5em]

\multirow{3}{*}{$(f_{6})^2$} & Med & \textbf{9.92E+11} & \textbf{9.92E+11} & 1.01E+12 & 1.01E+12 & 1.07E+12
\\
& Avg & \textbf{1.00E+12} & \textbf{1.01E+12} & 1.01E+12 & 1.01E+12 & 1.07E+12\\
& Std & \textbf{3.83E+10} & \textbf{4.59E+10} & 5.31E+09 & 6.37E+09 & 1.49E+10\\

\addlinespace[0.5em]

\multirow{3}{*}{$(f_{7})^2$} & Med & \textbf{8.08E-43} & \textbf{8.59E-43} & 5.71E+10 & 1.16E+01 & 2.82E+04\\
& Avg & \textbf{8.15E-43} & \textbf{8.89E-43} & 5.85E+10 & 1.14E+01 & 1.16E+05\\
& Std & \textbf{1.54E-43} & \textbf{1.46E-43} & 7.10E+09 & 5.94E+00 & 1.93E+05\\

\addlinespace[0.5em]

\multirow{3}{*}{$(f_{8})^2$} & Med & 7.66E+07 & \textbf{4.85E-07} & 1.39E+19 & 2.38E+19 & 1.80E+23\\
& Avg & 7.88E+07 & \textbf{1.20E-06} & 9.90E+19 & 1.45E+21 & 3.90E+23\\
& Std & 2.63E+07 & \textbf{2.16E-06} & 2.35E+20 & 4.60E+21 & 7.28E+23\\

\addlinespace[0.5em]

\multirow{3}{*}{$(f_{9})^2$} & Med & \textbf{2.81E+16} & \textbf{2.46E+16} & 4.17E+16 & 3.48E+16 & \textbf{2.73E+16}\\
& Avg & \textbf{2.65E+16} & \textbf{2.56E+16} & 4.13E+16 & 3.84E+16 & \textbf{2.76E+16}\\
& Std & \textbf{9.74E+15} & \textbf{8.92E+15} & 1.22E+16 & 1.01E+16 & \textbf{5.18E+15}\\

\addlinespace[0.5em]

\multirow{3}{*}{$(f_{10})^2$} & Med & \textbf{8.20E+15} & \textbf{8.20E+15} & 8.27E+15 & 8.28E+15 & 8.60E+15\\
& Avg & \textbf{8.32E+15} & \textbf{8.36E+15} & 8.26E+15 & 8.31E+15 & 8.59E+15\\
& Std & \textbf{2.75E+14} & \textbf{3.19E+14} & 5.90E+13 & 1.91E+14 & 7.88E+13\\

\addlinespace[0.5em]

\multirow{3}{*}{$(f_{11})^2$} & Med & \textbf{2.50E-27} & \textbf{2.51E-26} & 9.78E+13 & 1.07E+14 & 2.80E+11\\
& Avg & \textbf{2.35E-20} & \textbf{1.29E-24} & 1.04E+14 & 1.06E+14 & 3.21E+11\\
& Std & \textbf{7.68E-20} & \textbf{4.28E-24} & 2.50E+13 & 2.10E+13 & 1.85E+11\\

\addlinespace[0.5em]

\multirow{3}{*}{$(f_{12})^2$} & Med & 8.29E+05 & 7.97E+05 & 9.95E+05 & 1.01E+06 & \textbf{6.16E+04}\\
& Avg & 7.95E+22 & 8.06E+05 & 9.29E+05 & 9.32E+05 & \textbf{1.75E+05}\\
& Std & 2.27E+23 & 1.01E+05 & 1.40E+05 & 1.63E+05 & \textbf{2.19E+05}\\

\addlinespace[0.5em]

\multirow{3}{*}{$(f_{13})^2$} & Med & 7.70E+11 & 1.84E+12 & 2.13E+12 & 2.04E+12 & \textbf{9.39E+10}\\
& Avg & 1.31E+12 & 1.57E+12 & 2.25E+12 & 2.07E+12 & \textbf{9.32E+11}\\
& Std & 1.09E+12 & 1.01E+12 & 6.00E+11 & 4.81E+11 & \textbf{1.60E+12}\\

\addlinespace[0.5em]

\multirow{3}{*}{$(f_{14})^2$} & Med & 3.64E+14 & 7.11E+14 & 1.71E+19 & 1.47E+18 & \textbf{5.33E+13}\\
& Avg & 4.83E+14 & 5.85E+15 & 1.82E+19 & 3.85E+18 & \textbf{5.98E+13}\\
& Std & 3.08E+14 & 2.65E+16 & 1.33E+19 & 5.88E+18 & \textbf{1.81E+13}\\

\addlinespace[0.5em]

\multirow{3}{*}{$(f_{15})^2$} & Med & 5.28E+12 & 5.12E+12 & 4.94E+12 & 4.99E+12 & \textbf{3.25E+11}\\
& Avg & 5.36E+12 & 5.23E+12 & 4.87E+12 & 5.10E+12 & \textbf{1.49E+12}\\
& Std & 1.10E+12 & 1.03E+12 & 8.43E+11 & 9.27E+11 & \textbf{2.99E+12}\\
\bottomrule
\end{tabularx}
\end{table}

\begin{table}
\setlength{\tabcolsep}{0.5em}
 \caption{Results for the square rooted CEC'2013 set}
\label{tab:resultsSqrtCEC}
\scriptsize
\begin{tabularx}{\columnwidth}{FSBBBBB}
\toprule
 Func & Stats & CBCC-IRRG & CCFR2-IRRG & CBCC-RDG3 & CCFR2-RDG3 & SHADE-ILS \\
\midrule

\multirow{3}{*}{$\sqrt{f_{1}}$} & Med & 8.86E-10 & 8.78E-10 & 2.71E-02 & 1.72E-03 & \textbf{0.00E+00}\\
& Avg & 9.20E-10 & 9.16E-10 & 2.83E-02 & 1.88E-03 & \textbf{2.52E-14}\\
& Std & 1.75E-10 & 1.35E-10 & 1.67E-02 & 1.01E-03 & \textbf{5.91E-14}\\

\addlinespace[0.5em]

\multirow{3}{*}{$\sqrt{f_{2}}$} & Med & 4.83E+01 & 4.83E+01 & 6.92E+01 & 6.84E+01 & \textbf{3.19E+01}\\
& Avg & 4.84E+01 & 4.85E+01 & 6.90E+01 & 6.83E+01 & \textbf{3.21E+01}\\
& Std & 9.51E-01 & 1.00E+00 & 1.55E+00 & 1.31E+00 & \textbf{1.32E+00}\\

\addlinespace[0.5em]

\multirow{3}{*}{$\sqrt{f_{3}}$} & Med & 4.49E+00 & \textbf{4.49E+00} & 4.52E+00 & 4.52E+00 & \textbf{4.48E+00}\\
& Avg & 4.49E+00 & \textbf{4.49E+00} & 4.52E+00 & 4.52E+00 & \textbf{4.48E+00}\\
& Std & 1.26E-02 & \textbf{1.20E-02} & 5.65E-03 & 5.28E-03 & \textbf{1.43E-03}\\

\addlinespace[0.5em]

\multirow{3}{*}{$\sqrt{f_{4}}$} & Med & \textbf{1.07E+02} & \textbf{9.68E+01} & 1.77E+04 & 6.68E+03 & 1.58E+04\\
& Avg & \textbf{1.18E+02} & \textbf{1.05E+02} & 1.51E+04 & 6.82E+03 & 1.63E+04\\
& Std & \textbf{6.10E+01} & \textbf{9.66E+01} & 4.15E+03 & 5.65E+02 & 2.90E+03\\

\addlinespace[0.5em]

\multirow{3}{*}{$\sqrt{f_{5}}$} & Med & 1.51E+03 & 1.47E+03 & 1.43E+03 & 1.45E+03 & \textbf{1.18E+03}\\
& Avg & 1.49E+03 & 1.45E+03 & 1.43E+03 & 1.42E+03 & \textbf{1.16E+03}\\
& Std & 1.25E+02 & 1.47E+02 & 1.21E+02 & 1.21E+02 & \textbf{9.35E+01}\\

\addlinespace[0.5em]

\multirow{3}{*}{$\sqrt{f_{6}}$} & Med & \textbf{9.98E+02} & \textbf{9.98E+02} & 1.00E+03 & 1.02E+03 & 1.02E+03\\
& Avg & \textbf{1.00E+03} & \textbf{1.00E+03} & 1.01E+03 & 1.01E+03 & 1.02E+03\\
& Std & \textbf{9.22E+00} & \textbf{1.36E+01} & 1.09E+01 & 1.21E+01 & 3.42E+00\\

\addlinespace[0.5em]

\multirow{3}{*}{$\sqrt{f_{7}}$} & Med & \textbf{3.00E-11} & \textbf{2.99E-11} & 4.83E+02 & 7.67E-01 & 1.04E+01\\
& Avg & \textbf{3.01E-11} & \textbf{2.99E-11} & 4.82E+02 & 8.07E-01 & 1.39E+01\\
& Std & \textbf{1.11E-12} & \textbf{1.48E-12} & 1.10E+01 & 1.75E-01 & 9.68E+00\\

\addlinespace[0.5em]

\multirow{3}{*}{$\sqrt{f_{8}}$} & Med & 9.95E+01 & \textbf{5.46E-03} & 3.84E+06 & 4.40E+06 & 1.10E+06\\
& Avg & 9.97E+01 & \textbf{7.77E-03} & 4.56E+06 & 4.45E+06 & 1.02E+06\\
& Std & 9.71E+00 & \textbf{7.70E-03} & 1.89E+06 & 7.34E+05 & 3.66E+05\\

\addlinespace[0.5em]

\multirow{3}{*}{$\sqrt{f_{9}}$} & Med & \textbf{1.28E+04} & \textbf{1.25E+04} & 1.48E+04 & 1.48E+04 & \textbf{1.25E+04}\\
& Avg & \textbf{1.25E+04} & \textbf{1.24E+04} & 1.51E+04 & 1.47E+04 & \textbf{1.25E+04}\\
& Std & \textbf{1.03E+03} & \textbf{1.22E+03} & 2.44E+03 & 7.60E+02 & \textbf{6.39E+02}\\

\addlinespace[0.5em]

\multirow{3}{*}{$\sqrt{f_{10}}$} & Med & \textbf{9.52E+03} & \textbf{9.52E+03} & 9.53E+03 & 9.53E+03 & 9.63E+03\\
& Avg & \textbf{9.55E+03} & \textbf{9.59E+03} & 9.53E+03 & 9.55E+03 & 9.62E+03\\
& Std & \textbf{7.85E+01} & \textbf{1.03E+02} & 1.32E+01 & 6.65E+01 & 2.32E+01\\

\addlinespace[0.5em]

\multirow{3}{*}{$\sqrt{f_{11}}$} & Med & 4.81E-07 & \textbf{2.74E-09} & 2.82E+03 & 2.79E+03 & 6.76E+02\\
& Avg & 8.93E-06 & \textbf{2.68E-09} & 2.83E+03 & 2.77E+03 & 7.00E+02\\
& Std & 1.27E-05 & \textbf{7.49E-10} & 1.43E+02 & 1.66E+02 & 1.31E+02\\

\addlinespace[0.5em]

\multirow{3}{*}{$\sqrt{f_{12}}$} & Med & 3.05E+01 & 3.00E+01 & 3.16E+01 & 3.17E+01 & \textbf{1.12E-04}\\
& Avg & 1.66E+05 & 3.02E+01 & 3.09E+01 & 3.12E+01 & \textbf{1.50E+00}\\
& Std & 3.39E+05 & 1.07E+00 & 1.38E+00 & 1.16E+00 & \textbf{3.35E+00}\\

\addlinespace[0.5em]

\multirow{3}{*}{$\sqrt{f_{13}}$} & Med & 1.23E+03 & \textbf{8.38E+02} & 1.18E+03 & 1.25E+03 & \textbf{7.65E+02}\\
& Avg & 1.08E+03 & \textbf{9.56E+02} & 1.20E+03 & 1.23E+03 & \textbf{7.46E+02}\\
& Std & 2.63E+02 & \textbf{3.07E+02} & 6.48E+01 & 8.42E+01 & \textbf{2.31E+02}\\

\addlinespace[0.5em]

\multirow{3}{*}{$\sqrt{f_{14}}$} & Med & 4.20E+03 & 5.13E+03 & 4.14E+04 & 6.10E+04 & \textbf{2.67E+03}\\
& Avg & 4.34E+03 & 4.64E+03 & 4.20E+04 & 5.93E+04 & \textbf{2.70E+03}\\
& Std & 9.63E+02 & 9.36E+02 & 1.58E+04 & 2.97E+04 & \textbf{1.77E+02}\\

\addlinespace[0.5em]

\multirow{3}{*}{$\sqrt{f_{15}}$} & Med & 1.51E+03 & 1.49E+03 & 1.51E+03 & 1.49E+03 & \textbf{6.66E+02}\\
& Avg & 1.52E+03 & 1.49E+03 & 1.49E+03 & 1.50E+03 & \textbf{6.87E+02}\\
& Std & 7.99E+01 & 6.46E+01 & 7.31E+01 & 7.98E+01 & \textbf{2.85E+02}\\
\bottomrule
\end{tabularx}
\end{table}

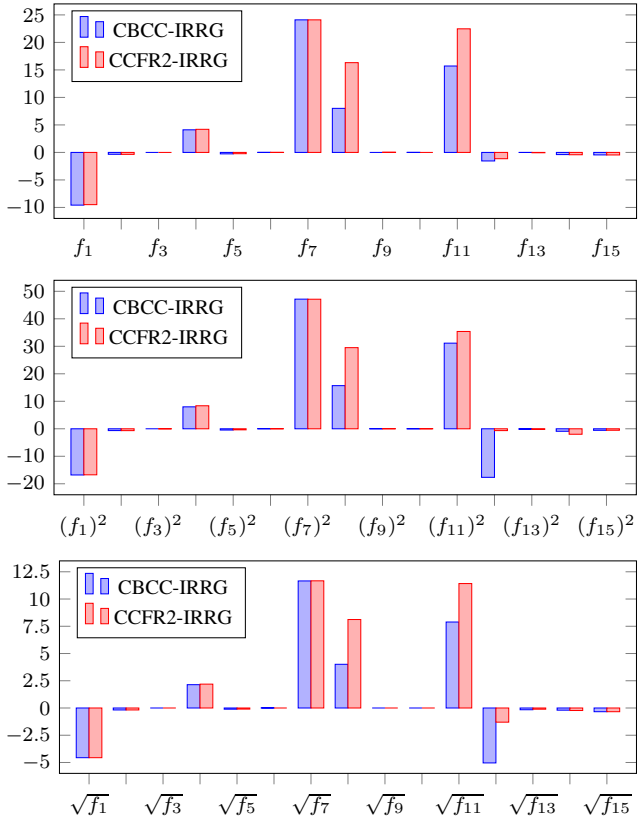
\begin{figure}
    \centering
	\subfloat{
		\resizebox{0.974\linewidth}{!}{%
			\begin{tikzpicture}
                \begin{axis} [
                    ybar = 0cm,
                    bar width = 0.5em,
                    xmin = 1,
                    xmax = 15,
                    ymin = -12,
                    ymax = 27,
                    xticklabels = {$f_{1}$, , $f_{3}$, , $f_{5}$, , $f_{7}$, , $f_{9}$, , $f_{11}$, , $f_{13}$, , $f_{15}$},
                    xtick = data,
                    ytick = {-10, -5, 0, 5, 10, 15, 20, 25},
                    enlarge x limits = {abs = .8},
                    width = 9.5cm,
                    height = 4.5cm,
                    legend pos=north west,
                    legend style={font=\fontsize{8}{0}\selectfont},
                    xticklabel style ={font=\fontsize{8}{0}\selectfont},
                    yticklabel style ={font=\fontsize{8}{0}\selectfont},
                    xtick pos=left
                ]
                 
                \addplot coordinates {(1,-9.58) (2,-0.34) (3,0) (4,4.11) (5,-0.26) (6, 0.01) (7, 24.10) (8,8.01) (9,0) (10, 0.01) (11, 15.71) (12, -1.55) (13, -0.01) (14, -0.39) (15, -0.45)};
                \addplot coordinates {(1,-9.49) (2,-0.35) (3,0) (4,4.20) (5,-0.23) (6, 0.01) (7, 24.10) (8,16.32) (9,0.04) (10, 0) (11, 22.47) (12, -1.14) (13, -0.06) (14, -0.42) (15, -0.46)};

                \legend {CBCC-IRRG, CCFR2-IRRG};
                \end{axis}
            \end{tikzpicture}
		}
	}
	\hfil
	\\[4pt]
	\subfloat{
		\resizebox{0.985\linewidth}{!}{%
			\begin{tikzpicture}
                \begin{axis} [
                    ybar = 0cm,
                    bar width = 0.5em,
                    xmin = 1,
                    xmax = 15,
                    ymin = -24,
                    ymax = 54,
                    xticklabels = {$(f_{1})^2$, , $(f_{3})^2$, , $(f_{5})^2$, , $(f_{7})^2$, , $(f_{9})^2$, , $(f_{11})^2$, , $(f_{13})^2$, , $(f_{15})^2$},
                    xtick = data,
                    ytick = {-20, -10, 0, 10, 20, 30, 40, 50},
                    enlarge x limits = {abs = .8},
                    width = 9.5cm,
                    height = 4.5cm,
                    legend pos=north west,
                    legend style={font=\fontsize{8}{0}\selectfont},
                    xticklabel style ={font=\fontsize{8}{0}\selectfont},
                    yticklabel style ={font=\fontsize{8}{0}\selectfont},
                    xtick pos=left,
                ]
                 
                \addplot coordinates {(1,-16.8) (2,-0.65) (3,0) (4,7.98) (5,-0.46) (6, 0.03) (7, 47.15) (8,15.69) (9,0.02) (10, 0.01) (11, 31.14) (12, -17.66) (13, -0.15) (14, -0.91) (15, -0.56)};
                \addplot coordinates {(1,-16.74) (2,-0.67) (3,-0.01) (4,8.37) (5,-0.38) (6, 0.02) (7, 47.12) (8,29.51) (9,0.03) (10, 0.01) (11, 35.39) (12, -0.66) (13, -0.23) (14, -1.99) (15, -0.54)};
                \legend {CBCC-IRRG, CCFR2-IRRG};
                \end{axis}
            \end{tikzpicture}
		}
	}
	\hfil
	\\[4pt]
    \subfloat{
		\resizebox{0.978\linewidth}{!}{%
			\begin{tikzpicture}
                \begin{axis} [
                    ybar = 0cm,
                    bar width = 0.5em,
                    xmin = 1,
                    xmax = 15,
                    ymin = -6,
                    ymax = 13.5,
                    xticklabels = {$\sqrt{f_{1}}$, , $\sqrt{f_{3}}$, , $\sqrt{f_{5}}$, , $\sqrt{f_{7}}$, , $\sqrt{f_{9}}$, , $\sqrt{f_{11}}$, , $\sqrt{f_{13}}$, , $\sqrt{f_{15}}$},
                    xtick = data,
                    ytick = {-5, -2.5, 0, 2.5, 5, 7.5, 10, 12.5},
                    enlarge x limits = {abs = .8},
                    width = 9.5cm,
                    height = 4.5cm,
                    legend pos=north west,
                    legend style={font=\fontsize{8}{0}\selectfont},
                    xticklabel style ={font=\fontsize{8}{0}\selectfont},
                    yticklabel style ={font=\fontsize{8}{0}\selectfont},
                    xtick pos=left,
                ]
                 
                \addplot coordinates {(1,-4.56) (2,-0.18) (3,0) (4,2.14) (5,-0.11) (6, 0.01) (7, 11.66) (8,4.01) (9,0.00) (10, 0.00) (11, 7.89) (12, -5.04) (13, -0.16) (14, -0.20) (15, -0.34)};
                \addplot coordinates {(1,-4.56) (2,-0.18) (3,0) (4,2.19) (5,-0.1) (6, 0.00) (7, 11.67) (8,8.12) (9,0) (10, 0) (11, 11.42) (12, -1.30) (13, -0.11) (14, -0.23) (15, -0.34)};
                \legend {CBCC-IRRG, CCFR2-IRRG};
                \end{axis}
            \end{tikzpicture}
		}
	}
	\caption{Orders of magnitude of differences in results between IRRG-embedded CCs and SHADE-ILS (positive numbers indicate CCs' wins, negative numbers otherwise)}
	\label{fig:res:difSize}
\end{figure}

In Fig.~\ref{fig:res:difSize}, we compare the results' quality differences between IRRG-embedded CCs and SHADE-ILS. When CCFR2-IRRG or CBCC-IRRG outperform SHADE-ILS, the differences in the quality of the results are significantly high for some functions (e.g., IRRG-embedded CC are up to $10^{25}$ times better for $f_7$). On the other hand, when SHADE-ILS outperforms IRRG-embedded CCs, the differences in the quality of the results are relatively low (except for $F_1$ and $F_{12}$ for the comparison to CBCC-IRRG). This observation shows that the high-quality decomposition may be the key to reaching the high-quality results for some problems.

\section{Comparison with Fast Variable Interdependence Learning}
\label{sec:fvil}

Fast variable interdependence learning~(FVIL)~\cite{fvil} is a single-parameter decomposition strategy based on monotonicity checking. In our experiments, as suggested in~\cite{fvil}, we used the maximum number of interaction checks between two sets of variables $X_1$ and $X_2$, i.e., a user-defined parameter $N$, set to $10$. In this section, we present the results obtained for functions from the standard CEC'2013, squared CEC'2013, and squared rooted CEC'2013 sets.

Table~\ref{tab:accuracy} presents the decomposition accuracy of FVIL in comparison to the minimal values achieved by IRRG. FVIL never reported false linkage. The values of $\rho_2$ were always equal to $100\%$. However, except for $F_6$ and $F_{10}$, FVIL found only a few of existing interactions. The corresponding $\rho_1$ values were low. Nevertheless, FVIL achieved relatively high values of $\rho_3$. The $\rho_3$ measure compares the resulted interaction matrix with the ideal one. The ideal interaction matrices of the considered functions are sparse. Therefore, missing a lot of existing interactions results in the high value of $\rho_3$.

FVIL and IRRG identified correctly that $F_1$--$F_3$ are fully separable functions. For the partially separable functions, FVIL found several times the perfect decomposition only for functions from $F_6$ and $F_{10}$, whereas IRRG provided the ideal interaction matrix in all runs. Considering only the overlapping functions, we can observe that FVIL tends to create many single-element groups. On the other hand, IRRG often joined all variables into one group. FVIL never found any existing interactions for $F_{15}$. IRRG identified correctly that functions from $F_{15}$ are fully non-separable.

Similarly to IRRG and RDG3, FVIL was embedded into two CC frameworks, CBCC and CCFR2. Optimization methods created after embedding FVIL into CBCC and CCFR2 are denoted as CBCC-FVIL and CCFR2-FVIL. To make a fair comparison between FVIL-embedded CC frameworks and CC frameworks after embedding IRRG and RDG3, especially for the fully separable functions and the partially separable functions with separable variables, we also joined all separable variables into groups of size $\epsilon_s = 100$ in both CBCC-FVIL and CCFR2-FVIL.

Table~\ref{tab:resFVILvsIRRGvsRDG3} presents the effectiveness comparison of FVIL against IRRG and RDG3 in terms of the optimization results obtained after embedding these decomposition strategies into CBCC or CCFR2. The comparison is verified by the unpaired Wilcoxon test (significance level of $5\%$) with the Holm-Bonferroni correction. According to Table~\ref{tab:resFVILvsIRRGvsRDG3}, we can state that CBCC-IRRG outperforms CBCC-FVIL as well as CCFR2-IRRG reports significantly better results than CCFR2-FVIL. The only exceptions where IRRG-embedded CC frameworks were not significantly better than FVIL-embedded CC frameworks are the fully separable functions and $F_{12}$. 
Both decomposition strategies correctly decomposed fully separable problems. Functions in $F_{12}$ are the only functions in the standard and modified CEC'2013 sets that consist of separable subfunctions, and the order of their decision vectors is not rearranged. Since FVIL did not find any existing interactions for $f_{12}$, $(f_{12})^2$, and $\sqrt{f_{12}}$, all variables are joined into groups of size $\epsilon_s = 100$, where groups consist of adjacent variables. Functions in $F_{12}$ are overlapping, and their subfunctions are two-dimensional. Thus, many pairs of directly dependent variables are placed into one group. The detailed optimization results may be found in Tables~\ref{tab:resultsCec}, \ref{tab:resultsSquare}, and \ref{tab:resultsSqrt}.

\begin{table*}
\caption{Decomposition accuracy achieved by FVIL and IRRG}
\label{tab:accuracy}
\scriptsize
\begin{tabularx}{\linewidth}{CCCCCCCCCCCCC}
\toprule
 \multirow{4}{*}{Func} & \multicolumn{9}{>{\hsize=\dimexpr9\hsize+9\tabcolsep+\arrayrulewidth\relax}C}{FVIL} & \multicolumn{3}{>{\hsize=\dimexpr3\hsize+3\tabcolsep+\arrayrulewidth\relax}C}{IRRG} \\
  \cmidrule{2-13}
  & \multicolumn{3}{>{\hsize=\dimexpr3\hsize+3\tabcolsep+\arrayrulewidth\relax}C}{$\rho_1$} & \multicolumn{3}{>{\hsize=\dimexpr3\hsize+3\tabcolsep+\arrayrulewidth\relax}C}{$\rho_2$} & \multicolumn{3}{>{\hsize=\dimexpr3\hsize+3\tabcolsep+\arrayrulewidth\relax}C}{$\rho_3$} & $\rho_1$ & $\rho_2$ & $\rho_3$ \\
  \cmidrule{2-13}
  & Med & Avg & Std & Med & Avg & Std & Med & Avg & Std & Min & Min & Min \\
\midrule

$f_{1}$ & N/A & N/A & N/A & \textbf{100\%} & \textbf{100\%} & \textbf{0\%} & \textbf{100\%} & \textbf{100\%} & \textbf{0\%} & N/A & \textbf{100\%} & \textbf{100\%} \\
$(f_{1})^2$ & N/A & N/A & N/A & \textbf{100\%} & \textbf{100\%} & \textbf{0\%} & \textbf{100\%} & \textbf{100\%} & \textbf{0\%} & N/A & \textbf{100\%} & \textbf{100\%} \\
$\sqrt{f_{1}}$ & N/A & N/A & N/A & \textbf{100\%} & \textbf{100\%} & \textbf{0\%} & \textbf{100\%} & \textbf{100\%} & \textbf{0\%} & N/A & \textbf{100\%} & \textbf{100\%} \\

\addlinespace[0.5em]

$f_{2}$ & N/A & N/A & N/A & \textbf{100\%} & \textbf{100\%} & \textbf{0\%} & \textbf{100\%} & \textbf{100\%} & \textbf{0\%} & N/A & \textbf{100\%} & \textbf{100\%} \\
$(f_{2})^2$ & N/A & N/A & N/A & \textbf{100\%} & \textbf{100\%} & \textbf{0\%} & \textbf{100\%} & \textbf{100\%} & \textbf{0\%} & N/A & \textbf{100\%} & \textbf{100\%} \\
$\sqrt{f_{2}}$ & N/A & N/A & N/A & \textbf{100\%} & \textbf{100\%} & \textbf{0\%} & \textbf{100\%} & \textbf{100\%} & \textbf{0\%} & N/A & \textbf{100\%} & \textbf{100\%} \\

\addlinespace[0.5em]

$f_{3}$ & N/A & N/A & N/A & \textbf{100\%} & \textbf{100\%} & \textbf{0\%} & \textbf{100\%} & \textbf{100\%} & \textbf{0\%} & N/A & \textbf{100\%} & \textbf{100\%} \\
$(f_{3})^2$ & N/A & N/A & N/A & \textbf{100\%} & \textbf{100\%} & \textbf{0\%} & \textbf{100\%} & \textbf{100\%} & \textbf{0\%} & N/A & \textbf{100\%} & \textbf{100\%} \\
$\sqrt{f_{3}}$ & N/A & N/A & N/A & \textbf{100\%} & \textbf{100\%} & \textbf{0\%} & \textbf{100\%} & \textbf{100\%} & \textbf{0\%} & N/A & \textbf{100\%} & \textbf{100\%} \\

\addlinespace[0.5em]

$f_{4}$ & 1.34\% & 1.35\% & 0.18\% & \textbf{100\%} & \textbf{100\%} & \textbf{0\%} & 98.30\% & 98.30\% & 0\% & \textbf{100\%} & \textbf{100\%} & \textbf{100\%} \\
$(f_{4})^2$ & 1.38\% & 1.36\% & 0.17\% & \textbf{100\%} & \textbf{100\%} & \textbf{0\%} & 98.30\% & 98.30\% & 0\% & \textbf{100\%} & \textbf{100\%} & \textbf{100\%} \\
$\sqrt{f_{4}}$ & 1.35\% & 1.33\% & 0.17\% & \textbf{100\%} & \textbf{100\%} & \textbf{0\%} & 98.30\% & 98.30\% & 0\% & \textbf{100\%} & \textbf{100\%} & \textbf{100\%} \\

\addlinespace[0.5em]

$f_{5}$ & 11.27\% & 11.44\% & 2.03\% & \textbf{100\%} & \textbf{100\%} & \textbf{0\%} & 98.47\% & 98.48\% & 0.04\% & \textbf{100\%} & \textbf{100\%} & \textbf{100\%} \\
$(f_{5})^2$ & 10.85\% & 11.44\% & 2.19\% & \textbf{100\%} & \textbf{100\%} & \textbf{0\%} & 98.47\% & 98.48\% & 0.04\% & \textbf{100\%} & \textbf{100\%} & \textbf{100\%} \\
$\sqrt{f_{5}}$ & 11.72\% & 11.78\% & 1.74\% & \textbf{100\%} & \textbf{100\%} & \textbf{0\%} & 98.48\% & 98.48\% & 0.03\% & \textbf{100\%} & \textbf{100\%} & \textbf{100\%} \\

\addlinespace[0.5em]

$f_{6}$ & 98.88\% & 97.63\% & 3.78\% & \textbf{100\%} & \textbf{100\%} & \textbf{0\%} & 99.98\% & 99.96\% & 0.07\% & \textbf{100\%} & \textbf{100\%} & \textbf{100\%} \\
$(f_{6})^2$ & 99.15\% & 97.50\% & 4.66\% & \textbf{100\%} & \textbf{100\%} & \textbf{0\%} & 99.99\% & 99.96\% & 0.08\% & \textbf{100\%} & \textbf{100\%} & \textbf{100\%} \\
$\sqrt{f_{6}}$ & 98.76\% & 97.45\% & 4.84\% & \textbf{100\%} & \textbf{100\%} & \textbf{0\%} & 99.98\% & 99.96\% & 0.08\% & \textbf{100\%} & \textbf{100\%} & \textbf{100\%} \\

\addlinespace[0.5em]

$f_{7}$ & 1.30\% & 1.33\% & 0.20\% & \textbf{100\%} & \textbf{100\%} & \textbf{0\%} & 98.30\% & 98.30\% & 0\% & \textbf{100\%} & \textbf{100\%} & \textbf{100\%} \\
$(f_{7})^2$ & 1.27\% & 1.31\% & 0.21\% & \textbf{100\%} & \textbf{100\%} & \textbf{0\%} & 98.30\% & 98.30\% & 0\% & \textbf{100\%} & \textbf{100\%} & \textbf{100\%} \\
$\sqrt{f_{7}}$ & 1.29\% & 1.30\% & 0.17\% & \textbf{100\%} & \textbf{100\%} & \textbf{0\%} & 98.36\% & 98.30\% & 0\% & \textbf{100\%} & \textbf{100\%} & \textbf{100\%} \\

\addlinespace[0.5em]

$f_{8}$ & 0.83\% & 0.83\% & 0.08\% & \textbf{100\%} & \textbf{100\%} & \textbf{0\%} & 93.27\% & 93.27\% & 0.01\% & \textbf{100\%} & \textbf{100\%} & \textbf{100\%} \\
$(f_{8})^2$ & 0.80\% & 0.81\% & 0.07\% & \textbf{100\%} & \textbf{100\%} & \textbf{0\%} & 93.27\% & 93.27\% & 0\% & \textbf{100\%} & \textbf{100\%} & \textbf{100\%} \\
$f_{8}$ & 0.78\% & 0.79\% & 0.08\% & \textbf{100\%} & \textbf{100\%} & \textbf{0\%} & 93.27\% & 93.27\% & 0.01\% & \textbf{100\%} & \textbf{100\%} & \textbf{100\%} \\

\addlinespace[0.5em]

$f_{9}$ & 9.04\% & 9.36\% & 1.08\% & \textbf{100\%} & \textbf{100\%} & \textbf{0\%} & 93.83\% & 93.85\% & 0.07\% & \textbf{100\%} & \textbf{100\%} & \textbf{100\%} \\
$(f_{9})^2$ & 9.45\% & 9.25\% & 0.81\% & \textbf{100\%} & \textbf{100\%} & \textbf{0\%} & 93.86\% & 93.85\% & 0.05\% & \textbf{100\%} & \textbf{100\%} & \textbf{100\%} \\
$\sqrt{f_{9}}$ & 9.19\% & 9.23\% & 0.72\% & \textbf{100\%} & \textbf{100\%} & \textbf{0\%} & 93.84\% & 93.84\% & 0.05\% & \textbf{100\%} & \textbf{100\%} & \textbf{100\%} \\

\addlinespace[0.5em]

$f_{10}$ & 98.81\% & 97.92\% & 1.89\% & \textbf{100\%} & \textbf{100\%} & \textbf{0\%} & 99.92\% & 99.86\% & 0.13\% & \textbf{100\%} & \textbf{100\%} & \textbf{100\%} \\
$(f_{10})^2$ & 98.81\% & 98.64\% & 1.08\% & \textbf{100\%} & \textbf{100\%} & \textbf{0\%} & 99.92\% & 99.91\% & 0.07\% & \textbf{100\%} & \textbf{100\%} & \textbf{100\%} \\
$\sqrt{f_{10}}$ & 98.48\% & 97.72\% & 2.51\% & \textbf{100\%} & \textbf{100\%} & \textbf{0\%} & 99.90\% & 99.85\% & 0.17\% & \textbf{100\%} & \textbf{100\%} & \textbf{100\%} \\

\addlinespace[0.5em]

$f_{11}$ & 1.15\% & 1.14\% & 0.10\% & \textbf{100\%} & \textbf{100\%} & \textbf{0\%} & 93.30\% & 93.30\% & 0.01\% & \textbf{100\%} & \textbf{100\%} & \textbf{100\%} \\
$(f_{11})^2$ & 1.14\% & 1.14\% & 0.10\% & \textbf{100\%} & \textbf{100\%} & \textbf{0\%} & 93.30\% & 93.30\% & 0.01\% & \textbf{100\%} & \textbf{100\%} & \textbf{100\%} \\
$\sqrt{f_{11}}$ & 1.11\% & 1.11\% & 0.10\% & \textbf{100\%} & \textbf{100\%} & \textbf{0\%} & 93.29\% & 93.29\% & 0.01\% & \textbf{100\%} & \textbf{100\%} & \textbf{100\%} \\

\addlinespace[0.5em]

$f_{12}$ & 0\% & 0\% & 0\% & \textbf{100\%} & \textbf{100\%} & \textbf{0\%} & \textbf{99.80\%} & \textbf{99.80\%} & \textbf{0\%} & \textbf{98.20\%} & 70.67\% & 70.72\% \\
$(f_{12})^2$ & 0\% & 0\% & 0\% & \textbf{100\%} & \textbf{100\%} & \textbf{0\%} & \textbf{99.80\%} & \textbf{99.80\%} & \textbf{0\%} & \textbf{98.60\%} & 34.31\% & 34.44\% \\
$\sqrt{f_{12}}$ & 0\% & 0\% & 0\% & \textbf{100\%} & \textbf{100\%} & \textbf{0\%} & \textbf{99.80\%} & \textbf{99.80\%} & \textbf{0\%} & \textbf{98.30\%} & 59.55\% & 59.63\% \\

\addlinespace[0.5em]

$f_{13}$ & 1.16\% & 1.17\% & 0.08\% & \textbf{100\%} & \textbf{100\%} & \textbf{0\%} & \textbf{91.86\%} & \textbf{91.86\%} & \textbf{0.01\%} & \textbf{98.99\%} & 0\% & 8.23\% \\
$(f_{13})^2$ & 1.17\% & 1.17\% & 0.10\% & \textbf{100\%} & \textbf{100\%} & \textbf{0\%} & \textbf{91.86\%} & \textbf{91.86\%} & \textbf{0.01\%} & \textbf{99.33\%} & 0\% & 8.23\% \\
$\sqrt{f_{13}}$ & 1.21\% & 1.22\% & 0.11\% & \textbf{100\%} & \textbf{100\%} & \textbf{0\%} & \textbf{91.86\%} & \textbf{91.87\%} & \textbf{0.01\%} & \textbf{99.41\%} & 0\% & 8.23\% \\

\addlinespace[0.5em]

$f_{14}$ & 1.17\% & 1.17\% & 0.09\% & \textbf{100\%} & \textbf{100\%} & \textbf{0\%} & \textbf{91.86\%} & \textbf{91.86\%} & \textbf{0.01\%} & \textbf{99.05\%} & 0\% & 8.23\% \\
$(f_{14})^2$ & 1.15\% & 1.15\% & 0.08\% & \textbf{100\%} & \textbf{100\%} & \textbf{0\%} & \textbf{91.86\%} & \textbf{91.86\%} & \textbf{0.01\%} & \textbf{98.92\%} & 0\% & 8.23\% \\
$\sqrt{f_{14}}$ & 1.13\% & 1.15\% & 0.11\% & \textbf{100\%} & \textbf{100\%} & \textbf{0\%} & \textbf{91.86\%} & \textbf{91.86\%} & \textbf{0.01\%} & \textbf{99.24\%} & 0\% & 8.23\% \\

\addlinespace[0.5em]

$f_{15}$ & 0\% & 0\% & 0\% & N/A & N/A & N/A & 0\% & 0\% & 0\% & \textbf{100\%} & N/A & \textbf{100\%} \\
$(f_{15})^2$ & 0\% & 0\% & 0\% & N/A & N/A & N/A & 0\% & 0\% & 0\% & \textbf{100\%} & N/A & \textbf{100\%} \\
$\sqrt{f_{15}}$ & 0\% & 0\% & 0\% & N/A & N/A & N/A & 0\% & 0\% & 0\% & \textbf{100\%} & N/A & \textbf{100\%} \\

\bottomrule
\end{tabularx}
\end{table*}

\begin{table*}
\setlength{\tabcolsep}{0.5em}

\caption{Effectiveness comparison of FVIL against IRRG and RDG3}
\label{tab:resFVILvsIRRGvsRDG3}
\scriptsize
\begin{tabularx}{\linewidth}{CCCCCCCCCCCCC}
\toprule
        \multirow{3}{*}{$i$} & \multicolumn{3}{>{\hsize=\dimexpr3\hsize+3\tabcolsep+\arrayrulewidth\relax}C}{CBCC-IRRG vs. CBCC-FVIL}  & \multicolumn{3}{>{\hsize=\dimexpr3\hsize+3\tabcolsep+\arrayrulewidth\relax}C}{CCFR2-IRRG vs. CCFR2-FVIL} & \multicolumn{3}{>{\hsize=\dimexpr3\hsize+3\tabcolsep+\arrayrulewidth\relax}C}{CBCC-RDG3 vs. CBCC-FVIL} & \multicolumn{3}{>{\hsize=\dimexpr3\hsize+3\tabcolsep+\arrayrulewidth\relax}C}{CCFR2-RDG3 vs. CCFR2-FVIL} \\
        \cmidrule{2-13}
       & $f_i$ & $(f_i)^2$ & $\sqrt{f_i}$    & $f_i$ & $(f_i)^2$ & $\sqrt{f_i}$ & $f_i$ & $(f_i)^2$ & $\sqrt{f_i}$ & $f_i$ & $(f_i)^2$ & $\sqrt{f_i}$    \\
       & w/t/l & w/t/l & w/t/l & w/t/l & w/t/l & w/t/l & w/t/l & w/t/l & w/t/l & w/t/l & w/t/l & w/t/l \\
       \midrule
       
1--3  & 0/\textbf{3}/0    & 0/\textbf{3}/0   & 0/\textbf{3}/0   & 0/\textbf{3}/0    & 0/\textbf{3}/0   & 0/\textbf{3}/0 & 0/\textbf{2}/1 & 0/0/\textbf{3} & 0/0/\textbf{3} & 0/\textbf{2}/1 & 0/0/\textbf{3} & 0/0/\textbf{3} \\
4--7  & \textbf{3}/1/0    & \textbf{4}/0/0   & \textbf{3}/1/0  & \textbf{4}/0/0    & \textbf{3}/1/0   & \textbf{3}/1/0 & \textbf{4}/0/0 & \textbf{3}/0/1 & \textbf{3}/0/1 & \textbf{3}/1/0 & \textbf{3}/0/1 & \textbf{3}/0/1 \\
8--11  & \textbf{3}/1/0   & \textbf{3}/1/0   & \textbf{3}/1/0   & \textbf{3}/1/0    & \textbf{3}/1/0   & \textbf{3}/1/0 & \textbf{3}/1/0 & \textbf{3}/1/0 & \textbf{3}/0/1 & \textbf{3}/1/0 & \textbf{3}/1/0 & \textbf{3}/1/0 \\
12--14 & \textbf{2}/0/1    & \textbf{2}/0/1   & \textbf{2}/0/1   & \textbf{2}/1/0    & \textbf{2}/0/1   & \textbf{2}/0/1 & \textbf{2}/1/0 & \textbf{2}/0/1 & \textbf{2}/0/1 & \textbf{2}/1/0 & \textbf{2}/0/1 & \textbf{2}/0/1 \\
15   & \textbf{1}/0/0    & \textbf{1}/0/0   & \textbf{1}/0/0   & \textbf{1}/0/0    & \textbf{1}/0/0   & \textbf{1}/0/0 & \textbf{1}/0/0 & \textbf{1}/0/0 & \textbf{1}/0/0 & \textbf{1}/0/0 & \textbf{1}/0/0 & \textbf{1}/0/0 \\
\midrule
Total  & \textbf{9}/5/1   & \textbf{10}/4/1 & \textbf{9}/5/1 & \textbf{10}/5/0  & \textbf{9}/5/1 & \textbf{9}/5/1 & \textbf{10}/4/1 & \textbf{9}/1/5 & \textbf{9}/0/6 & \textbf{9}/5/1 & \textbf{9}/1/5 & \textbf{9}/1/5 \\
 
 \bottomrule
\end{tabularx}
\end{table*}

\begin{table}
\setlength{\tabcolsep}{0.5em}
 \caption{Results for the standard CEC'2013 set}
\label{tab:resultsCec}
\scriptsize
\begin{tabularx}{\columnwidth}{CCCCCC}
\toprule
 Func & Stats & CBCC-IRRG & CCFR2-IRRG & CBCC-FVIL & CCFR2-FVIL \\

\midrule

\multirow{3}{*}{$f_1$} & Med & \textbf{7.87E-19} & \textbf{6.78E-19} & \textbf{7.12E-19} & \textbf{1.11E-18}  \\
& Avg & \textbf{9.16E-19} & \textbf{7.41E-19} & \textbf{8.60E-19} & \textbf{1.05E-18} \\
& Std & \textbf{3.35E-19} & \textbf{1.89E-19} & \textbf{3.00E-19} & \textbf{3.12E-19} \\

\addlinespace[0.5em]

\multirow{3}{*}{$f_2$} & Med & \textbf{2.35E+03} & \textbf{2.35E+03} & \textbf{2.35E+03} & \textbf{2.36E+03} \\
& Avg & \textbf{2.34E+03} & \textbf{2.36E+03} & \textbf{2.36E+03} & \textbf{2.35E+03}  \\
& Std & \textbf{9.64E+01} & \textbf{1.24E+02} & \textbf{9.06E+01} & \textbf{1.06E+02}  \\

\addlinespace[0.5em]

\multirow{3}{*}{$f_3$} & Med & \textbf{2.02E+01} & \textbf{2.02E+01} & \textbf{2.02E+01} & \textbf{2.02E+01} \\
& Avg & \textbf{2.02E+01} & \textbf{2.02E+01} & \textbf{2.01E+01} & \textbf{2.02E+01}  \\
& Std & \textbf{1.07E-01} & \textbf{1.07E-01} & \textbf{1.00E-01} & \textbf{1.05E-01}  \\

\addlinespace[0.5em]

\multirow{3}{*}{$f_4$} & Med & \textbf{1.58E+04} & 2.08E+03 & 2.46E+11 & 3.13E+11 \\
& Avg & \textbf{1.97E+04} & 1.62E+04 & 2.67E+11 & 3.51E+11 \\
& Std & \textbf{2.00E+04} & 3.03E+04 & 8.28E+10 & 1.57E+11 \\

\addlinespace[0.5em]

\multirow{3}{*}{$f_5$} & Med & \textbf{2.42E+06} & \textbf{2.22E+06} & 8.36E+06 & 9.86E+06 \\
& Avg & \textbf{2.32E+06} & \textbf{2.17E+06} & 8.67E+06 & 9.79E+06 \\
& Std & \textbf{4.97E+05} & \textbf{3.23E+05} & 2.29E+06 & 1.96E+06 \\

\addlinespace[0.5em]

\multirow{3}{*}{$f_6$} & Med & \textbf{9.96E+05} & \textbf{9.96E+05} & \textbf{9.96E+05} & 9.96E+05 \\
& Avg & \textbf{1.00E+06} & \textbf{1.01E+06} & \textbf{1.01E+06} & 1.00E+06 \\
& Std & \textbf{2.27E+04} & \textbf{2.27E+04} & \textbf{2.72E+04} & 2.18E+04 \\

\addlinespace[0.5em]

\multirow{3}{*}{$f_7$} & Med & \textbf{9.43E-22} & \textbf{9.25E-22} & 1.69E+09 & 1.77E+09 \\
& Avg & \textbf{9.33E-22} & \textbf{9.27E-22} & 2.43E+09 & 1.96E+09 \\
& Std & \textbf{9.62E-23} & \textbf{7.24E-23} & 1.84E+09 & 1.13E+09 \\

\addlinespace[0.5em]

\multirow{3}{*}{$f_8$} & Med & 9.13E+03 & \textbf{3.50E-05} & 1.43E+16 & 1.61E+16 \\
& Avg & 9.12E+03 & \textbf{4.47E-05} & 1.57E+16 & 1.76E+16 \\
& Std & 1.37E+03 & \textbf{2.99E-05} & 8.50E+15 & 8.39E+15 \\

\addlinespace[0.5em]

\multirow{3}{*}{$f_9$} & Med & \textbf{1.59E+08} & \textbf{1.50E+08} & 8.64E+08 & 8.66E+08 \\
& Avg & \textbf{1.64E+08} & \textbf{1.49E+08} & 8.53E+08 & 9.05E+08 \\
& Std & \textbf{2.76E+07} & \textbf{3.19E+07} & 1.94E+08 & 2.74E+08 \\

\addlinespace[0.5em]

\multirow{3}{*}{$f_{10}$} & Med & \textbf{9.05E+07} & \textbf{9.05E+07} & \textbf{9.05E+07} & \textbf{9.06E+07} \\
& Avg & \textbf{9.10E+07} & \textbf{9.17E+07} & \textbf{9.13E+07} & \textbf{9.19E+07} \\
& Std & \textbf{1.27E+06} & \textbf{1.88E+06} & \textbf{1.69E+06} & \textbf{1.96E+06} \\

\addlinespace[0.5em]

\multirow{3}{*}{$f_{11}$} & Med & 5.49E-12 & \textbf{6.70E-18} & 4.77E+11 & 4.41E+11 \\
& Avg & 1.01E-10 & \textbf{1.73E-17} & 4.94E+11 & 4.72E+11 \\
& Std & 1.56E-10 & \textbf{3.36E-17} & 2.81E+11 & 2.12E+11 \\

\addlinespace[0.5em]

\multirow{3}{*}{$f_{12}$} & Med & 8.12E+02 & 9.04E+02 & \textbf{6.77E+02} & 8.14E+02 \\
& Avg & 2.32E+03 & 9.06E+02 & \textbf{6.83E+02} & 8.32E+02 \\
& Std & 7.17E+03 & 7.18E+01 & \textbf{7.86E+01} & 4.63E+01 \\

\addlinespace[0.5em]

\multirow{3}{*}{$f_{13}$} & Med & \textbf{1.34E+06} & \textbf{1.50E+06} & 9.87E+09 & 1.03E+10 \\
& Avg & \textbf{1.14E+06} & \textbf{1.30E+06} & 1.03E+10 & 1.06E+10 \\
& Std & \textbf{5.61E+05} & \textbf{5.80E+05} & 3.69E+09 & 4.59E+09 \\

\addlinespace[0.5em]

\multirow{3}{*}{$f_{14}$} & Med & \textbf{1.60E+07} & \textbf{1.53E+07} & 2.49E+11 & 2.19E+11 \\
& Avg & \textbf{1.89E+07} & \textbf{2.01E+07} & 2.50E+11 & 2.41E+11 \\
& Std & \textbf{8.75E+06} & \textbf{1.14E+07} & 8.36E+10 & 9.40E+10 \\

\addlinespace[0.5em]

\multirow{3}{*}{$f_{15}$} & Med & \textbf{2.22E+06} & \textbf{2.27E+06} & 3.41E+08 & 5.95E+08 \\
& Avg & \textbf{2.21E+06} & \textbf{2.29E+06} & 3.95E+08 & 6.16E+08 \\
& Std & \textbf{2.35E+05} & \textbf{1.91E+05} & 1.70E+08 & 1.29E+08 \\
\bottomrule
\end{tabularx}
\end{table}

\begin{table}
\setlength{\tabcolsep}{0.5em}
 \caption{Results for the squared CEC'2013 set}
\label{tab:resultsSquare}
\scriptsize
\begin{tabularx}{\columnwidth}{CCCCCC}
\toprule
 Func & Stats & CBCC-IRRG & CCFR2-IRRG & CBCC-FVIL & CCFR2-FVIL \\
\midrule

\multirow{3}{*}{$(f_{1})^2$} & Med & \textbf{5.75E-37} & \textbf{4.90E-37} & \textbf{7.76E-37} & \textbf{5.19E-37} \\
& Avg & \textbf{9.18E-37} & \textbf{8.10E-37} & \textbf{9.74E-37} & \textbf{7.42E-37} \\
& Std & \textbf{6.67E-37} & \textbf{6.99E-37} & \textbf{5.92E-37} & \textbf{4.25E-37} \\

\addlinespace[0.5em]

\multirow{3}{*}{$(f_{2})^2$} & Med & \textbf{5.25E+06} & \textbf{5.53E+06} & \textbf{5.21E+06} & \textbf{5.39E+06} \\
& Avg & \textbf{5.34E+06} & \textbf{5.53E+06} & \textbf{5.22E+06} & \textbf{5.35E+06} \\
& Std & \textbf{5.25E+05} & \textbf{5.09E+05} & \textbf{5.05E+05} & \textbf{4.44E+05} \\

\addlinespace[0.5em]

\multirow{3}{*}{$(f_{3})^2$} & Med & \textbf{4.07E+02} & \textbf{4.07E+02} & \textbf{4.07E+02} & \textbf{4.08E+02} \\
& Avg & \textbf{4.06E+02} & \textbf{4.08E+02} & \textbf{4.06E+02} & \textbf{4.10E+02} \\
& Std & \textbf{5.12E+00} & \textbf{3.74E+00} & \textbf{4.11E+00} & \textbf{3.71E+00} \\

\addlinespace[0.5em]

\multirow{3}{*}{$(f_{4})^2$} & Med & \textbf{1.53E+07} & \textbf{3.72E+07} & 6.97E+22 & 8.40E+22 \\
& Avg & \textbf{9.45E+08} & \textbf{3.88E+08} & 1.04E+23 & 1.85E+23 \\
& Std & \textbf{3.52E+09} & \textbf{5.12E+08} & 1.20E+23 & 2.36E+23 \\

\addlinespace[0.5em]

\multirow{3}{*}{$(f_{5})^2$} & Med & \textbf{5.18E+12} & \textbf{4.24E+12} & 8.99E+13 & 8.72E+13 \\
& Avg & \textbf{5.44E+12} & \textbf{4.46E+12} & 8.98E+13 & 8.90E+13 \\
& Std & \textbf{1.84E+12} & \textbf{1.46E+12} & 3.44E+13 & 3.93E+13 \\

\addlinespace[0.5em]

\multirow{3}{*}{$(f_{6})^2$} & Med & \textbf{9.92E+11} & \textbf{9.92E+11} & \textbf{9.92E+11} & \textbf{9.92E+11} \\
& Avg & \textbf{1.00E+12} & \textbf{1.01E+12} & \textbf{1.00E+12} & \textbf{1.00E+12} \\
& Std & \textbf{3.83E+10} & \textbf{4.59E+10} & \textbf{3.62E+10} & \textbf{3.85E+10} \\

\addlinespace[0.5em]

\multirow{3}{*}{$(f_{7})^2$} & Med & \textbf{8.08E-43} & \textbf{8.59E-43} & 2.25E+18 & 2.47E+18 \\
& Avg & \textbf{8.15E-43} & \textbf{8.89E-43} & 3.12E+18 & 4.63E+18 \\
& Std & \textbf{1.54E-43} & \textbf{1.46E-43} & 2.93E+18 & 4.47E+18 \\

\addlinespace[0.5em]

\multirow{3}{*}{$(f_{8})^2$} & Med & 7.66E+07 & \textbf{4.85E-07} & 1.50E+32 & 1.66E+32 \\
& Avg & 7.88E+07 & \textbf{1.20E-06} & 1.89E+32 & 2.13E+32 \\
& Std & 2.63E+07 & \textbf{2.16E-06} & 1.45E+32 & 1.77E+32 \\

\addlinespace[0.5em]

\multirow{3}{*}{$(f_{9})^2$} & Med & \textbf{2.81E+16} & \textbf{2.46E+16} & 7.91E+17 & 7.04E+17 \\
& Avg & \textbf{2.65E+16} & \textbf{2.56E+16} & 8.81E+17 & 6.32E+17 \\
& Std & \textbf{9.74E+15} & \textbf{8.92E+15} & 4.54E+17 & 1.89E+17 \\

\addlinespace[0.5em]

\multirow{3}{*}{$(f_{10})^2$} & Med & \textbf{8.20E+15} & \textbf{8.20E+15} & \textbf{8.21E+15} & \textbf{8.20E+15} \\
& Avg & \textbf{8.32E+15} & \textbf{8.36E+15} & \textbf{8.39E+15} & \textbf{8.42E+15} \\
& Std & \textbf{2.75E+14} & \textbf{3.19E+14} & \textbf{3.26E+14} & \textbf{3.62E+14} \\

\addlinespace[0.5em]

\multirow{3}{*}{$(f_{11})^2$} & Med & \textbf{2.50E-27} & \textbf{2.51E-26} & 1.77E+23 & 1.15E+23 \\
& Avg & \textbf{2.35E-20} & \textbf{1.29E-24} & 2.30E+23 & 2.37E+23 \\
& Std & \textbf{7.68E-20} & \textbf{4.28E-24} & 2.05E+23 & 2.95E+23 \\

\addlinespace[0.5em]

\multirow{3}{*}{$(f_{12})^2$} & Med & 8.29E+05 & 7.97E+05 & \textbf{4.86E+05} & 6.63E+05 \\
& Avg & 7.95E+22 & 8.06E+05 & \textbf{5.13E+05} & 6.80E+05 \\
& Std & 2.27E+23 & 1.01E+05 & \textbf{9.55E+04} & 4.96E+04 \\

\addlinespace[0.5em]

\multirow{3}{*}{$(f_{13})^2$} & Med & \textbf{7.70E+11} & \textbf{1.84E+12} & 1.11E+20 & 6.81E+19 \\
& Avg & \textbf{1.31E+12} & \textbf{1.57E+12} & 1.67E+20 & 7.64E+19 \\
& Std & \textbf{1.09E+12} & \textbf{1.01E+12} & 1.94E+20 & 5.69E+19 \\

\addlinespace[0.5em]

\multirow{3}{*}{$(f_{14})^2$} & Med & \textbf{3.64E+14} & \textbf{7.11E+14} & 3.38E+22 & 5.10E+22 \\
& Avg & \textbf{4.83E+14} & \textbf{5.85E+15} & 5.01E+22 & 6.71E+22 \\
& Std & \textbf{3.08E+14} & \textbf{2.65E+16} & 4.37E+22 & 5.58E+22 \\

\addlinespace[0.5em]

\multirow{3}{*}{$(f_{15})^2$} & Med & \textbf{5.28E+12} & \textbf{5.12E+12} & 1.17E+17 & 1.64E+17 \\
& Avg & \textbf{5.36E+12} & \textbf{5.23E+12} & 1.72E+17 & 1.61E+17 \\
& Std & \textbf{1.10E+12} & \textbf{1.03E+12} & 1.65E+17 & 4.28E+16 \\
\bottomrule
\end{tabularx}
\end{table}

\begin{table}
\setlength{\tabcolsep}{0.5em}
 \caption{Results for the square rooted CEC'2013 set}
\label{tab:resultsSqrt}
\scriptsize
\begin{tabularx}{\columnwidth}{CCCCCC}
\toprule
 Func & Stats & CBCC-IRRG & CCFR2-IRRG & CBCC-FVIL & CCFR2-FVIL \\
\midrule

\multirow{3}{*}{$\sqrt{f_{1}}$} & Med & \textbf{8.86E-10} & \textbf{8.78E-10} & \textbf{8.33E-10} & \textbf{8.83E-10} \\
& Avg & \textbf{9.20E-10} & \textbf{9.16E-10} & \textbf{8.72E-10} & \textbf{9.18E-10} \\
& Std & \textbf{1.75E-10} & \textbf{1.35E-10} & \textbf{1.33E-10} & \textbf{1.72E-10} \\

\addlinespace[0.5em]

\multirow{3}{*}{$\sqrt{f_{2}}$} & Med & \textbf{4.83E+01} & \textbf{4.83E+01} & \textbf{4.81E+01} & \textbf{4.89E+01} \\
& Avg & \textbf{4.84E+01} & \textbf{4.85E+01} & \textbf{4.83E+01} & \textbf{4.86E+01} \\
& Std & \textbf{9.51E-01} & \textbf{1.00E+00} & \textbf{1.09E+00} & \textbf{1.06E+00} \\

\addlinespace[0.5em]

\multirow{3}{*}{$\sqrt{f_{3}}$} & Med & \textbf{4.49E+00} & \textbf{4.49E+00} & \textbf{4.49E+00} & \textbf{4.49E+00} \\
& Avg & \textbf{4.49E+00} & \textbf{4.49E+00} & \textbf{4.49E+00} & \textbf{4.50E+00}` \\
& Std & \textbf{1.26E-02} & \textbf{1.20E-02} & \textbf{1.03E-02} & \textbf{1.05E-02}` \\

\addlinespace[0.5em]

\multirow{3}{*}{$\sqrt{f_{4}}$} & Med & \textbf{1.07E+02} & \textbf{9.68E+01} & 5.53E+05 & 5.79E+05 \\
& Avg & \textbf{1.18E+02} & \textbf{1.05E+02} & 5.65E+05 & 6.15E+05 \\
& Std & \textbf{6.10E+01} & \textbf{9.66E+01} & 8.89E+04 & 1.20E+05 \\

\addlinespace[0.5em]

\multirow{3}{*}{$\sqrt{f_{5}}$} & Med & \textbf{1.51E+03} & \textbf{1.47E+03} & 2.80E+03 & 3.01E+03 \\
& Avg & \textbf{1.49E+03} & \textbf{1.45E+03} & 2.84E+03 & 3.04E+03 \\
& Std & \textbf{1.25E+02} & \textbf{1.47E+02} & 3.62E+02 & 3.66E+02 \\

\addlinespace[0.5em]

\multirow{3}{*}{$\sqrt{f_{6}}$} & Med & \textbf{9.98E+02} & \textbf{9.98E+02} & \textbf{9.98E+02} & 9.98E+02 \\
& Avg & \textbf{1.00E+03} & \textbf{1.00E+03} & \textbf{1.00E+03} & 1.00E+03 \\
& Std & \textbf{9.22E+00} & \textbf{1.36E+01} & \textbf{1.26E+01} & 1.11E+01 \\

\addlinespace[0.5em]

\multirow{3}{*}{$\sqrt{f_{7}}$} & Med & \textbf{3.00E-11} & \textbf{2.99E-11} & 4.14E+04 & 4.33E+04 \\
& Avg & \textbf{3.01E-11} & \textbf{2.99E-11} & 4.36E+04 & 4.94E+04 \\
& Std & \textbf{1.11E-12} & \textbf{1.48E-12} & 1.15E+04 & 1.90E+04 \\

\addlinespace[0.5em]

\multirow{3}{*}{$\sqrt{f_{8}}$} & Med & 9.95E+01 & \textbf{5.46E-03} & 1.10E+08 & 1.24E+08 \\
& Avg & 9.97E+01 & \textbf{7.77E-03} & 1.14E+08 & 1.28E+08 \\
& Std & 9.71E+00 & \textbf{7.70E-03} & 2.01E+07 & 2.42E+07 \\

\addlinespace[0.5em]

\multirow{3}{*}{$\sqrt{f_{9}}$} & Med & \textbf{1.28E+04} & \textbf{1.25E+04} & 3.02E+04 & 2.97E+04 \\
& Avg & \textbf{1.25E+04} & \textbf{1.24E+04} & 3.04E+04 & 2.94E+04 \\
& Std & \textbf{1.03E+03} & \textbf{1.22E+03} & 4.25E+03 & 4.32E+03 \\

\addlinespace[0.5em]

\multirow{3}{*}{$\sqrt{f_{10}}$} & Med & \textbf{9.52E+03} & \textbf{9.52E+03} & \textbf{9.52E+03} & 9.52E+03 \\
& Avg & \textbf{9.55E+03} & \textbf{9.59E+03} & 9.55E+03 & 9.59E+03 \\
& Std & \textbf{7.85E+01} & \textbf{1.03E+02} & 7.68E+01 & 1.02E+02 \\

\addlinespace[0.5em]

\multirow{3}{*}{$\sqrt{f_{11}}$} & Med & 4.81E-07 & \textbf{2.74E-09} & 6.34E+05 & 6.06E+05 \\
& Avg & 8.93E-06 & \textbf{2.68E-09} & 6.31E+05 & 5.96E+05 \\
& Std & 1.27E-05 & \textbf{7.49E-10} & 1.58E+05 & 1.79E+05 \\

\addlinespace[0.5em]

\multirow{3}{*}{$\sqrt{f_{12}}$} & Med & 3.05E+01 & 3.00E+01 & \textbf{2.62E+01} & 2.85E+01 \\
& Avg & 1.66E+05 & 3.02E+01 & \textbf{2.63E+01} & 2.85E+01 \\
& Std & 3.39E+05 & 1.07E+00 & \textbf{1.27E+00} & 1.04E+00 \\

\addlinespace[0.5em]

\multirow{3}{*}{$\sqrt{f_{13}}$} & Med & \textbf{1.23E+03} & \textbf{8.38E+02} & 9.41E+04 & 9.50E+04\\
& Avg & \textbf{1.08E+03} & \textbf{9.56E+02} & 9.70E+04 & 9.60E+04 \\
& Std & \textbf{2.63E+02} & \textbf{3.07E+02} & 1.44E+04 & 1.13E+04 \\

\addlinespace[0.5em]

\multirow{3}{*}{$\sqrt{f_{14}}$} & Med & \textbf{4.20E+03} & \textbf{5.13E+03} & 4.87E+05 & 4.95E+05 \\
& Avg & \textbf{4.34E+03} & \textbf{4.64E+03} & 4.80E+05 & 5.01E+05 \\
& Std & \textbf{9.63E+02} & \textbf{9.36E+02} & 8.52E+04 & 9.80E+04 \\

\addlinespace[0.5em]

\multirow{3}{*}{$\sqrt{f_{15}}$} & Med & \textbf{1.51E+03} & \textbf{1.49E+03} & 2.02E+04 & 2.51E+04 \\
& Avg & \textbf{1.52E+03} & \textbf{1.49E+03} & 2.01E+04 & 2.48E+04 \\
& Std & \textbf{7.99E+01} & \textbf{6.46E+01} & 4.16E+03 & 4.25E+03 \\
\bottomrule
\end{tabularx}
\end{table}

\section{Real-World Optimization Problem with Non-Additively Separable Subproblems}

As a real-world optimization problem with non-additively separable subproblems, we consider a multi-path routing problem in computer and communication networks \cite{Pioro_Book_2004, Cideon_1999, Banner_2007}. The multi-path routing problem is an important optimization problem in many types of computer and communication networks, e.g., Internet \cite{He_2008}, mobile ad hoc networks (MANETs) \cite{Mueller_2003}, wireless sensor networks (WSNs) \cite{Radi_2012}, and elastic optical networks \cite{Zhu_2013}.

\subsection{Problem Description}

In contrast to the traditional routing approach that transmits all traffic of a given demand along a single path, the multi-path routing approach splits the traffic among several paths. Multi-path routing provides path diversity, a common requirement in computer and communication networks that implies splitting demand volumes into more than one routing path. There are several motivations behind multi-path routing. First, multi-path routing increases the survivability of the network since, in the case of a single link failure (the most common network failure), not all routing paths are affected by the failure. Thus, a part of the demand can still be transmitted in the network. Second, multi-path routing improves the efficiency of the network in terms of various QoS (Quality of Service) parameters, including congestion and throughput \cite{Pioro_Book_2004, Cideon_1999, Banner_2007, He_2008}.

The considered network is modelled as a graph $G = (V, E)$, where $V$ is a set of nodes (vertices), and $E$ is a set of edges (directed links). We use a \textit{link-path formulation of multicommodity flows} to model network flows. It should be noted that the theory of multicommodity flows originally proposed in the context of transport networks is commonly used to model flows in computer and communication networks \cite{Assad_Networks_1978,Pioro_Book_2004}. A commodity (which we here call a \textit{demand}) is defined by a source node, destination node and volume (bit-rate). We assume that set $D$ contains all demands that must be established in the network. For each demand $d \in D$, there is a set of candidate paths $P(d)$ connecting the origin and destination node of demand $d$. The set of candidate paths $P(d)$ can contain either all possible routing paths that can be calculated in the graph or only a selected subset of paths. In our case, we use the latter approach, i.e., we consider a limited number of candidate routing paths. Routing paths are defined with constant~${\delta}_{e,d,p}$, i.e., ${\delta}_{e,d,p}$ is 1 if path~$p \in P(d)$ of demand~$d \in D$ contains link~$e \in E$ and 0 otherwise. In the context of multi-path routing, we assume that the demand volume $h_d$ is to be transmitted in the network using any number of routing paths included in set $P(d)$. To this end, we use a decision variable $x_{d,p}$, which indicates the fraction (percentage) of demand $d$ allocated to path $p \in P(d)$. As the objective function, we apply the average delay function, which is a commonly used performance metric for computer networks \cite{Fratta_Networks_1973, Pioro_Book_2004}.

The considered multi-path routing optimization problem is formulated as follows. 

\noindent\textbf{sets} 

\begin{tabular}{p{0,5cm} p{6,5cm}}
$E$   & links\\
\end{tabular}

\begin{tabular}{p{0,5cm} p{6,5cm}}
$D$   & demands\\
\end{tabular}

\begin{tabular}{p{0,5cm} p{6,5cm}}
$P(d)$  & candidate paths for flows realizing demand~$d$\\
\end{tabular}

\noindent\textbf{constants} 

\begin{tabular}{p{0,5cm} p{6,5cm}}
${\delta}_{e,d,p}$   & =1, if link~$e$ belongs to path~$p$ realizing demand~$d$; 0, otherwise \\
\end{tabular}

\begin{tabular}{p{0,5cm} p{6,5cm}}
$h_d$   & volume (bit-rate) of demand~$d$\\
\end{tabular}

\begin{tabular}{p{0,5cm} p{6,5cm}}
$c_e$   & capacity of link~$e$\\
\end{tabular}

\noindent\textbf{variables} 

\begin{tabular}{p{0,5cm} p{6,5cm}}
$x_{d,p}$   &fraction of demand $d$ flow allocated to path $p$ (continuous non-negative)\\
\end{tabular}

\begin{tabular}{p{0,5cm} p{6,5cm}}
$f_{e}$   &flow on link~$e$ (continuous non-negative) \\
\end{tabular}

\noindent\textbf{objective} 
\begin{subequations} \label{Model:CON/A/FA/Cost/Link-path}
\allowdisplaybreaks
\begin{flalign}
&\text{minimize} \quad F= {\sum\limits_{e\in E}  \frac{f_e}{c_e - f_e }} &
\label{Model:a}
\end{flalign}
\noindent\textbf{constraints} 
\begin{flalign}
& f_e = {\sum\limits_{d\in D} \sum\limits_{p\in P(d)}} {\delta}_{e,d,p} x_{d,p} h_d, \quad e\in E 
\label{Model:b}&\\
& {\sum\limits_{p\in P(d)} } x_{d,p}=1, \quad d\in D 
\label{Model:c}&\\
& f_e \leq c_e, \quad e \in E
\label{Model:d}&
\end{flalign}
\end{subequations}

Objective (\ref{Model:a}) is to minimize the average delay in the network. Note that the objective function is convex. Equality (\ref{Model:b}) defines the value of variable~$f_e$, i.e., the overall flow on link $e$ calculated as a sum of all demands that use that link. Constraint~(\ref{Model:c}) ensures that the whole demand is realized in the network accounting for all candidate routing paths included in set $P(d)$.  Finally, (\ref{Model:d}) is a link capacity constraint formulated to guarantee that the flow of each link given as a sum of all demands that use this link cannot exceed the link capacity.

\subsection{Non-Additive Separability}
\label{sec:cn:sep}

\begin{figure}
    \centering
        \begin{tikzpicture}[node distance={24mm}, main/.style = {draw, circle}, every node/.style={draw}, every newellipse node/.style={inner sep=0pt}] 
            \node[main] (1) {$n_1^*$};
            \node[main] (2) [above right of=1] {$n_2^*$};
            \node[main] (3) [below right of=1] {$n_3^*$};
            \node[main] (4) [above right of=3] {$n_4^*$};
            
            \draw (1) -- (2) node [midway, above left, draw=none] {$e_1^*$};
            \draw (1) -- (3) node [midway, below left, draw=none] {$e_2^*$};
            \draw (2) -- (3) node [midway, right, draw=none] {$e_3^*$};
            \draw (2) -- (4) node [midway, above right, draw=none] {$e_4^*$};
            \draw (3) -- (4) node [midway, below right, draw=none] {$e_5^*$};
        \end{tikzpicture}
    \caption{Exemplary computer network consisting of $4$ nodes and $5$ links}
    \label{fig:computer_network}
\end{figure}
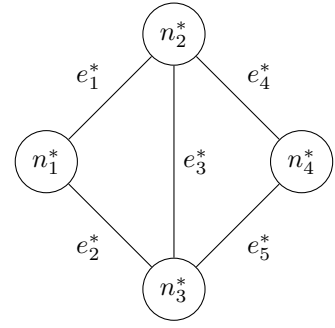

Some instances of the multi-path routing problem are non-additively separable. Let us analyze the following example. We consider a computer network presented in Fig.~\ref{fig:computer_network} and two demands $d_1^*$ and $d_2^*$. The origin node of $d_1^*$ is $n_1^*$, and its destination node is $n_3^*$. For $d_2^*$, we consider paths that connect $n_2^*$ and $n_4^*$. We select two paths for each demand. The paths are as follows.
\begin{itemize}
    \item The first path for $d_1^*$: $e_2^*$.
    \item The second path for $d_1^*$: $e_1^* \rightarrow e_3^*$.
    \item The first path for $d_2^*$: $e_4^*$.
    \item The second path for $d_2^*$: $e_3^* \rightarrow e_5^*$.
\end{itemize}
Therefore, flow on links can be calculated as follows
\begin{subequations}
    \begin{flalign}
       f_{e_1^*} = x_{d_1^*,2} h_{d_1^*}
    \end{flalign}
    \begin{flalign}
       f_{e_2^*} = x_{d_1^*,1} h_{d_1^*}
    \end{flalign}
    \begin{flalign}
       f_{e_3^*} = x_{d_1^*,2} h_{d_1^*} + x_{d_2^*,2} h_{d_2^*}
    \end{flalign}
    \begin{flalign}
       f_{e_4^*} = x_{d_2^*,1} h_{d_2^*}
    \end{flalign}
    \begin{flalign}
       f_{e_5^*} = x_{d_2^*,2} h_{d_2^*}
    \end{flalign}
\end{subequations}
as well as objective
\begin{equation}
    \begin{aligned}
        F &= \frac{x_{d_1^*,2} h_{d_1^*}}{c_{e_1^*} - x_{d_1^*,2} h_{d_1^*}} + \frac{x_{d_1^*,1} h_{d_1^*}}{c_{e_2^*} - x_{d_1^*,1} h_{d_1^*}} \\
          &+ \frac{x_{d_1^*,2} h_{d_1^*} + x_{d_2^*,2} h_{d_2^*}}{c_{e_3^*} - x_{d_1^*,2} h_{d_1^*} - x_{d_2^*,2} h_{d_2^*}} \\
          &+ \frac{x_{d_2^*,1} h_{d_2^*}}{c_{e_4^*} -x_{d_2^*,1} h_{d_2^*}} + \frac{x_{d_2^*,2} h_{d_2^*}}{c_{e_5^*} - x_{d_2^*,2} h_{d_2^*}}
    \end{aligned}
\end{equation}

Variable pairs $(x_{d_1^*,1}$, $x_{d_1^*,2})$ and $(x_{d_2^*,1}$, $x_{d_2^*,2}$) interact because each pair handles the same demand (constraint (\ref{Model:c})).
On the other hand, variable pairs $(x_{d_1^*,1}, x_{d_2^*,2})$ and $(x_{d_1^*,2}, x_{d_2^*,1})$ do not interact. The underlying structure of the considered problem depends on link capacities and demand volumes that determine whether variables $x_{d_1^*,2}$ and $x_{d_2^*,2}$ interact or not. Suppose, $c_{e_3^*}$ is much larger than the other link capacities and much larger than the sum of $h_{d_1^*}$ and $h_{d_2^*}$. In that case, it is highly probable that variables $x_{d_1^*,2}$ and $x_{d_2^*,2}$ will not interact because link $e_3^*$ will handle both demands simultaneously. Thus, it is possible that each of these two variables ($x_{d_2^*,1}$ and $x_{d_2^*,2}$) can be optimized separately and the value of the other variable will not influence this process. If variables $x_{d_2^*,1}$ and $x_{d_2^*,2}$ are not interacting, then we can identify two separable subproblems that are non-additively separable because
\begin{equation}
    \frac{\partial^2 F}{\partial x_{d_1^*,2} \partial x_{d_2^*,2}} = \frac{2 h_{d_1^*} h_{d_2^*} c_{e_3^*}}{(c_{e_3^*} - x_{d_1^*,2} h_{d_1^*} - x_{d_2^*,2} h_{d_2^*})^2} \ne 0
\end{equation}
Otherwise, $x_{d_1^*,2}$ and $x_{d_2^*,2}$ are shared variables, and the problem is an overlapping one.

\subsection{Solution Encoding}

The multi-path routing problem is a constrained optimization problem. First, objective (\ref{Model:a}) is modified to handle constraint (\ref{Model:d}). The modification is as follows. If $f_e \ge c_e$, then we set the outcome of the fitness function to positive infinity. Such a modification works like penalty function~\cite{penalty_1, penalty_2}.

Second, to handle constraint (\ref{Model:c}) we define the following solution encoding. In the considered problem, for each demand $d$, we can represent its fractions using $|P(d)| - 1$ continuous variables $z_{d,1}$, ..., $z_{d,|P(d)|-1}$, where $\forall_{j \in \{1, ..., |P(d)|-1\}} z_{d,j} \in [0, 1]$. Let $\pi$ be a permutation of $\{1, ..., |P(d)|-1\}$ such that it indicates the ascending order of values $\{z_{d,1}, ..., z_{d,|P(d)|-1}\}$. Then $x_{d,p}$ can be calculated as follows
\begin{equation}
    \scriptsize
        x_{d,p} = 
    \begin{cases}
        z_{d,p} &, p < |P(d)| \land \pi(p) = 1 \\
        (1 - \sum_{j \in \{1, ...,\pi(p)-1\}} x_{d,\pi^{-1}(j)}) z_{d,p}  &, p < |P(d)| \land \pi(p) > 1 \\
        1 - \sum_{j \in \{1, ..., |P(d)|-1\}} x_{d,j} &, p = |P(d)|
    \end{cases}
\end{equation}
where $\pi^{-1}$ denotes the inverse function of $\pi$. Let us provide an example. For $z_{d^*,1} = 0.6$, $z_{d^*,2} = 0.3$, and $z_{d^*,3} = 0.9$, we get $\pi(1) = 2$, $\pi(2) = 1$, $\pi(3) = 3$, $x_{d^*,1} = 0.42$, $x_{d^*,2} = 0.3$, $x_{d^*,3} = 0.252$, and $x_{d^*,4} = 0.028$. The ascending order of values $\{z_{d,1}, ..., z_{d,|P(d)|-1}\}$ is used to eliminate plateaus from the fitness function. Please note that such a representation leads to obtain only feasible solutions in terms of constraint (\ref{Model:c}). Finally, solutions of each problem instance will be encoded using $\sum_{d \in D} (|P(d)| - 1)$ decision variables.

For such an encoding, we can show that the exemplary instance presented in Section~\ref{sec:cn:sep} can be either non-additively separable or overlapping for different values of link capacities and demand volumes. The example can be encoded using two decision variables, $z_{d_1^*,1}$ and $z_{d_2^*,1}$. Because of the two dimensions, this representation can be fully non-additively separable or fully non-separable. However, the fully non-separability means that the considered problem instance is overlapping, and the shared variables are $x_{d_1^*,2}$ and $x_{d_2^*,2}$. Let us consider the following values of link capacities and demand volumes: $c_{e_1^*} = 115$, $c_{e_2^*} = 120$, $c_{e_3^*} = 5000$, $c_{e_4^*} = 130$, $c_{e_5^*} = 125$, $h_{d_1^*} = 100$, and $h_{d_2^*} = 110$. We examine four different scenarios.
\begin{enumerate}
    \item \label{enum:cn:1} The second demand is fully allocated to its second path ($z_{d_2^*,1} = 0$). We search for the optimal fraction of the first demand flow (Fig.~\ref{fig:cn:sep:z1}).
    \item \label{enum:cn:2} The second demand is fully allocated to its first path ($z_{d_2^*,1} = 1$). We search for the optimal fraction of the first demand flow (Fig.~\ref{fig:cn:sep:z1}).
    \item \label{enum:cn:3} The first demand is fully allocated to its second path ($z_{d_1^*,1} = 0$). We search for the optimal fraction of the second demand flow (Fig.~\ref{fig:cn:sep:z2}).
    \item \label{enum:cn:4} The first demand is fully allocated to its first path ($z_{d_1^*,1} = 1$). We search for the optimal fraction of the second demand flow (Fig.~\ref{fig:cn:sep:z2}).
\end{enumerate}
According to Fig.~\ref{fig:cn:sep:z1}, the optimal solutions of cases \ref{enum:cn:1} and \ref{enum:cn:2} (the best value of $z_{d_1^*,1}$) are the same. Thus, it does not matter whether the second demand is fully allocated to a path that consists of common link $e_3^*$ or it is fully allocated to the first path, which does not use the common link. A similar situation is presented in Fig.~\ref{fig:cn:sep:z2}, which corresponds to scenarios \ref{enum:cn:3} and \ref{enum:cn:4}. The best value of $z_{d_2^*,1}$ can be found regardless of a decision on whether the first demand is fully allocated to a path consisting of the common link or not. The analysis of these four cases shows that variables $z_{d_1^*,1}$ and $z_{d_2^*,1}$ are not interacting. Thus, the considered problem instance is non-additively separable.

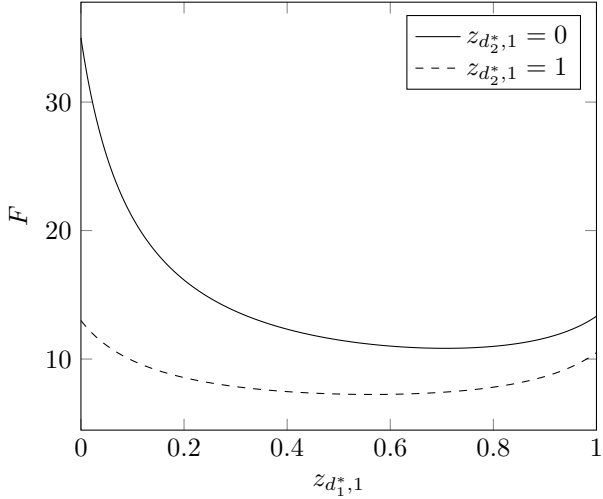
\begin{figure}
    \centering
    \begin{tikzpicture}
			\begin{axis}[%
			xmin=0,
			xmax=1,
			legend entries={$z_{d_2^*,1} = 0$, $z_{d_2^*,1} = 1$},
            legend pos = north east,
            xlabel = $z_{d_1^*,1}$,
            ylabel = $F$
			]
            \addplot[color=black,domain=0:1,samples=500]{((1-x)*100)/(115-(1-x)*100)+(x*100)/(120-x*100)+((1-x)*100+1*110)/(220-(1-x)*100-1*110)+(0*110)/(130-0*110)+(1*110)/(125-1*110)};
			\addplot[color=black,domain=0:1,dashed,samples=500]{((1-x)*100)/(115-(1-x)*100)+(x*100)/(120-x*100)+((1-x)*100+0*110)/(220-(1-x)*100-0*110)+(1*110)/(130-1*110)+(0*110)/(125-0*110)};
			\end{axis}
			\end{tikzpicture}
    \caption{Average delay in the network for different values of $z_{d_1^*,1}$ and constant values of $z_{d_2^*,1}$ (non-separable)}
    \label{fig:cn:nonsep}
\end{figure}

However, if we change the value of capacity of the common link, i.e., $c_{e_3^*}$, from $5000$ to $220$, then the considered problem instance is overlapping. According to Fig.~\ref{fig:cn:nonsep}, for two various values of $z_{d_2^*,1}$, the best value of $z_{d_1^*,1}$ differs.

\begin{figure}
    \centering
    \begin{tikzpicture}
			\begin{axis}[%
			xmin=0,
			xmax=1,
			legend entries={$z_{d_2^*,1} = 0$, $z_{d_2^*,1} = 1$},
            legend pos = north east,
            xlabel = $z_{d_1^*,1}$,
            ylabel = $F$
			]
            \addplot[color=black,domain=0:1,samples=500]{((1-x)*100)/(115-(1-x)*100)+(x*100)/(120-x*100)+((1-x)*100+1*110)/(5000-(1-x)*100-1*110)+(0*110)/(130-0*110)+(1*110)/(125-1*110)};
			\addplot[color=black,domain=0:1,dashed,samples=500]{((1-x)*100)/(115-(1-x)*100)+(x*100)/(120-x*100)+((1-x)*100+0*110)/(5000-(1-x)*100-0*110)+(1*110)/(130-1*110)+(0*110)/(125-0*110)};
			\end{axis}
			\end{tikzpicture}
    \caption{Average delay in the network for different values of $z_{d_1^*,1}$ and constant values of $z_{d_2^*,1}$ (possibly separable)}
    \label{fig:cn:sep:z1}
\end{figure}
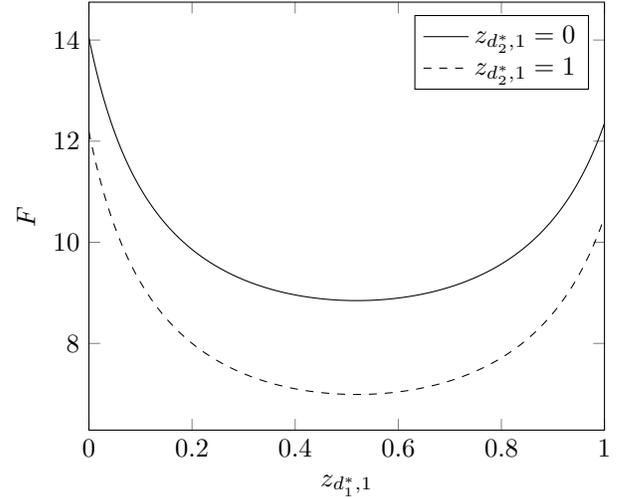

\begin{figure}
    \centering
    \begin{tikzpicture}
			\begin{axis}[%
			xmin=0,
			xmax=1,
			legend entries={$z_{d_1^*,1} = 0$, $z_{d_1^*,1} = 1$},
            legend pos = north east,
            xlabel = $z_{d_2^*,1}$,
            ylabel = $F$
			]
            \addplot[color=black,domain=0:1,samples=500]{(1*100)/(115-1*100)+(0*100)/(120-0*100)+(1*100+(1-x)*110)/(5000-1*100-(1-x)*110)+(x*110)/(130-x*110)+((1-x)*110)/(125-(1-x)*110)};
			\addplot[color=black,domain=0:1,dashed,samples=500]{(0*100)/(115-0*100)+(1*100)/(120-1*100)+(0*100+(1-x)*110)/(5000-0*100-(1-x)*110)+(x*110)/(130-x*110)+((1-x)*110)/(125-(1-x)*110)};
			\end{axis}
			\end{tikzpicture}
    \caption{Average delay in the network for different values of $z_{d_2^*,1}$ and constant values of $z_{d_1^*,1}$ (possibly separable)}
    \label{fig:cn:sep:z2}
\end{figure}

\subsection{Test Cases}

To conduct experiments, we randomly generated $120$ different test cases. Each test case consists of $63$ demands. For each demand, we created $17$ different paths. Thus, each test case has $1008$ dimensions. Such values were chosen to generate test cases whose size is similar to the size of functions from the CEC'2013 set. We considered six computer networks, i.e., 104, 114, 128, 144, 162, and g120~\cite{networks}. Their link capacities were generated from $10^6$ to $5 \cdot 10^8$. To consider the different underlying problem structure, we randomly generated demand volumes from four different ranges, such as $[1, 50]$, $[10, 500]$, $[100, 5 \cdot 10^3]$, and $[10^3, 5 \cdot 10^4]$. Five test cases were generated for each computer network and range of demand volumes. Let us group all test cases into four categories based on the range of demand volumes.
\begin{enumerate}
    \item C1: Demand volumes from $1$ to $50$.
    \item C2: Demand volumes from $10$ to $500$.
    \item C3: Demand volumes from $100$ to $5 \cdot 10^3$.
    \item C4: Demand volumes from $10^3$ to $5 \cdot 10^4$.
\end{enumerate}
For each pair of optimization method and test case, we conduct a single experiment.

\subsection{Optimization Methods’ Setup}
Since the size of generated instances of the multi-path routing problem is almost the same as the size of CEC'2013 functions, we used the same optimization method configuration. The computation budget was also set to $3 \cdot 10^6$ FFEs.

\subsection{Optimization Results}

Table~\ref{tab:ranking} presents the ranking of the considered optimization method for the multi-path routing problem. We performed the sign tests (significance level of $5\%$) with the Holm-Bonferroni correction to confirm whether the optimization results were significant. CBCC-IRRG achieves the smallest average ranking, whereas another IRRG-embedded CC framework takes second place along with SHADE-ILS. CBCC-RDG3 achieved the highest average ranking because RDG3 probably reports false linkage for the multi-path routing problem that consists of non-additively separable subproblems. The fourth place goes to CCFR2-FVIL. Similarly to $F_{12}$ (see Section~\ref{sec:fvil}), the order of variables is not rearranged for the considered instances of the multi-path routing problem. Thus, joining separable variables into groups of size $\epsilon_s = 100$ may mitigate the missing linkage issue.

The detailed optimization results are reported in Tables~\ref{tab:results_1_50}, \ref{tab:results_10_500}, \ref{tab:results_100_5000}, and \ref{tab:results_1000_50000}. Note that along with increasing the lower and upper ranges of demand volumes, the test cases become more overlapping. Moreover, their overlapping subproblems become more conflicting. Changes in the structure of test cases cause the decomposition provided by IRRG to become less beneficial for CC frameworks. 

\begin{table*}
    \setlength{\tabcolsep}{0.5em}
    \caption{Ranking of considered optimization methods for multi-path routing problem}
    \label{tab:ranking}
    \scriptsize
    \begin{tabularx}{\textwidth}{CCCCCCCC}
        \toprule
        Test case category & CBCC-IRRG & CCFR2-IRRG & CBCC-RDG3 & CCFR2-RDG3 & CBCC-FVIL & CCFR2-FVIL & SHADE-ILS \\
        \midrule
        
        C1 & \textbf{1} & \textbf{1} & 7 & 6 & 5 & 4 & 3 \\
        C2 & \textbf{1} & \textbf{1} & 7 & 6 & 5 & 4 & 3 \\
        C3 & 2 & \textbf{1} & 5 & 5 & 7 & 4 & 3 \\
        C4 & 6 & 2 & 7 & 3 & 3 & 3 & \textbf{1} \\
        \midrule
        Average & 2.50 & \textbf{1.25} & 6.50 & 5.00 & 5.00 & 3.75 & 2.50 \\
        \bottomrule
    \end{tabularx}
\end{table*}

\begin{table*}
    \setlength{\tabcolsep}{0.5em}
    \caption{Results for multi-path routing problem instances that consider demand volumes from $1$ to $50$ (C1)}
    \label{tab:results_1_50}
    \scriptsize
    \begin{tabularx}{\textwidth}{CCCCCCCC}
        \toprule
        Test case & CBCC-IRRG & CCFR2-IRRG & CBCC-RDG3 & CCFR2-RDG3 & CBCC-FVIL & CCFR2-FVIL & SHADE-ILS \\
        \midrule
        
        104\_1\_50\_0 & \textbf{2.50E-05} & \textbf{2.50E-05} & 5.11E-05 & 4.92E-05 & 3.93E-05 & 3.75E-05 & 2.71E-05 \\
        104\_1\_50\_1 & \textbf{2.53E-05} & \textbf{2.53E-05} & 4.52E-05 & 4.51E-05 & 3.77E-05 & 3.67E-05 & 2.72E-05 \\
        104\_1\_50\_2 & \textbf{2.21E-05} & \textbf{2.21E-05} & 3.00E-05 & 3.06E-05 & 3.22E-05 & 3.12E-05 & 2.36E-05 \\
        104\_1\_50\_3 & \textbf{2.42E-05} & \textbf{2.42E-05} & 6.01E-05 & 3.18E-05 & 3.27E-05 & 3.14E-05 & 2.52E-05 \\
        104\_1\_50\_4 & \textbf{2.46E-05} & \textbf{2.46E-05} & 3.11E-05 & 3.45E-05 & 4.07E-05 & 3.42E-05 & 2.60E-05 \\
        
        \addlinespace[0.5em]

        114\_1\_50\_0 & \textbf{2.26E-05} & \textbf{2.26E-05} & 4.98E-05 & 3.03E-05 & 3.58E-05 & 3.15E-05 & 2.43E-05 \\
        114\_1\_50\_1 & \textbf{2.10E-05} & \textbf{2.10E-05} & 4.29E-05 & 2.91E-05 & 3.11E-05 & 3.18E-05 & 2.14E-05 \\
        114\_1\_50\_2 & \textbf{1.91E-05} & \textbf{1.91E-05} & 3.80E-05 & 2.65E-05 & 2.90E-05 & 2.96E-05 & 1.99E-05 \\
        114\_1\_50\_3 & \textbf{2.23E-05} & \textbf{2.23E-05} & 3.36E-05 & 3.33E-05 & 3.20E-05 & 3.08E-05 & 2.28E-05 \\
        114\_1\_50\_4 & \textbf{2.13E-05} & \textbf{2.13E-05} & 3.94E-05 & 3.92E-05 & 3.26E-05 & 2.91E-05 & 2.20E-05 \\
        
        \addlinespace[0.5em]
        
        128\_1\_50\_0 & \textbf{1.87E-05} & \textbf{1.87E-05} & 3.77E-05 & 2.98E-05 & 2.72E-05 & 2.68E-05 & 1.91E-05 \\
        128\_1\_50\_1 & \textbf{1.85E-05} & \textbf{1.85E-05} & 3.49E-05 & 2.65E-05 & 2.59E-05 & 2.50E-05 & 1.90E-05 \\
        128\_1\_50\_2 & \textbf{1.64E-05} & \textbf{1.64E-05} & 3.11E-05 & 3.11E-05 & 2.31E-05 & 2.20E-05 & 1.69E-05 \\
        128\_1\_50\_3 & 1.81E-05 & \textbf{1.78E-05} & 3.13E-05 & 3.14E-05 & 2.53E-05 & 2.56E-05 & 1.81E-05 \\
        128\_1\_50\_4 & \textbf{1.60E-05} & \textbf{1.60E-05} & 2.79E-05 & 2.12E-05 & 2.59E-05 & 2.61E-05 & 1.63E-05 \\
        
        \addlinespace[0.5em]
        
        144\_1\_50\_0 & \textbf{1.84E-05} & \textbf{1.84E-05} & 3.06E-05 & 2.47E-05 & 2.50E-05 & 2.36E-05 & 1.89E-05 \\
        144\_1\_50\_1 & \textbf{1.78E-05} & \textbf{1.78E-05} & 2.90E-05 & 2.86E-05 & 2.25E-05 & 2.26E-05 & 1.80E-05 \\
        144\_1\_50\_2 & \textbf{1.54E-05} & \textbf{1.54E-05} & 2.63E-05 & 2.60E-05 & 1.98E-05 & 1.87E-05 & 1.55E-05 \\
        144\_1\_50\_3 & \textbf{1.76E-05} & \textbf{1.76E-05} & 2.82E-05 & 2.52E-05 & 2.19E-05 & 2.23E-05 & 1.83E-05 \\
        144\_1\_50\_4 & \textbf{1.56E-05} & \textbf{1.56E-05} & 2.41E-05 & 2.35E-05 & 2.25E-05 & 2.20E-05 & 1.65E-05 \\
        
        \addlinespace[0.5em]
        
        162\_1\_50\_0 & \textbf{1.64E-05} & \textbf{1.64E-05} & 2.09E-05 & 2.12E-05 & 2.19E-05 & 2.20E-05 & 1.71E-05 \\
        162\_1\_50\_1 & \textbf{1.51E-05} & \textbf{1.51E-05} & 2.26E-05 & 2.25E-05 & 2.06E-05 & 1.89E-05 & 1.57E-05 \\
        162\_1\_50\_2 & \textbf{1.38E-05} & \textbf{1.38E-05} & 2.19E-05 & 2.18E-05 & 1.87E-05 & 1.71E-05 & 1.42E-05 \\
        162\_1\_50\_3 & \textbf{1.52E-05} & \textbf{1.52E-05} & 2.11E-05 & 2.15E-05 & 1.98E-05 & 2.00E-05 & 1.55E-05 \\
        162\_1\_50\_4 & \textbf{1.47E-05} & \textbf{1.47E-05} & 2.36E-05 & 2.19E-05 & 1.92E-05 & 1.91E-05 & 1.59E-05 \\
        
        \addlinespace[0.5em]
        
        g120\_1\_50\_0 & \textbf{2.47E-05} & \textbf{2.47E-05} & 3.25E-05 & 3.25E-05 & 3.21E-05 & 3.20E-05 & 2.51E-05 \\
        g120\_1\_50\_1 & \textbf{2.82E-05} & \textbf{2.82E-05} & 3.49E-05 & 3.48E-05 & 3.48E-05 & 3.31E-05 & 2.87E-05 \\
        g120\_1\_50\_2 & \textbf{2.21E-05} & \textbf{2.21E-05} & 3.22E-05 & 3.22E-05 & 3.23E-05 & 2.87E-05 & 2.25E-05 \\
        g120\_1\_50\_3 & \textbf{2.57E-05} & \textbf{2.57E-05} & 3.27E-05 & 3.33E-05 & 3.17E-05 & 3.29E-05 & 2.63E-05 \\
        g120\_1\_50\_4 & \textbf{2.63E-05} & \textbf{2.63E-05} & 3.25E-05 & 3.21E-05 & 3.40E-05 & 3.39E-05 & 2.70E-05 \\
        
        \bottomrule
    \end{tabularx}
\end{table*}

\begin{table*}
    \setlength{\tabcolsep}{0.5em}
    \caption{Results for multi-path routing problem instances that consider demand volumes from $10$ to $500$ (C2)}
    \label{tab:results_10_500}
    \scriptsize
    \begin{tabularx}{\textwidth}{CCCCCCCC}
        \toprule
        Test case & CBCC-IRRG & CCFR2-IRRG & CBCC-RDG3 & CCFR2-RDG3 & CBCC-FVIL & CCFR2-FVIL & SHADE-ILS \\
        \midrule
        
        104\_10\_500\_0 & \textbf{2.59E-04} & \textbf{2.59E-04} & 4.99E-04 & 5.06E-04 & 4.00E-04 & 3.82E-04 & 2.70E-04 \\
        104\_10\_500\_1 & 3.97E-04 & \textbf{2.38E-04} & 4.31E-04 & 4.30E-04 & 3.79E-04 & 3.38E-04 & 2.43E-04 \\
        104\_10\_500\_2 & \textbf{2.71E-04} & \textbf{2.71E-04} & 3.70E-04 & 3.68E-04 & 4.43E-04 & 3.96E-04 & 2.86E-04 \\
        104\_10\_500\_3 & \textbf{2.50E-04} & \textbf{2.50E-04} & 6.26E-04 & 3.26E-04 & 3.36E-04 & 3.30E-04 & 2.56E-04 \\
        104\_10\_500\_4 & 3.96E-04 & \textbf{2.30E-04} & 2.84E-04 & 5.40E-04 & 3.48E-04 & 3.29E-04 & 2.34E-04 \\
        
        \addlinespace[0.5em]
        
        114\_10\_500\_0 & \textbf{2.34E-04} & \textbf{2.34E-04} & 3.20E-04 & 3.10E-04 & 3.41E-04 & 3.11E-04 & 2.40E-04 \\
        114\_10\_500\_1 & \textbf{1.96E-04} & \textbf{1.96E-04} & 3.87E-04 & 2.62E-04 & 3.05E-04 & 3.08E-04 & 2.10E-04 \\
        114\_10\_500\_2 & \textbf{2.26E-04} & \textbf{2.26E-04} & 4.67E-04 & 3.10E-04 & 3.49E-04 & 3.39E-04 & 2.41E-04 \\
        114\_10\_500\_3 & \textbf{2.17E-04} & \textbf{2.17E-04} & 4.09E-04 & 3.24E-04 & 3.07E-04 & 3.05E-04 & 2.21E-04 \\
        114\_10\_500\_4 & \textbf{1.99E-04} & \textbf{1.99E-04} & 3.78E-04 & 3.78E-04 & 2.99E-04 & 3.07E-04 & 2.05E-04 \\
        
        \addlinespace[0.5em]
        
        128\_10\_500\_0 & \textbf{1.94E-04} & \textbf{1.94E-04} & 3.85E-04 & 3.08E-04 & 2.68E-04 & 2.76E-04 & 2.00E-04 \\
        128\_10\_500\_1 & \textbf{1.72E-04} & \textbf{1.72E-04} & 2.57E-04 & 2.51E-04 & 2.35E-04 & 2.37E-04 & 1.85E-04 \\
        128\_10\_500\_2 & \textbf{1.95E-04} & \textbf{1.95E-04} & 3.92E-04 & 3.77E-04 & 2.97E-04 & 2.64E-04 & 2.03E-04 \\
        128\_10\_500\_3 & 1.77E-04 & \textbf{1.73E-04} & 3.22E-04 & 3.19E-04 & 2.40E-04 & 2.28E-04 & 1.80E-04 \\
        128\_10\_500\_4 & 1.45E-04 & \textbf{1.44E-04} & 2.91E-04 & 2.04E-04 & 2.49E-04 & 2.45E-04 & 1.48E-04 \\
        
        \addlinespace[0.5em]
        
        144\_10\_500\_0 & \textbf{1.88E-04} & \textbf{1.88E-04} & 2.99E-04 & 2.96E-04 & 2.63E-04 & 2.54E-04 & 1.97E-04 \\
        144\_10\_500\_1 & \textbf{1.62E-04} & \textbf{1.62E-04} & 2.80E-04 & 2.29E-04 & 2.21E-04 & 2.22E-04 & 1.70E-04 \\
        144\_10\_500\_2 & 1.84E-04 & \textbf{1.82E-04} & 2.99E-04 & 2.95E-04 & 2.54E-04 & 2.42E-04 & 1.91E-04 \\
        144\_10\_500\_3 & \textbf{1.73E-04} & \textbf{1.73E-04} & 2.86E-04 & 2.32E-04 & 2.24E-04 & 2.22E-04 & 1.76E-04 \\
        144\_10\_500\_4 & \textbf{1.43E-04} & \textbf{1.43E-04} & 2.23E-04 & 2.27E-04 & 2.14E-04 & 2.14E-04 & 1.46E-04 \\
        
        \addlinespace[0.5em]
        
        162\_10\_500\_0 & \textbf{1.64E-04} & \textbf{1.64E-04} & 2.12E-04 & 2.10E-04 & 2.27E-04 & 2.20E-04 & 1.67E-04 \\
        162\_10\_500\_1 & \textbf{1.42E-04} & \textbf{1.42E-04} & 2.48E-04 & 2.46E-04 & 1.97E-04 & 1.90E-04 & 1.51E-04 \\
        162\_10\_500\_2 & \textbf{1.64E-04} & 1.68E-04 & 2.59E-04 & 2.15E-04 & 2.33E-04 & 2.18E-04 & 1.69E-04 \\
        162\_10\_500\_3 & \textbf{1.48E-04} & \textbf{1.48E-04} & 2.34E-04 & 2.13E-04 & 2.08E-04 & 1.88E-04 & 1.50E-04 \\
        162\_10\_500\_4 & \textbf{1.38E-04} & \textbf{1.38E-04} & 2.18E-04 & 2.03E-04 & 1.98E-04 & 1.78E-04 & 1.41E-04 \\
        
        \addlinespace[0.5em]
        
        g120\_10\_500\_0 & \textbf{2.50E-04} & \textbf{2.50E-04} & 3.23E-04 & 3.04E-04 & 3.13E-04 & 3.06E-04 & 2.55E-04 \\
        g120\_10\_500\_1 & \textbf{2.64E-04} & \textbf{2.64E-04} & 3.32E-04 & 3.32E-04 & 3.32E-04 & 3.26E-04 & 2.70E-04 \\
        g120\_10\_500\_2 & \textbf{2.65E-04} & \textbf{2.65E-04} & 3.64E-04 & 3.66E-04 & 3.65E-04 & 3.26E-04 & 2.70E-04 \\
        g120\_10\_500\_3 & \textbf{2.41E-04} & \textbf{2.41E-04} & 3.37E-04 & 2.82E-04 & 2.97E-04 & 3.02E-04 & 2.46E-04 \\
        g120\_10\_500\_4 & 2.49E-04 & \textbf{2.44E-04} & 3.10E-04 & 3.09E-04 & 3.18E-04 & 3.01E-04 & 2.52E-04 \\
        
        \bottomrule
    \end{tabularx}
\end{table*}

\begin{table*}
    \setlength{\tabcolsep}{0.5em}
    \caption{Results for multi-path routing problem instances that consider demand volumes from $100$ to $5 \cdot 10^3$ (C3)}
    \label{tab:results_100_5000}
    \scriptsize
    \begin{tabularx}{\textwidth}{CCCCCCCC}
        \toprule
        Test case & CBCC-IRRG & CCFR2-IRRG & CBCC-RDG3 & CCFR2-RDG3 & CBCC-FVIL & CCFR2-FVIL & SHADE-ILS \\
        \midrule
        
        104\_100\_5000\_0 & 4.03E-03 & 4.54E-03 & 4.20E-03 & 4.18E-03 & 3.23E-03 & 3.13E-03 & \textbf{2.29E-03} \\
        104\_100\_5000\_1 & 4.16E-03 & \textbf{1.89E-03} & 3.51E-03 & 3.51E-03 & 3.08E-03 & 2.97E-03 & 1.97E-03 \\
        104\_100\_5000\_2 & \textbf{2.25E-03} & 5.11E-03 & 2.98E-03 & 2.98E-03 & 3.52E-03 & 3.24E-03 & 2.35E-03 \\
        104\_100\_5000\_3 & 5.33E-03 & \textbf{2.26E-03} & 5.45E-03 & 3.09E-03 & 3.04E-03 & 3.00E-03 & 2.35E-03 \\
        104\_100\_5000\_4 & 2.98E-03 & 5.90E-03 & 3.05E-03 & 3.02E-03 & 3.56E-03 & 3.53E-03 & \textbf{2.39E-03} \\
        
        \addlinespace[0.5em]
        
        114\_100\_5000\_0 & \textbf{2.05E-03} & 2.06E-03 & 4.31E-03 & 3.11E-03 & 2.93E-03 & 2.75E-03 & 2.15E-03 \\
        114\_100\_5000\_1 & \textbf{1.74E-03} & 3.75E-03 & 2.28E-03 & 2.32E-03 & 2.57E-03 & 2.62E-03 & 1.82E-03 \\
        114\_100\_5000\_2 & \textbf{1.94E-03} & 3.08E-03 & 2.62E-03 & 3.67E-03 & 2.86E-03 & 2.71E-03 & 1.99E-03 \\
        114\_100\_5000\_3 & 3.15E-03 & \textbf{2.03E-03} & 3.75E-03 & 2.69E-03 & 2.97E-03 & 2.82E-03 & 2.09E-03 \\
        114\_100\_5000\_4 & \textbf{1.91E-03} & 1.95E-03 & 3.78E-03 & 3.72E-03 & 3.19E-03 & 2.94E-03 & 2.00E-03 \\
        
        \addlinespace[0.5em]
        
        128\_100\_5000\_0 & 3.56E-03 & \textbf{1.67E-03} & 2.66E-03 & 2.62E-03 & 2.36E-03 & 2.11E-03 & 1.76E-03 \\
        128\_100\_5000\_1 & \textbf{1.48E-03} & 1.49E-03 & 2.57E-03 & 2.58E-03 & 2.09E-03 & 2.06E-03 & 1.50E-03 \\
        128\_100\_5000\_2 & 1.70E-03 & \textbf{1.57E-03} & 3.00E-03 & 2.96E-03 & 2.18E-03 & 2.19E-03 & 1.61E-03 \\
        128\_100\_5000\_3 & 1.82E-03 & \textbf{1.64E-03} & 2.50E-03 & 2.73E-03 & 2.32E-03 & 2.26E-03 & 1.66E-03 \\
        128\_100\_5000\_4 & \textbf{1.43E-03} & \textbf{1.43E-03} & 2.72E-03 & 2.08E-03 & 2.51E-03 & 2.27E-03 & 1.48E-03 \\
        
        \addlinespace[0.5em]
        
        144\_100\_5000\_0 & \textbf{1.66E-03} & 1.67E-03 & 2.05E-03 & 2.47E-03 & 2.13E-03 & 2.10E-03 & 1.70E-03 \\
        144\_100\_5000\_1 & \textbf{1.44E-03} & \textbf{1.44E-03} & 2.36E-03 & 1.96E-03 & 1.88E-03 & 1.87E-03 & 1.49E-03 \\
        144\_100\_5000\_2 & 1.49E-03 & \textbf{1.47E-03} & 2.43E-03 & 2.34E-03 & 2.03E-03 & 1.84E-03 & 1.48E-03 \\
        144\_100\_5000\_3 & \textbf{1.64E-03} & \textbf{1.64E-03} & 2.67E-03 & 2.64E-03 & 2.12E-03 & 2.05E-03 & 1.68E-03 \\
        144\_100\_5000\_4 & \textbf{1.42E-03} & \textbf{1.42E-03} & 2.26E-03 & 2.22E-03 & 2.15E-03 & 2.11E-03 & 1.46E-03 \\
        
        \addlinespace[0.5em]
        
        162\_100\_5000\_0 & 1.85E-03 & \textbf{1.47E-03} & 2.23E-03 & 1.89E-03 & 2.05E-03 & 1.86E-03 & 1.51E-03 \\
        162\_100\_5000\_1 & \textbf{1.24E-03} & \textbf{1.24E-03} & 2.18E-03 & 2.10E-03 & 1.82E-03 & 1.78E-03 & 1.27E-03 \\
        162\_100\_5000\_2 & 1.62E-03 & \textbf{1.37E-03} & 2.22E-03 & 2.13E-03 & 1.81E-03 & 1.74E-03 & 1.41E-03 \\
        162\_100\_5000\_3 & \textbf{1.46E-03} & \textbf{1.46E-03} & 1.94E-03 & 1.96E-03 & 1.90E-03 & 1.92E-03 & 1.52E-03 \\
        162\_100\_5000\_4 & 1.34E-03 & \textbf{1.33E-03} & 2.24E-03 & 1.98E-03 & 2.00E-03 & 1.90E-03 & 1.38E-03 \\
        
        \addlinespace[0.5em]
        
        g120\_100\_5000\_0 & 2.13E-03 & \textbf{2.10E-03} & 2.94E-03 & 2.91E-03 & 2.90E-03 & 2.69E-03 & 2.19E-03 \\
        g120\_100\_5000\_1 & \textbf{2.28E-03} & \textbf{2.28E-03} & 2.81E-03 & 2.80E-03 & 2.92E-03 & 2.82E-03 & 2.32E-03 \\
        g120\_100\_5000\_2 & \textbf{2.24E-03} & \textbf{2.24E-03} & 3.17E-03 & 3.18E-03 & 3.03E-03 & 3.07E-03 & 2.33E-03 \\
        g120\_100\_5000\_3 & \textbf{2.39E-03} & 2.40E-03 & 2.74E-03 & 2.78E-03 & 2.88E-03 & 2.91E-03 & 2.43E-03 \\
        g120\_100\_5000\_4 & \textbf{2.54E-03} & \textbf{2.54E-03} & 3.21E-03 & 3.13E-03 & 3.25E-03 & 3.34E-03 & 2.61E-03 \\
        
        \bottomrule
    \end{tabularx}
\end{table*}

\begin{table*}
    \setlength{\tabcolsep}{0.5em}
    \caption{Results for multi-path routing problem instances that consider demand volumes from $10^3$ to $5 \cdot 10^4$ (C4)}
    \label{tab:results_1000_50000}
    \scriptsize
    \begin{tabularx}{\textwidth}{CCCCCCCC}
        \toprule
        Test case & CBCC-IRRG & CCFR2-IRRG & CBCC-RDG3 & CCFR2-RDG3 & CBCC-FVIL & CCFR2-FVIL & SHADE-ILS \\
        \midrule
        
        104\_1000\_50000\_0 & 5.35E-02 & 2.88E-02 & 5.22E-02 & 4.24E-02 & 3.21E-02 & 3.16E-02 & \textbf{2.37E-02} \\
        104\_1000\_50000\_1 & 3.10E-02 & 2.90E-02 & 4.13E-02 & 4.01E-02 & 3.62E-02 & 3.57E-02 & \textbf{2.53E-02} \\
        104\_1000\_50000\_2 & 3.43E-02 & 3.31E-02 & 3.48E-02 & 3.51E-02 & 3.52E-02 & 3.57E-02 & \textbf{2.75E-02} \\
        104\_1000\_50000\_3 & 3.65E-02 & 3.66E-02 & 7.43E-02 & 3.82E-02 & 4.10E-02 & 3.85E-02 & \textbf{2.95E-02} \\
        104\_1000\_50000\_4 & 2.71E-02 & 2.77E-02 & 3.05E-02 & 3.08E-02 & 3.51E-02 & 3.04E-02 & \textbf{2.31E-02} \\
        
        \addlinespace[0.5em]
        
        114\_1000\_50000\_0 & 2.69E-02 & 2.77E-02 & 3.97E-02 & 2.88E-02 & 2.85E-02 & 3.05E-02 & \textbf{2.22E-02} \\
        114\_1000\_50000\_1 & 2.63E-02 & 2.59E-02 & 4.01E-02 & 2.75E-02 & 3.20E-02 & 2.86E-02 & \textbf{2.18E-02} \\
        114\_1000\_50000\_2 & 2.86E-02 & 2.83E-02 & 2.99E-02 & 3.03E-02 & 3.42E-02 & 3.24E-02 & \textbf{2.37E-02} \\
        114\_1000\_50000\_3 & 3.31E-02 & 3.26E-02 & 3.95E-02 & 3.91E-02 & 3.81E-02 & 3.80E-02 & \textbf{2.75E-02} \\
        114\_1000\_50000\_4 & 2.35E-02 & 2.31E-02 & 3.48E-02 & 3.44E-02 & 2.65E-02 & 2.65E-02 & \textbf{1.94E-02} \\
        
        \addlinespace[0.5em]
        
        128\_1000\_50000\_0 & 2.24E-02 & 2.23E-02 & 3.32E-02 & 2.51E-02 & 2.37E-02 & 2.43E-02 & \textbf{1.87E-02} \\
        128\_1000\_50000\_1 & 3.50E-02 & 2.13E-02 & 3.24E-02 & 2.98E-02 & 2.44E-02 & 2.46E-02 & \textbf{1.80E-02} \\
        128\_1000\_50000\_2 & 2.26E-02 & 2.24E-02 & 3.50E-02 & 3.26E-02 & 2.57E-02 & 2.40E-02 & \textbf{1.89E-02} \\
        128\_1000\_50000\_3 & 2.59E-02 & 2.67E-02 & 4.35E-02 & 4.08E-02 & 3.01E-02 & 2.83E-02 & \textbf{2.28E-02} \\
        128\_1000\_50000\_4 & 1.76E-02 & 1.76E-02 & 2.45E-02 & 2.43E-02 & 2.25E-02 & 2.16E-02 & \textbf{1.40E-02} \\
        
        \addlinespace[0.5em]
        
        144\_1000\_50000\_0 & 2.10E-02 & 3.03E-02 & 2.45E-02 & 2.51E-02 & 2.38E-02 & 2.38E-02 & \textbf{1.82E-02} \\
        144\_1000\_50000\_1 & 1.93E-02 & 1.97E-02 & 2.64E-02 & 2.28E-02 & 2.28E-02 & 2.18E-02 & \textbf{1.68E-02} \\
        144\_1000\_50000\_2 & 3.41E-02 & 2.02E-02 & 2.83E-02 & 2.11E-02 & 2.20E-02 & 2.17E-02 & \textbf{1.83E-02} \\
        144\_1000\_50000\_3 & 2.52E-02 & 2.45E-02 & 3.72E-02 & 3.15E-02 & 2.79E-02 & 2.67E-02 & \textbf{2.19E-02} \\
        144\_1000\_50000\_4 & 2.77E-02 & 1.77E-02 & 2.13E-02 & 2.09E-02 & 2.01E-02 & 2.01E-02 & \textbf{1.44E-02} \\
        
        \addlinespace[0.5em]
       
        162\_1000\_50000\_0 & 2.60E-02 & 3.15E-02 & 2.22E-02 & 2.03E-02 & 2.14E-02 & 1.93E-02 & \textbf{1.58E-02} \\
        162\_1000\_50000\_1 & 1.72E-02 & 1.70E-02 & 2.44E-02 & 2.31E-02 & 2.03E-02 & 1.98E-02 & \textbf{1.49E-02} \\
        162\_1000\_50000\_2 & 2.75E-02 & 1.86E-02 & 2.44E-02 & 2.39E-02 & 2.15E-02 & 2.08E-02 & \textbf{1.60E-02} \\
        162\_1000\_50000\_3 & 3.22E-02 & 2.10E-02 & 3.07E-02 & 2.56E-02 & 2.45E-02 & 2.30E-02 & \textbf{1.98E-02} \\
        162\_1000\_50000\_4 & 1.59E-02 & 1.51E-02 & 2.09E-02 & 1.87E-02 & 1.73E-02 & 1.72E-02 & \textbf{1.37E-02} \\
        
        \addlinespace[0.5em]
        
        g120\_1000\_50000\_0 & 3.76E-02 & 3.57E-02 & 3.32E-02 & 3.34E-02 & 3.23E-02 & 3.25E-02 & \textbf{2.31E-02} \\
        g120\_1000\_50000\_1 & 3.01E-02 & 3.01E-02 & 3.36E-02 & 3.36E-02 & 3.45E-02 & 3.35E-02 & \textbf{2.79E-02} \\
        g120\_1000\_50000\_2 & 3.87E-02 & 3.88E-02 & 3.30E-02 & 3.28E-02 & 3.16E-02 & 3.14E-02 & \textbf{2.44E-02} \\
        g120\_1000\_50000\_3 & 4.63E-02 & 3.73E-02 & 4.35E-02 & 3.61E-02 & 3.63E-02 & 3.67E-02 & \textbf{3.07E-02} \\
        g120\_1000\_50000\_4 & 3.84E-02 & 3.76E-02 & 2.87E-02 & 2.92E-02 & 3.11E-02 & 2.96E-02 & \textbf{2.37E-02} \\
        
        \bottomrule
    \end{tabularx}
\end{table*}

\ifCLASSOPTIONcaptionsoff
  \newpage
\fi

\bibliographystyle{IEEEtran}
\bibliography{bibl-ps}
